\pdfoutput=1
\documentclass{article}
\PassOptionsToPackage{numbers,sort&compress}{natbib}
\usepackage[final,main]{neurips_2026}

\usepackage[utf8]{inputenc}
\usepackage[T1]{fontenc}
\usepackage{url}
\usepackage{booktabs}
\usepackage{multirow}
\usepackage{tabularx}
\usepackage{longtable}
\usepackage{array}
\usepackage{float}
\usepackage{placeins}
\usepackage{needspace}
\usepackage{amsfonts}
\usepackage{amsmath,amssymb,amsthm}
\usepackage{mathtools}
\usepackage{nicefrac}
\usepackage{microtype}
\usepackage{xcolor}
\usepackage{algorithm}
\usepackage{algpseudocode}
\usepackage{graphicx}
\usepackage{tikz}
\usetikzlibrary{arrows.meta, positioning, shapes.geometric, fit, backgrounds, calc}
\usepackage{hyperref}
\usepackage[capitalize,noabbrev]{cleveref}
\crefname{proposition}{Proposition}{Propositions}
\Crefname{proposition}{Proposition}{Propositions}
\crefname{theorem}{Theorem}{Theorems}
\Crefname{theorem}{Theorem}{Theorems}
\crefname{corollary}{Corollary}{Corollaries}
\Crefname{corollary}{Corollary}{Corollaries}
\crefname{lemma}{Lemma}{Lemmas}
\Crefname{lemma}{Lemma}{Lemmas}
\crefname{assumption}{Assumption}{Assumptions}
\Crefname{assumption}{Assumption}{Assumptions}

\newcommand{\Uale}{U_{\mathrm{ale}}}
\newcommand{\Uepi}{U_{\mathrm{epi}}}
\newcommand{\Utot}{U_{\mathrm{total}}}

\newcommand{\tauhat}{\hat{\tau}}
\newcommand{\lambdastar}{\lambda^\star}
\newcommand{\wlam}{w_\lambda}
\newcommand{\Cset}{\mathcal{C}}
\newcommand{\Mmodel}{\mathcal{M}}
\newcommand{\Dcal}{\mathcal{D}_{\mathrm{cal}}}
\newcommand{\Vzero}{\mathcal{V}_{0}}
\newcommand{\Vone}{\mathcal{V}_{1}}
\newcommand{\Vtwo}{\mathcal{V}_{2}}

\newcommand{\APSS}{\mathrm{APSS}}
\newcommand{\snorm}{\tilde{s}}

\theoremstyle{plain}
\newtheorem{theorem}{Theorem}
\newtheorem{proposition}{Proposition}

\newtheorem{corollary}{Corollary}
\theoremstyle{definition}

\newtheorem{assumption}{Assumption}
\theoremstyle{remark}
\newtheorem{remark}{Remark}

\title{Matchability-Aware Conformal Prediction for Open-Ended Language Model Generation}

\author{%
  Ilia Azizi \\
  University of Lausanne, Switzerland \\
  BegooAI, Switzerland \\
  \href{mailto:ilia@begoo.ai}{\texttt{ilia@begoo.ai}}
}
\hypersetup{pdfauthor={Ilia Azizi},pdftitle={Matchability-Aware Conformal Prediction for Open-Ended Language Model Generation}}

\begin{document}

\maketitle

\begin{abstract}
Large language models (LLMs) often do not produce calibrated uncertainty estimates for their answers. We propose CAELUM, a framework that builds semantic prediction sets while separating aleatoric from epistemic uncertainty. CAELUM samples answers using an ensemble of clarification prompts, clusters them by semantic equivalence, and estimates the aleatoric and epistemic components via kernel-based von Neumann entropy before calibrating nonconformity scores with split conformal prediction. We use a weight $w_\lambda(x)=1+U_{\mathrm{ale}}(x)+\lambda\,U_{\mathrm{epi}}(x)$ to control for the contribution of the epistemic uncertainty of each instance and fit a coefficient $\lambda^\star$ to minimize prediction set size. When the objective is nonflat and selection is stable across splits, $\lambda^\star$ could be a diagnostic, where large values may indicate epistemic dominance (retrieve, abstain) and $\lambda^\star\to 0$ aleatoric dominance (clarify). For individual queries, a log ratio $R(x)$ compares the weighted epistemic component with the aleatoric component. Theoretically, we demonstrate that the coverage is capped by \emph{matchability}, the fraction of items whose reference answer appears in any sampled cluster. Empirically, this matchability can be predicted from the original question features using a lightweight classifier. We use those predictions to define a stratum and derive a conformal coverage bound within it. Across $30$ LLM and dataset combinations (4--32B), mean coverage is $0.74$ in the learned stratum versus $0.65$ overall, averaged over the $78$ evaluations using the learned classifier.
\mbox{Project page:} \url{https://unco3892.github.io/caelum/}
\end{abstract}

\section{Introduction}\label{sec:introduction}

Large language models (LLMs)~\citep{openai2023gpt4,touvron2023llama} are increasingly deployed in settings where model hallucinations can have practical consequences~\citep{farquhar2024detecting}. Assessing reliability also requires understanding underspecified requests, interactions, and how uncertainty is communicated to users~\citep{kirchhof2025reassessment}. Conformal methods~\citep{vovk2005,angelopoulos2024} offer finite-sample marginal coverage guarantees for regression and classification tasks under exchangeability. However, considerable obstacles exist to applying conformal methods to open-ended text generation tasks, including a large response space, discrepancies between semantic and exact-match response accuracy, and limitations of a single conformal parameter in separating input ambiguity from knowledge gaps.

\begin{figure}[t]
\centering
\setlength{\abovecaptionskip}{4pt}
\input{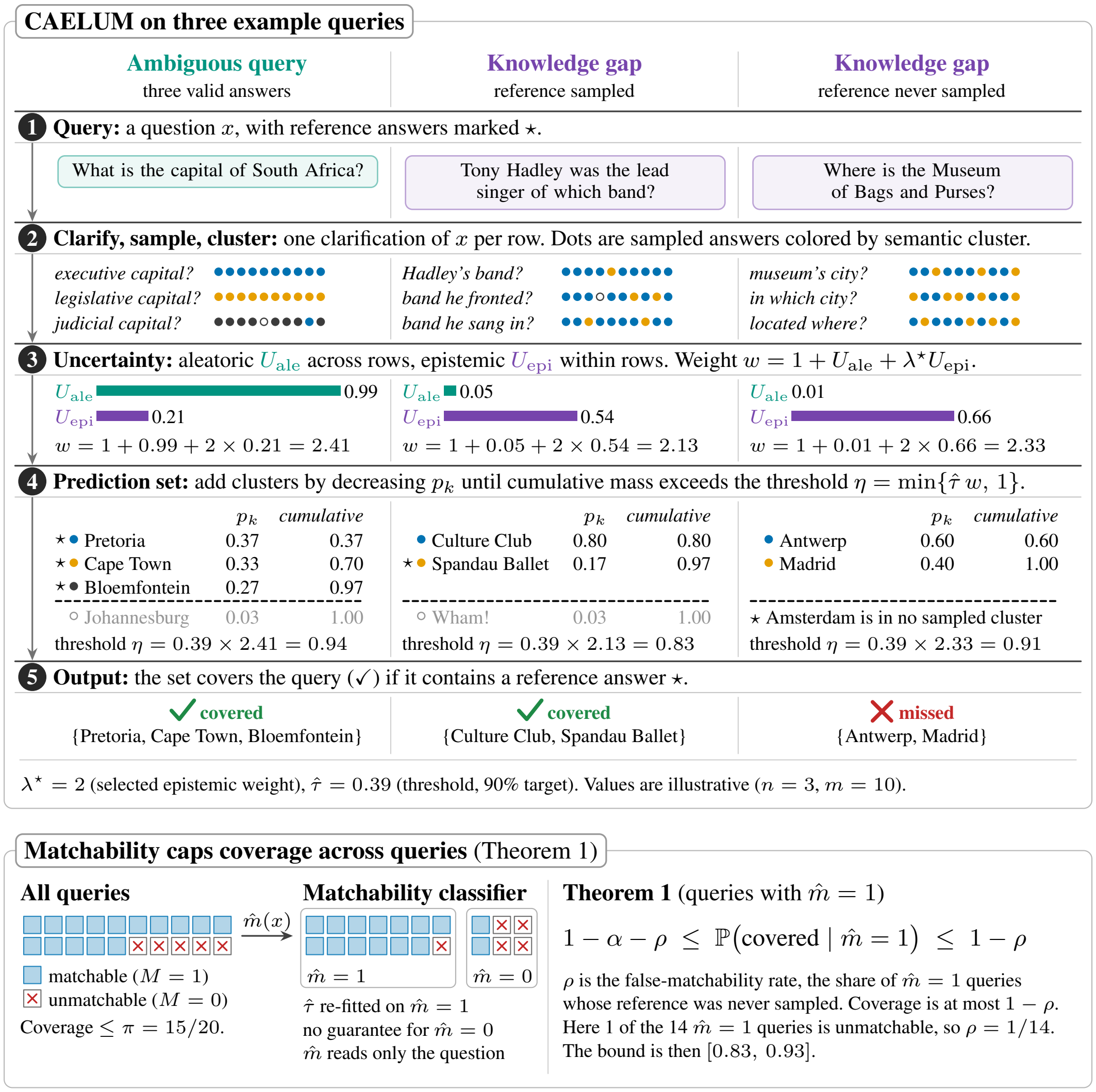}
\caption{\textbf{Method overview.} Each top-panel column shows the same five steps. Disagreement between clarifications raises $\Uale$, and disagreement within a clarification raises $\Uepi$. The weight $w$ sets the effective threshold $\eta=\min\{\hat\tau w,1\}$ for selecting clusters in \cref{eq:prediction_set}. The third column's reference answer is absent from sampled support, so expanding the set cannot include it. The lower panel shows the matchability bound. Main experiments use $n=5$ and kernel entropy.}

\label{fig:overview}
\end{figure}

We introduce Calibrated Aleatoric-Epistemic Language Uncertainty Modeling (CAELUM), which constructs semantic prediction sets from clarification ensembles. The idea is to distinguish the two sources of uncertainty~\citep{kendall2017uncertainties,hullermeier2021aleatoric}, namely aleatoric and epistemic uncertainty. \emph{Aleatoric uncertainty} arises from input ambiguity and is irreducible given the query. For example, ``What is the capital of South Africa?'' has three valid answers depending on which branch of government the user means. \emph{Epistemic uncertainty} reflects the model's lack of knowledge, for example where training data are sparse. Distinguishing these sources can inform different downstream responses.

Existing methods address different aspects of uncertainty estimation and calibration in open-ended language generation. Semantic entropy~\citep{kuhn2023semantic,farquhar2024detecting} clusters semantically equivalent LLM generations and computes entropy over the resulting meaning distribution. Kernel language entropy~\citep{nikitin2024kernel} extends this approach using semantic similarity kernels and von Neumann entropy. These methods use uncertainty estimates to detect hallucinations, without decomposing uncertainty into aleatoric and epistemic components or constructing prediction sets with conformal coverage guarantees. The spectral framework of \citet{spectraluncertainty2025} decomposes uncertainty into aleatoric and epistemic components but evaluates uncertainty \emph{ranking} without constructing prediction sets. ConU~\citep{du2024conu} calibrates prediction sets for free-form generation using a self-consistency measure without explicitly separating the two uncertainty components. CAELUM combines the spectral decomposition with an adaptation of the two-parameter calibration approach of \citet{azizi2026clear}, makes the matchability ceiling explicit, and derives a Mondrian coverage bound for a learned matchability stratum (see \cref{app:related} for further discussion). CAELUM and its aleatoric-only baseline have similar average set sizes (\cref{tab:multiseed-apss}).

\paragraph{What CAELUM does.}
For a given query $x$, we generate $n$ clarifications and $m$ LLM responses for each clarification. We perform semantic clustering over the $nm$ responses and calculate three uncertainty scores using a kernel applied to response embeddings: the pooled uncertainty $\Utot(x)$, the mean per-clarification uncertainty $\Uepi(x)$, and the empirical residual $\Uale(x) = \max(\Utot(x) - \Uepi(x),0)$. \Cref{fig:overview} shows the pipeline using cluster Shannon entropy. We construct a conformal prediction set $\Cset_\lambda(x)$ using an adaptive prediction set (APS) score divided by the weight
\begin{equation}\label{eq:weight_intro}
    \wlam(x) \;=\; 1 \,+\, \Uale(x) \,+\, \lambda \cdot \Uepi(x), \qquad \lambda \geq 0,
\end{equation}
tuning $\lambdastar$ on the selection split $\Vone$ to minimize average prediction-set size (APSS) over a finite grid and calibrating the conformal threshold $\tauhat$ on a disjoint calibration split $\Vtwo$. The per-instance score $R(x) = \log(\lambdastar \Uepi(x) + \epsilon) - \log(\Uale(x) + \epsilon)$, with $\epsilon=10^{-6}$, summarizes the relative contributions of weighted epistemic and aleatoric uncertainty for individual queries.

\noindent \textbf{The contributions} are as follows:
\begin{enumerate}
    \item We identify the \emph{matchability ceiling}, an upper bound on the coverage of prediction sets restricted to sampled clusters. A coverage bound for a learned matchability stratum makes the false matchability penalty explicit (\cref{thm:stratified}).
    \item We construct semantic prediction sets using clarification ensembles, kernel uncertainty decomposition, and uncertainty-weighted APS scores. Separate splits are used for both selecting the epistemic uncertainty weight, $\lambda^\star$, to minimize average validation set size over a finite grid and setting the conformal threshold.
    \item In our main experiments, we evaluate support prediction, set efficiency, and uncertainty ranking for three generators and six datasets. A finite-grid diagnostic tests whether the set-size objective can distinguish between different weights. Experiments with larger generators test whether the support limitation also holds for larger generators.
\end{enumerate}

\section{Method}\label{sec:method}

\subsection{Problem setup}\label{sec:setup}

Given a query $x$ in the query space $\mathcal X$, a black-box LLM $\Mmodel$ from which we can sample responses, and a calibration dataset of queries and reference answers $\Dcal = \{(x_i, y_i^*)\}_{i=1}^N$, we seek to construct a \textbf{semantic prediction set} $\Cset(x)$ over semantic clusters of answers. The ideal target for conformal inference is marginal coverage
\begin{equation}\label{eq:coverage}
    \mathbb{P}\bigl(y^*_{N+1} \in \Cset(x_{N+1})\bigr) \geq 1 - \alpha,
\end{equation}
where $y^* \in \Cset(x)$ means that the reference cluster selected by a fixed semantic matching rule is included in $\Cset(x)$. This rule is applied to the full collection of sampled clusters before selecting the prediction set. The parameter $\alpha \in (0,1)$ specifies the desired error rate. The probability averages over the calibration data, the test example, and the sampling randomness. In our setting, $\Cset(x)$ is restricted to clusters of sampled answers from the model. Let $M(x,y^*)=1$ denote \emph{matchability}, meaning that the same rule selects a sampled cluster for the reference $y^*$. If $M=0$, no threshold over the sampled clusters can include the reference $y^*$. We distinguish support-conditional coverage,
\begin{equation}\label{eq:support_coverage}
    \mathbb{P}\bigl(y^*_{N+1} \in \Cset(x_{N+1}) \mid M(x_{N+1}, y^*_{N+1})=1\bigr) \geq 1-\alpha,
\end{equation}
which holds for the support-restricted calibration procedure under the assumptions of \cref{prop:coverage}, including $\mathbb{P}(M=1)>0$, from population coverage, which is bounded above by $\pi := \mathbb{P}(M=1)$.

\subsection{Sampling and semantic clustering}\label{sec:clustering}

We use the two-stage sampling approach described by ICE~\citep{hou2023decomposing}. We sample $n$ clarifications $W_1, \ldots, W_n$ from the LLM for $x$. For factual questions, clarifications are alternative ways of stating $x$. For ambiguous questions, clarifications are intended to disambiguate $x$. We sample $m$ responses $y_{ij} \sim \Mmodel(\cdot \mid W_i)$ for each clarification. We use a greedy embedding algorithm to group the $nm$ sampled responses into approximate semantic clusters $\{G_1(x), \ldots, G_K(x)\}$. The first unassigned response in generation order becomes a cluster seed. All remaining unassigned responses whose cosine similarity to that seed is at least $\delta$ join its cluster. We repeat this process until every response has been assigned. The clusters are ordered by decreasing proportion $p_k(x) = |G_k(x)|/(nm)$, and the most frequent response in each cluster is selected as its representative. Other methods, including NLI-based bidirectional entailment~\citep{kuhn2023semantic}, can be used in place of our clustering approach.

\subsection{Uncertainty decomposition}\label{sec:decomposition}

We adapt the ICE decomposition~\citep{hou2023decomposing} to semantic clusters and report results using the kernel-based von Neumann entropy (VNE) formulation of \citet{spectraluncertainty2025} (\cref{app:vne}). We present the cluster-Shannon formulation below for clarity. In this formulation, \emph{total uncertainty} is the entropy of the pooled cluster proportions $p_k^{\mathrm{pool}}(x)=p_k(x)$, $\Utot(x) = -\sum_{k} p_k^{\mathrm{pool}}(x) \log p_k^{\mathrm{pool}}(x)$. \emph{Epistemic uncertainty} is the mean entropy within clarifications, using the same pooled cluster assignments. We restrict these assignments to the $m$ answers of clarification $W_i$ to obtain $p_k^{(i)}(x)$, and then compute
\begin{equation}\label{eq:epistemic}
    \Uepi(x) = \tfrac{1}{n}\sum_{i=1}^{n} H_i(x), \qquad H_i(x) = -\sum_{k} p_k^{(i)}(x) \log p_k^{(i)}(x).
\end{equation}
This quantity measures response variability within a clarification. A generated clarification may not fully resolve the ambiguity in the original question. In this Shannon formulation, $\Uepi(x)=0$ when all responses within each clarification belong to a single cluster, even if the cluster differs across clarifications. \emph{Aleatoric uncertainty} is the residual $\Uale(x) = \Utot(x) - \Uepi(x)$. With the same number of responses per clarification, the aleatoric residual is nonnegative in the shared-cluster Shannon formulation due to the concavity of entropy.

In the VNE formulation, the same encoder and RBF kernel are used for the pooled responses and the responses within each clarification. The entropies are not divided by the logarithm of the sample count. With equal numbers of responses per clarification, entropy concavity also guarantees a nonnegative residual in the shared kernel feature space. We use $\Uale(x) = \max(\Utot(x)-\Uepi(x),0)$ to guard against numerical error (\cref{app:vne}). The Shannon formulation follows the identity $H(Y\mid X) = H(Y\mid X,C) + \mathcal{I}(Y;C\mid X)$, with clarifications $C$ as the conditioning variable. Both formulations require only sampling access to the LLM.

\subsection{Conformal semantic prediction sets}\label{sec:conformal}

\paragraph{Nonconformity score.} For a query $x$ with clusters ordered by mass $p_1(x) \geq \cdots \geq p_K(x)$, let $k^*$ be the index of the cluster selected by the fixed reference-matching rule for $y^*$, when a match exists. We use a deterministic APS-style score~\citep{romano2020aps}, given by the cumulative mass preceding that cluster,
\begin{equation}\label{eq:score}
    s(x, y^*) = \sum_{j=1}^{k^* - 1} p_j(x).
\end{equation}
The APS score is $0$ for questions whose selected reference cluster is first in the ordering and can approach $1$ when that cluster occurs late in the ordering. If none of the clusters matches the reference, we set $M=0$, assign $s(x,y^*)=1$, and mark the question as uncoverable from any subset of sampled clusters. We use both deterministic and randomized APS scores. The randomized variant is $s_{\mathrm{rand}}(x,y^*) = \sum_{j<k^*}p_j(x) + U\,p_{k^*}(x)$, where an independent $U\sim\mathrm{Unif}(0,1)$ is drawn for each item and held fixed across candidate values of $\lambda$. Unmatched items have score $1$ under both schemes. The prediction sets in \cref{eq:prediction_set} use no additional random draw. For matchable items, $s_{\mathrm{rand}}\geq s$ means that $s_{\mathrm{rand}}(x,y^*)/\wlam(x)\leq\hat\tau$ is sufficient for reference inclusion under the cumulative-threshold rule. Deterministic APS scoring is used for the main table and selective-answering results, while seeded randomized APS scoring is used for the CAELUM multi-seed results and the stratified analysis.

\paragraph{Uncertainty-weighted threshold.} The per-instance weight
\begin{equation}\label{eq:weight}
    \wlam(x) = 1 + \Uale(x) + \lambda \cdot \Uepi(x), \qquad \lambda \geq 0,
\end{equation}
scales the conformal threshold. Since cluster proportions sum to one, we cap the effective threshold at one,
\[
    \eta_\lambda(x;\tau)=\min\{\tau\cdot\wlam(x),\,1\}.
\]
We include cluster $G_j(x)$ if the cumulative mass of all preceding clusters is at most this threshold,
\begin{equation}\label{eq:prediction_set}
    \Cset_\lambda(x) = \bigl\{G_j(x): \sum_{\ell<j} p_\ell(x) \leq \eta_\lambda(x;\tauhat)\bigr\}.
\end{equation}
For a fixed threshold $\tau$, increasing $\lambda$ can only expand a query's prediction set. Smaller $\lambda$ reduces the contribution of $\Uepi$ relative to $\Uale$. This monotonicity need not hold when the threshold is recalibrated for each $\lambda$.

\paragraph{Split calibration.} We use three disjoint splits to select $\lambda$ on $\Vone$, calibrate the threshold on $\Vtwo$, and evaluate on test. The encoder, $\delta$, and matching rule may also be tuned on $\Vone$, then fixed for calibration and testing (\cref{app:matcher-audit}). We apply \emph{score normalization},
\begin{equation}\label{eq:normalized_score}
    \snorm_\lambda(x, y^*) = \frac{s(x, y^*)}{\wlam(x)},
\end{equation}
so the conformal quantile $\tauhat_\lambda$ depends on $\lambda$. For randomized calibration, we replace $s$ by $s_{\mathrm{rand}}$ in this normalization and use the resulting scores in both calibration steps. Without normalization, the scores and calibrated threshold would be independent of $\lambda$. Increasing $\lambda$ would then only expand the sets, making $\lambda=0$ an APSS minimizer, possibly tied with other grid values (\cref{rem:weight_role}).

The quantile $Q_q$ is the upward empirical quantile of all calibration scores, including zeros. For a calibration subset of size $L$, the theoretical construction uses the score at rank $r=\lceil(1-\alpha)(L+1)\rceil$. If there are fewer than $r$ calibration items, including an empty calibration set, we set the threshold to $+\infty$ and include all sampled clusters. The quantile convention and this fallback are given in \cref{app:protocol}.

\textbf{Step 1: $\lambda$ selection on $\Vone$.} For each $\lambda$ in the grid $\Lambda$, compute $\snorm_{\lambda,i} = s_i / \wlam(x_i)$ on $\Vone$, set $\tauhat_\lambda = Q_{\lceil(1-\alpha)(|\Vone|+1)\rceil/|\Vone|}(\{\snorm_{\lambda,i}\}_{i \in \Vone})$, and choose
\begin{equation}\label{eq:lambda_selection}
    \lambdastar = \arg\min_{\lambda \in \Lambda} \;\frac{1}{|\Vone|} \sum_{i \in \Vone} |\Cset_\lambda(x_i)|.
\end{equation}
For this selection step, $\Cset_\lambda$ uses the provisional threshold $\tauhat_\lambda$. Ties are resolved by choosing the smallest minimizing weight.

\textbf{Step 2: Conformal threshold on $\Vtwo$.} With $\lambdastar$ fixed, compute
\begin{equation}\label{eq:tau}
    \tauhat = Q_{\lceil(1-\alpha)(|\Vtwo|+1)\rceil / |\Vtwo|}\bigl(\{\snorm_{\lambdastar,i}\}_{i \in \Vtwo}\bigr).
\end{equation}
The final prediction set for a new query $x$ is $\Cset_{\lambdastar}(x)$, as defined in \cref{eq:prediction_set}, with $\tauhat$ and $\lambdastar$. For deterministic scoring, $\snorm\leq\tauhat$ is equivalent to $s\leq\tauhat\,\wlam(x)$ since $\wlam(x)>0$. If $\tauhat\,\wlam(x)\geq1$, the prediction set includes all sampled clusters. For smaller values, we include clusters whose preceding cumulative mass is at most $\tauhat\,\wlam(x)$. In \cref{tab:main}, we calibrate the threshold using all of $\Vtwo$. A reference missing from the sampled clusters cannot be covered, so population coverage cannot exceed the matchability rate. The result below instead calibrates the threshold using only matched examples, $\Vtwo^M=\{i\in\Vtwo:M_i=1\}$.

\paragraph{When can the weight be identified?}
Write $b_{ij}=\sum_{\ell<j}p_\ell(x_i)$. On a finite grid $\Lambda$, the prediction-set size for $x_i$ can change only if a cluster boundary satisfies
\[
    \min_{\lambda\in\Lambda}\eta_\lambda(x_i;\hat\tau_\lambda)
    < b_{ij} \leq
    \max_{\lambda\in\Lambda}\eta_\lambda(x_i;\hat\tau_\lambda).
\]
Hence,
\[
    f_{\rm move}(\Vone)=|\Vone|^{-1}\sum_{i\in\Vone}
    \mathbf{1}\!\left\{
        \min_{\lambda\in\Lambda}|\Cset_\lambda(x_i)|
        <
        \max_{\lambda\in\Lambda}|\Cset_\lambda(x_i)|
    \right\}
\]
measures the fraction of validation items whose set sizes differ across grid points. If every validation item's set size stays the same across grid weights, APSS is constant. Changes in individual set sizes can also cancel in the average, so several weights may minimize APSS. We report the objective range and the fraction of grid weights attaining the minimum in \cref{tab:weight-stability}. The separate held-out movement audit and the validation exceptions are in \cref{app:stability}.

\begin{proposition}[Split coverage on generated support]\label{prop:coverage}
Assume $\mathbb{P}(M=1)>0$. Conditional on $\Vone$, assume exchangeability of the calibration examples and test example within the event $M=1$, including their sampled generations and any score randomization. Select $\lambdastar$ using only $\Vone$ and compute the conformal threshold from $\Vtwo^M=\{i\in\Vtwo:M_i=1\}$ at rank $r_M=\lceil(1-\alpha)(|\Vtwo^M|+1)\rceil$, using $+\infty$ if $r_M>|\Vtwo^M|$. The normalized scoring rule is therefore fixed when computing this threshold. By the standard split conformal argument~\citep{vovk2005,angelopoulos2024},
\[
    \mathbb{P}\bigl(
        y_{N+1}^* \in \Cset_{\lambdastar}(x_{N+1})
        \mid M(x_{N+1},y^*_{N+1})=1
    \bigr) \geq 1-\alpha.
\]
\end{proposition}
\noindent The calibration procedure adapts the two-split construction of CLEAR introduced by \citet[Definition~B.1 and Lemma~B.2]{azizi2026clear}, using semantic clusters as the output space. As the sampled support can omit the reference answer, \cref{prop:coverage} guarantees coverage conditional on matchability rather than unconditional coverage at the nominal level. See \cref{app:theory} for details.

\subsection{Stratified coverage via learned matchability}\label{sec:stratified}

\cref{prop:coverage} is conditional on generated support. On hard cells, matchability caps the population coverage at $\pi=\mathbb{P}(M=1)$ regardless of $\alpha$, so a cell with $\pi=0.4$ makes a $0.9$ nominal target unattainable. We make this dependence explicit via Mondrian conformal prediction~\citep{vovk2003mondrian,bostrom2020mondrian}, which restricts calibration to a stratum where the ceiling can be estimated and audited. Train a binary classifier $\hat m: \mathcal{X} \to \{0, 1\}$ on a held-out split $\Vzero$ (disjoint from $\Vone, \Vtwo$ and the test point) using only pre-generation features of $x$. Define the predicted-matchable subsample $\mathcal{S} = \{i \in \Vtwo : \hat m(x_i) = 1\}$ and let $r=\lceil(1-\alpha)(|\mathcal{S}|+1)\rceil$. For nonempty $\mathcal{S}$ with $r\leq|\mathcal{S}|$, the stratified threshold is
\begin{equation}\label{eq:tau_strat}
    \tauhat_{\mathrm{strat}}(\lambda) = Q_{\lceil(1-\alpha)(|\mathcal{S}|+1)\rceil/|\mathcal{S}|}\bigl(\{\snorm_{\lambda,i}\}_{i \in \mathcal{S}}\bigr).
\end{equation}
If $\mathcal{S}$ is empty or $r>|\mathcal{S}|$, use threshold $+\infty$ and return all sampled clusters. The resulting set $\Cset_{\lambda_0}^{\mathrm{strat}}$, for any $\lambda_0$ fixed or selected using $\Vone$, then satisfies the following two-sided guarantee.

\begin{theorem}[Matchability-stratified coverage with false matchability penalty]\label{thm:stratified}
Let $\hat m:\mathcal{X}\to\{0,1\}$ be any binary classifier trained on $\Vzero$ (disjoint from $\Vtwo$ and the test point), let $\lambda_0$ be fixed or selected using $\Vone$ (also disjoint from $\Vtwo$ and the test point), and let $\Cset_{\lambda_0}^{\mathrm{strat}}(x)$ use the stratified threshold $\tauhat_{\mathrm{strat}}(\lambda_0)$ from \cref{eq:tau_strat}, with the $+\infty$ convention above. Condition on $\Vzero$, $\Vone$, and any randomness used to fit $\hat m$ or select $\lambda_0$. Assume that the calibration and test examples, including their generated samples and any score randomization, are exchangeable under this conditioning. All probabilities below, including $\rho_{\hat m}$, are understood under the same conditioning. Assume $\mathbb{P}(\hat m=1)>0$. Then
\begin{equation}\label{eq:cond_cov}
    1 - \alpha - \rho_{\hat m} \;\leq\; \mathbb{P}\bigl(y_{N+1}^* \in \Cset_{\lambda_0}^{\mathrm{strat}}(x_{N+1}) \,\big|\, \hat m(x_{N+1}) = 1\bigr) \;\leq\; 1 - \rho_{\hat m},
\end{equation}
where $\rho_{\hat m}=\mathbb{P}(M=0\mid \hat m=1)$ is the false matchability rate in the predicted-matchable stratum. Without conditioning on membership in the predicted-matchable stratum, coverage satisfies
\begin{equation}\label{eq:marginal_lb}
    \mathbb{P}\bigl(y_{N+1}^* \in \Cset_{\lambda_0}^{\mathrm{strat}}(x_{N+1})\bigr) \geq (1-\alpha)\,\mathbb{P}\bigl(\hat m(x_{N+1})=1\bigr) - \mathbb{P}\bigl(M=0,\,\hat m(x_{N+1})=1\bigr).
\end{equation}
\end{theorem}
Our bound complements two related formulations. \citet{li2026setvalued} concurrently study feasibility-aware coverage at the population level. \citet{noorani2025beyond} estimate unseen probability mass after sampling and combine set construction with an adaptive query policy. Our taxonomy is instead a per-query prediction learned before generation, and $\rho_{\hat m}$ measures its errors within the selected population. Unseen mass for one sampled distribution need not identify whether that query's reference is absent (\cref{tab:missing-mass}).

If $\rho_{\hat m}=0$, the bound gives nominal coverage within the predicted-matchable stratum, which may still exclude some matchable queries. If $\rho_{\hat m}>\alpha$, the upper bound $1-\rho_{\hat m}$ is below the nominal target $1-\alpha$, so no threshold over sampled clusters can attain that target within the selected stratum. Using the learned matchability stratum as the Mondrian taxonomy expresses the support ceiling through a measurable parameter. We use $\lambda_0=0$ (the aleatoric-only baseline labeled \emph{Vanilla CP}) for the empirical evaluation in \cref{sec:experiments}. \Cref{app:matchable-with-lambdastar} describes the extension to $\lambda_0=\lambdastar$, and the proof and corollary on the upper bound are in \cref{app:stratified-proof}.

\subsection{Diagnostics}\label{sec:diagnostics}
The fitted $\lambda^\star$ controls how much within-clarification variability enters the score. An interpretation can be made only when the set-size objective distinguishes weights and the selected region is stable across splits. On a flat objective, neither a large coefficient nor a tie-breaking value of zero identifies the source of uncertainty. The per-query ratio
\begin{equation}\label{eq:diagnostic}
 R(x)=\log(\lambda^\star\Uepi(x)+\epsilon)-\log(\Uale(x)+\epsilon),\qquad\epsilon=10^{-6},
\end{equation}
describes the relative weighted components. It is also affected by instability in the coefficient and loses its epistemic contribution when $\lambda^\star=0$. We therefore evaluate $\Uepi$ directly for support failure ranking and treat $R$ as a descriptive statistic. Clarification, retrieval, and abstention are possible downstream interventions, but their benefits must be evaluated separately and cannot be inferred from the sign of $R$.

\section{Experiments}\label{sec:experiments}

\subsection{Experimental setup}

\paragraph{Datasets.} Six QA benchmarks, $2{,}104$ questions total. \emph{Factual:} TriviaQA (300)~\citep{joshi2017triviaqa}, OpenNQ (300)~\citep{kwiatkowski2019nq}, SimpleQA (500)~\citep{wei2024simpleqa}, PopQA (500)~\citep{mallen2023popqa}. \emph{Ambiguity benchmarks:} AmbigQA (200)~\citep{min2020ambigqa}, AmbigInst (304)~\citep{hou2023decomposing}. We use the curated subsets of \citet{spectraluncertainty2025} for TriviaQA, OpenNQ, AmbigQA and AmbigInst for direct comparability. The main table uses a 40\% $\Vone$ / 30\% $\Vtwo$ / 30\% Test split. Multi-seed summaries use four random splits, with seed values listed in \cref{app:implementation}. Stratified matchability experiments use a separate $\Vzero/\Vtwo/\mathrm{Test}$ split. Full dataset descriptions are in \cref{app:datasets}.

\paragraph{Models, baselines, metrics.} The main open-weight 4--9B LLMs are \textbf{Qwen3-8B}~\citep{qwen3}, \textbf{Ministral-3-8B-Instruct-2512}~\citep{ministral3} (FP8), and \textbf{Gemma-4-E4B-it}~\citep{gemma4}. Each model self-generates $n=5$ clarifications at temperature $0.7$ and $m=10$ answers per clarification at temperature $0.5$. Answers are clustered with all-mpnet-base-v2 at $\delta = 0.7$ (\cref{app:embedding} examines encoder sensitivity). The reference is matched to cluster representatives, and to raw cluster members for lexical tiers, by a multi-tier criterion combining substring, ROUGE-L, Jaccard, and embedding-cosine tests (\cref{app:implementation}). The six set rules are \textbf{CAELUM} ($\lambda = \lambdastar$, ours), \textbf{Vanilla CP} (the aleatoric-only weight $w=1+\Uale$, i.e., $\lambda = 0$), \textbf{Equal-weight} ($\lambda = 1$), \textbf{Epistemic-only} ($\lambda = 10^6$), a \textbf{self-consistency proxy} (SC proxy), and \textbf{Semantic Entropy CP} (our APS baseline weighted by $1+\Utot$). The SC proxy calibrates $1-\max_k p_k$ and uses the resulting cumulative-mass cutoff. It is inspired by ConU~\citep{du2024conu}, but does not reproduce ConU's reference-aligned candidate scores or set rule. \cref{tab:multiseed-apss} additionally reports \textbf{TrueAPS}, the textbook unweighted APS baseline $w(x)\equiv 1$, since the ``Vanilla'' label is still uncertainty-weighted by $\Uale$. The main table uses deterministic APS. In multi-seed comparisons, CAELUM, Vanilla, Equal-weight, Epistemic-only, and TrueAPS use randomized calibration scores, while the SE-CP baseline uses deterministic APS. Prediction sets remain deterministic. SE-CP and Equal-weight have the same weight whenever $\Utot=\Uale+\Uepi$, and their different performance across seeds cannot be attributed to the decomposition alone. Hallucination ranking compares the four signals $\Utot$ (the spectral-entropy baseline of \citet{spectraluncertainty2025}), $\Uepi$, $\Uale$, and $R(x)$. We report marginal coverage (target $1-\alpha = 0.9$), APSS, AUROC, and $\lambdastar$. The $\lambda$ grid is specified in \cref{app:implementation}. Additional analyses include Qwen3-14B and Qwen3-32B~\citep{qwen3_32b}.

For \cref{tab:main}, both CAELUM and the SC proxy calibrate their thresholds on the full $\Vtwo$ split and use the same unfiltered test questions, including unmatchable items. This differs from the support condition in ConU's correctness-coverage formulation (\cref{app:conu-support}).

\subsection{Results}

\paragraph{Support limits coverage. Small gains in set size may be possible.}
\Cref{tab:main} summarizes experimental results. CAELUM prediction sets are smaller on average than Vanilla prediction sets in 9 of 18 cells, the same average size in 7 of 18 cells, and only slightly larger by no more than $0.01$ in 2 of 18 cells. The largest reductions are $4.5\%$ for Ministral--AmbigQA and about $3.3\%$ for Qwen3--PopQA and Ministral--TriviaQA. The mean differences are smaller than variation across random splits, according to \cref{tab:multiseed-apss}. SC proxy prediction sets have coverage that is up to $6.5$ percentage points lower than CAELUM prediction sets. Comparisons of set sizes should consider differences in coverage. Analysis of post-hoc set-size comparisons in \cref{tab:ncapss} shows no clear winners and identifies some infeasible coverage levels.

\begin{figure}[t]
\centering
\includegraphics[width=\textwidth]{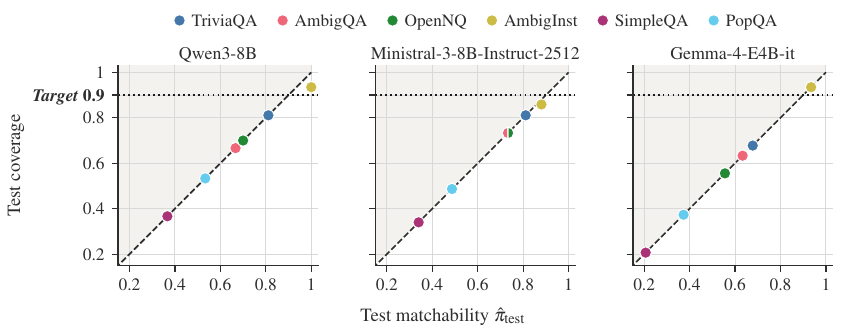}
\caption{Test coverage and matchability for 18 main model--dataset cells, one panel per LLM. Colors identify datasets. The shaded region above the diagonal is unreachable using sampled clusters. A two-color circle marks overlapping datasets. The horizontal line marks the $0.9$ target.}
\label{fig:cov_match}
\end{figure}

Test coverage for CP methods is close to the test support fraction in most cells (\cref{fig:cov_match}). A prediction set can contain all sampled clusters for a given query. Once this happens, weighting cannot add a reference absent from those clusters. Although estimates of $\lambda^\star$ vary greatly by dataset and generator, the small differences in APSS and variability in $\lambda^\star$ across splits suggest that $\lambda^\star$ cannot reliably be interpreted as a measure of intrinsic ambiguity in a dataset.

\begin{table}[t]
\caption{Main results with a target nominal 90\% coverage rate (deterministic APS). Test matchability $\hat\pi_{\rm test}$, the fraction of references matched to sampled clusters, and other test results are evaluated on the same held-out questions. \textbf{Bold} and \underline{underline} designate the smallest and second-smallest displayed set sizes among methods with the highest test coverage in that row, including ties, and $^\dagger$ marks sizes below that coverage. Equal/SE refers to numerically identical results from Equal-weight and SE-CP methods. SC proxy is a simple self-consistency heuristic. The fitted $\lambda^\star$ (deterministic seed SD) does not by itself identify an uncertainty regime. Full comparison results are reported in Table~\ref{tab:complete-methods}.}
\label{tab:main}
\centering
\footnotesize
\setlength{\tabcolsep}{3.5pt}
\renewcommand{\arraystretch}{1.1}
\begingroup
\small
\setlength{\tabcolsep}{3.2pt}
\renewcommand{\arraystretch}{1.12}
\begin{tabular*}{\linewidth}{@{\extracolsep{\fill}}lccccccccc@{}}
\toprule
& \multicolumn{3}{c}{Support / CAELUM} & \multicolumn{4}{c}{Average set size $\downarrow$} & \multicolumn{2}{c}{SC proxy} \\
\cmidrule(lr){2-4}\cmidrule(lr){5-8}\cmidrule(lr){9-10}
Dataset & $\hat\pi_{\rm test}$ & Cov. & $\lambda^\star$ & CAELUM & Vanilla & Equal/SE & Epi.-only & Cov. & Size \\
\midrule
\multicolumn{10}{@{}l}{\textbf{Qwen3-8B}}\\[3pt]
TriviaQA & 0.811 & 0.811 & 47.42 (14.5) & \underline{1.82} & 1.84 & 1.84 & \textbf{1.81} & 0.778 & 1.02$^\dagger$ \\
AmbigQA & 0.667 & 0.667 & 7.54 (33.9) & \textbf{1.30} & \textbf{1.30} & \textbf{1.30} & \textbf{1.30} & 0.650 & 1.00$^\dagger$ \\
OpenNQ & 0.700 & 0.700 & 0.00 (0.0) & \underline{1.20} & \underline{1.20} & \underline{1.20} & \underline{1.20} & 0.700 & \textbf{1.00} \\
AmbigInst & 1.000 & 0.935 & 0.00 (0.0) & \textbf{1.00} & \textbf{1.00} & \textbf{1.00} & \textbf{1.00} & 0.935 & \underline{1.05} \\
SimpleQA & 0.367 & 0.367 & 0.00 (12.2) & \underline{1.77} & \underline{1.77} & \underline{1.77} & \textbf{1.75} & 0.327 & 1.02$^\dagger$ \\
PopQA & 0.533 & 0.533 & 3.15 (38.7) & \textbf{1.74} & 1.80 & 1.79 & \underline{1.75} & 0.520 & 1.02$^\dagger$ \\
\midrule[0.3pt]
\addlinespace[2pt]
\multicolumn{10}{@{}l}{\textbf{Ministral-3-8B-Instruct-2512}}\\[3pt]
TriviaQA & 0.811 & 0.811 & 26.96 (25.5) & \textbf{2.27} & 2.34 & 2.37 & \underline{2.28} & 0.767 & 1.07$^\dagger$ \\
AmbigQA & 0.733 & 0.733 & 18.41 (24.5) & \textbf{3.90} & 4.08$^\dagger$ & 4.05$^\dagger$ & \underline{3.92} & 0.717 & 1.90$^\dagger$ \\
OpenNQ & 0.733 & 0.733 & 6.19 (14.8) & \textbf{1.61} & \underline{1.63} & \underline{1.63} & 1.60$^\dagger$ & 0.711 & 1.00$^\dagger$ \\
AmbigInst & 0.880 & 0.859 & 0.00 (25.5) & \textbf{2.32} & \textbf{2.32} & 2.75 & 2.99 & 0.859 & \underline{2.63} \\
SimpleQA & 0.340 & 0.340 & 0.32 (20.9) & 2.31 & \underline{2.30} & 2.31 & \textbf{2.29} & 0.287 & 1.00$^\dagger$ \\
PopQA & 0.487 & 0.487 & 1.43 (22.5) & \textbf{3.57} & \underline{3.62} & \textbf{3.57} & \textbf{3.57} & 0.447 & 1.12$^\dagger$ \\
\midrule[0.3pt]
\addlinespace[2pt]
\multicolumn{10}{@{}l}{\textbf{Gemma-4-E4B-it}}\\[3pt]
TriviaQA & 0.678 & 0.678 & 45.74 (13.6) & \underline{1.74} & 1.77 & 1.77 & \textbf{1.73} & 0.656 & 1.01$^\dagger$ \\
AmbigQA & 0.633 & 0.633 & 9.32 (5.2) & \textbf{2.17} & \underline{2.18} & \textbf{2.17} & \textbf{2.17} & 0.583 & 1.05$^\dagger$ \\
OpenNQ & 0.556 & 0.556 & 0.00 (15.0) & \textbf{1.51} & \textbf{1.51} & \textbf{1.51} & \textbf{1.51} & 0.544 & 1.00$^\dagger$ \\
AmbigInst & 0.935 & 0.935 & 6.17 (18.9) & \underline{1.96} & 1.97$^\dagger$ & 1.99 & \textbf{1.89} & 0.870 & 1.42$^\dagger$ \\
SimpleQA & 0.207 & 0.207 & 0.28 (0.1) & \textbf{1.66} & \textbf{1.66} & \underline{1.67} & \underline{1.67} & 0.180 & 1.00$^\dagger$ \\
PopQA & 0.373 & 0.373 & 8.51 (7.9) & \underline{1.94} & \textbf{1.93} & 1.95 & \textbf{1.93} & 0.347 & 1.02$^\dagger$ \\
\bottomrule
\end{tabular*}
\endgroup

\end{table}

\paragraph{Hallucination ranking.}
\cref{tab:auroc} presents AUROC for support failure classification, with ``ground truth not in any cluster'' as the positive class. AUROC for $\Uepi$ is higher than for $\Utot$ in $11$ out of $17$ non-degenerate cells in the main dataset split, with an average AUROC improvement of $+0.014$ and a maximum improvement of $+0.059$. Averaged over four random splits, AUROC for $\Uepi$ is higher than for $\Utot$ in $15$ out of $17$ non-degenerate cells, with an average improvement of $\bar\Delta_\textsc{epi}=+0.012$. We use $\Uepi$ as an indicator of support failures and $R(x)$ as a descriptive statistic for each query. See \cref{app:auroc_bars} for results of paired tests and other comparisons of AUROC with $R(x)$.

\paragraph{Coverage in the predicted-matchable subpopulation.}
We use \cref{thm:stratified} with $\lambda=0$ (Vanilla-CP score) and a logistic matchability model $\hat m$ trained on $\Vzero$ ($30\%$ of each cell). The inputs to $\hat m$ are question embeddings and surface features (\cref{app:implementation}). The same fixed matchability model $\hat m$ is used for $\Vtwo$ ($30\%$) and Test ($40\%$). Using $\lambda=0$ separates the effect of stratification by matchability from the effect of $\lambda$ weighting in \cref{tab:main}. A $\lambdastar$-weighted variant is described in \cref{app:matchable-with-lambdastar}.

Across $30$ cells (the $18$ cells of \cref{tab:main} plus Qwen3-14B and Qwen3-32B on the same six datasets) and $4$ random seeds ($n=120$ cell$\times$seed evaluations, omitting the duplicate $50$-question AmbigInst cache), estimated prediction-set coverage in the learned matchability strata is $0.74 \pm 0.15$, and overall estimated coverage is $0.65 \pm 0.17$ ($78$ rows from $27$ cells, learned classifier $\Vtwo$ AUC $0.67 \pm 0.08$). Rows with only one class or $\Vtwo$ AUC less than $0.55$ are labeled oracle-fallback reference results rather than learned-classifier results. Three cells (Qwen3-8B on AmbigInst and SimpleQA, Qwen3-14B on SimpleQA) use the oracle fallback on all four seeds. Including oracle-fallback rows, estimated coverage in matchability strata is $0.83 \pm 0.17$, and overall estimated coverage is $0.65 \pm 0.21$. The $\pm$ values are standard deviations across the corresponding cell$\times$seed evaluations.

Mean coverage gains across all rows include $+20$ percentage points for PopQA across the five models and $+12$ percentage points for TriviaQA. These summaries include oracle fallbacks. The largest learned gains occur for PopQA data. Despite these gains, $19$ of the $20$ model and seed evaluations on PopQA fail the empirical coverage check, which uses a heuristic tolerance. This check is distinct from the coverage bound in Theorem~\ref{thm:stratified} using $\rho_{\hat m}$. Neither more samples nor escalation to larger models closes this apparently knowledge-bound gap. Other information about operating points and deployment challenges is available in \cref{app:operating}. Point estimates for individual cells are available in \cref{fig:stratified} (\cref{app:stratified-figure}) and the supplement.

\paragraph{Cross-LLM transfer of matchability.}
To assess whether $\hat m$ reflects question difficulty shared across models or issues specific to the source LLM, we fit logistic regression models predicting match labels for one LLM on $2{,}104$ questions in our six datasets and apply these models to predict match labels for another LLM. \Cref{app:transfer} reports all six cross-generator directions with bootstrap intervals. The average in-sample AUC for these tasks is $0.80$, and the average cross-generator AUC is $0.78$ ($\bar\Delta=-0.02$). Overall, observed label agreement is $0.81$, well above independent-label agreement levels of $0.50$--$0.53$ at the observed marginal match rates of the two LLMs. The $95\%$ bootstrap lower confidence bound on target-label AUC for all six cross-generator directions exceeds $0.74$ (see \cref{tab:transfer-full,app:transfer} for complete confidence intervals). These results suggest some evidence of cross-generator transfer of match labels for these questions. Note that the logistic regression models use the same question records for both fitting and scoring. We are not assessing cross-generator transfer to new questions and topics.

\paragraph{Selective answering.}
Used as a routing signal (rank inputs by score and route the top-$k$ to a simulated oracle), $R(x)$ yields AUC improvements of $+0.040$ to $+0.061$ over random routing on three TriviaQA cells and $+0.052$ AUC improvement on Qwen3--AmbigQA, but no improvement over random routing on some tasks with low values of $\lambdastar$. Here, AUC measures the area under the accuracy curve as the escalation rate varies. $\Uepi$ yields more consistent improvements ($17/18$ tasks, per-task change in AUC of $[-0.005,+0.071]$, mainly in the range $[+0.02,+0.07]$). See \cref{tab:r_protocol} (\cref{app:r_protocol}) for details.

\paragraph{Detection and abstention.}
At a cost of 50 generated responses per query, the learned CAELUM detector adds an abstention label less often than two baseline rules, Good--Turing and CPQ. The average augmented coverage of the CAELUM detector is greater than $0.9$, as detailed in \cref{tab:missing-mass}. The abstention label contributes to augmented coverage if the reference answer does not match any sampled cluster. Test results for 29 model--dataset cells with judged support labels are separate from the pre-generation classifier studied in \cref{thm:stratified}. For more information about experimental procedures and measures of coverage, see \cref{app:missing-mass}.

\begin{table}[htbp]
\caption{Detection and abstention on 29 cells. CAELUM uses the variant in \cref{app:standardized-weight}. Means are over four splits per cell, then over cells. Parentheses count cells with defined precision. \textbf{Bold} and \underline{underline} mark the best and second-best AUROC, abstention rate, and precision. Arrows indicate preferred directions. The augmented coverage target is $0.9$.}
\label{tab:missing-mass}
\centering
\begingroup
\small
\setlength{\tabcolsep}{3.5pt}
\renewcommand{\arraystretch}{1.12}
\begin{tabular*}{\linewidth}{@{\extracolsep{\fill}}lrrrr@{}}
\toprule
Trigger signal / rule & AUROC $\uparrow$ & Abstention rate $\downarrow$ & Abstention precision $\uparrow$ & Aug. coverage \\
\midrule
CAELUM detector (ours) & \textbf{0.780} & \textbf{0.459} & \textbf{0.569} (28) & 0.924 \\
$\Uepi$ trigger~\citep{spectraluncertainty2025} & \underline{0.696} & \underline{0.506} & \underline{0.481} (28) & 0.918 \\
Good--Turing trigger~\citep{noorani2025beyond} & 0.550 & 0.878 & 0.371 (28) & 0.989 \\
CPQ calibration module~\citep{noorani2025beyond} & \textemdash & 0.848 & 0.384 (27) & 0.987 \\
\bottomrule
\end{tabular*}
\endgroup

\end{table}

\section{Discussion}\label{sec:discussion}

\paragraph{Support and calibration have different goals.}
A score can satisfy a conformal rank bound without the prediction set covering references absent from sampled support. Proposition~\ref{prop:coverage} addresses calibration conditional on true support. Theorem~\ref{thm:stratified} demonstrates how false matchability limits coverage in a learned stratum. The difference between overall coverage of $0.65$ and learned-stratum coverage of $0.74$ compares different populations within the same evaluations, not an overall $90\%$ coverage rate. Both classifier precision and selection rate should be reported.

\paragraph{Weight stability and uncertainty ranking.}
Mean cell APSS values for Vanilla and CAELUM are similar across four random seeds, each being about $2.21$. The mean difference in APSS, Vanilla minus CAELUM, is about $-0.005$. Differences between Vanilla and CAELUM are smaller than variation across splits in every cell. Flatness of the APSS grid search objective function can result in large differences in estimated coefficients and small differences in prediction sets. We find that in a 30-cell experiment, an average of $3.6\%$ of test items have different set sizes across the examined grid weights up to $\lambda=100$ (\cref{app:stability}). Results support the movement diagnostic rather than a regime-based interpretation of every $\lambda^\star$ value. $\Uepi$ performs slightly better than $\Utot$ on average in ranking support failures. $R$ is less consistent and does not provide a good substitute for $\Uepi$.

\paragraph{Scale of generator and interventions.}
A comparison using new generations for the same questions at the same sampling budget indicates that pooled heuristic match rates increase with model size from $61.1\%$ for Qwen3-8B to $66.0\%$ for Qwen3-32B. Match rates for Qwen3-32B are $36.6\%$ for SimpleQA and $57.4\%$ for PopQA. Separate 14B and 32B experiments inform transfer and ranking criteria, and coverage tracks $\hat\pi$ within $0.030$ on all 30 judged cells (\cref{app:historical-scale}). There are advantages and disadvantages to using larger generator models. Adding an abstention label to a prediction set allows augmented coverage to be evaluated alongside ordinary semantic coverage. Prediction sets with an abstention option have a different output space. Reports should include ordinary semantic coverage, augmented coverage, and abstention rate to evaluate coverage and abstention together.

\section{Conclusion, Limitations, and Future Work}\label{sec:conclusion}

\paragraph{Conclusion.}
CAELUM applies kernel von Neumann uncertainty quantification and conformal methods to produce semantic prediction sets for open-ended natural language problems. Across four random splits of 30 model--dataset combinations, matchability, or the proportion of questions whose reference matches a sampled cluster, limits achievable coverage. Mathematical results in this paper, such as \cref{thm:stratified}, give bounds on coverage that incorporate a penalty term for false matchability, $\rho_{\hat m}$. Classifiers trained on one generator have a mean transfer AUC of $0.78$ when used on the same questions from a different generator. $\Uepi$ performs better than total entropy on average at ranking support failures.

\paragraph{Limitations and future work.}
The decomposition method measures uncertainty within the clarification ensemble, which need not correspond to ground-truth ambiguity. The association between $\Uale$ and labeled ambiguity is weak. Additional ambiguity audit results show that $\Uale$ achieves an AUROC of $0.53$--$0.63$. CAELUM and Vanilla have similar set sizes across splits, and there can be several values of $\lambda^\star$ that minimize the objective function, so a selected value need not identify a particular uncertainty regime. Improvements in answer generation may increase matchability, while changes in match evaluation may alter its measured value. The judge audit revisits nonmatches identified heuristically but does not validate all positive matches. Learned-stratum coverage on some factoid datasets, including PopQA, remains below the nominal target. Transfer evaluations reuse questions across generators. Generating 50 answers per question is time-consuming. Current work focuses on generating short-answer QA responses with external references. We do not evaluate long-form responses or multi-turn agentic settings. One potential extension is a multimodal approach that uses source disagreement and other information about modalities to create calibration strata, as in modality-aware conformal regression by \citet{azizi2026modality}. Audits within each stratum would be needed to assess whether the reference answer is present in the sampled support. Treating prediction sets with limited support as having unconditional nominal coverage may encourage users to trust generated answers too readily. High refusal rates can prevent users from accessing potentially useful information.

\begin{ack}
This work used computational resources from the DCSR clusters at the University of Lausanne. BegooAI supported the LLM-as-judge evaluations by providing Gemini API keys and covering the associated API usage costs. The author has no additional competing financial interests to disclose.
\end{ack}

\bibliographystyle{unsrtnat}

\bibliography{references}

@inproceedings{abbasiyadkori2024believe,
  author    = {Abbasi Yadkori, Yasin and Kuzborskij, Ilja and Gy{\"o}rgy, Andr{\'a}s and Szepesv{\'a}ri, Csaba},
  title     = {To Believe or Not to Believe Your {LLM}: Iterative Prompting for Estimating Epistemic Uncertainty},
  booktitle = {Advances in Neural Information Processing Systems},
  volume    = {37},
  pages     = {58077--58117},
  year      = {2024},
  doi       = {10.52202/079017-1852},
  url       = {https://proceedings.neurips.cc/paper_files/paper/2024/hash/6aebba00fff5b6de7b488e496f80edd7-Abstract-Conference.html}
}

@inproceedings{ahdritz2025provable,
  author    = {Ahdritz, Gustaf and Gollakota, Aravind and Gopalan, Parikshit and Peale, Charlotte and Wieder, Udi},
  title     = {Provable Uncertainty Decomposition via Higher-Order Calibration},
  booktitle = {International Conference on Learning Representations},
  year      = {2025},
  url       = {https://proceedings.iclr.cc/paper_files/paper/2025/hash/122ea6470232ee5e79a2649243348005-Abstract-Conference.html}
}

@inproceedings{azizi2025semf,
  author = {Azizi, Ilia and Boldi, Marc-Olivier and Chavez-Demoulin, Val{\'e}rie},
  title = {{SEMF}: Supervised Expectation-Maximization Framework for Predicting Intervals},
  booktitle = {Proceedings of the Fourteenth Symposium on Conformal and Probabilistic Prediction with Applications},
  series = {Proceedings of Machine Learning Research},
  volume = {266},
  pages = {250--281},
  year = {2025},
  publisher = {PMLR},
  url = {https://proceedings.mlr.press/v266/azizi25a.html}
}

@inproceedings{azizi2026clear,
  author    = {Azizi, Ilia and Bodik, Juraj and Heiss, Jakob and Yu, Bin},
  title     = {{CLEAR}: Calibrated Learning for Epistemic and Aleatoric Risk},
  booktitle = {The Fourteenth International Conference on Learning Representations},
  year      = {2026},
  url       = {https://proceedings.iclr.cc/paper_files/paper/2026/hash/fff035a88e93fff416f66f8a52dd9067-Abstract-Conference.html}
}

@inproceedings{azizi2026modality,
  author    = {Azizi, Ilia},
  title     = {Conformal Calibration for Multi-Modal Regression with Missing Modalities},
  booktitle = {Proceedings of the Fifteenth Symposium on Conformal and Probabilistic Prediction with Applications},
  series    = {Proceedings of Machine Learning Research},
  volume    = {329},
  pages     = {343--377},
  year      = {2026},
  publisher = {PMLR},
  url       = {https://proceedings.mlr.press/v329/azizi26a.html}
}

@article{barber2020,
  author  = {Barber, Rina Foygel and Cand{\`e}s, Emmanuel J. and Ramdas, Aaditya and Tibshirani, Ryan J.},
  title   = {The Limits of Distribution-Free Conditional Predictive Inference},
  journal = {Information and Inference: A Journal of the IMA},
  volume  = {10},
  number  = {2},
  pages   = {455--482},
  year    = {2021},
  doi     = {10.1093/imaiai/iaaa017},
  note    = {First published online in 2020},
  url = {https://arxiv.org/abs/1903.04684}
}

@inproceedings{bostrom2020mondrian,
  author    = {Bostr{\"o}m, Henrik and Johansson, Ulf},
  title     = {Mondrian Conformal Regressors},
  booktitle = {Proceedings of the Ninth Symposium on Conformal and Probabilistic Prediction and Applications},
  series    = {Proceedings of Machine Learning Research},
  volume    = {128},
  pages     = {114--133},
  year      = {2020},
  publisher = {PMLR},
  url       = {https://proceedings.mlr.press/v128/bostrom20a.html}
}

@inproceedings{cabezas2025epistemic,
  author    = {Cruz Cabezas, Luben Miguel and Silva Santos, Vagner and Ramos, Thiago and Izbicki, Rafael},
  title     = {Epistemic Uncertainty in Conformal Scores: A Unified Approach},
  booktitle = {Proceedings of the Forty-First Conference on Uncertainty in Artificial Intelligence},
  series    = {Proceedings of Machine Learning Research},
  volume    = {286},
  pages     = {443--470},
  year      = {2025},
  publisher = {PMLR},
  url       = {https://proceedings.mlr.press/v286/cruz-cabezas25a.html}
}

@inproceedings{chan2025cip,
  author    = {Chan, Kwan Ho Ryan and Ge, Yuyan and Dobriban, Edgar and Hassani, Hamed and Vidal, Ren{\'e}},
  title     = {Conformal Information Pursuit for Interactively Guiding Large Language Models},
  booktitle = {Advances in Neural Information Processing Systems},
  year      = {2025},
  volume = {38},
  doi = {10.52202/085713-3021},
  url = {https://proceedings.neurips.cc/paper_files/paper/2025/hash/823a3d2cf462fb815978314023e48f65-Abstract-Conference.html}
}

@inproceedings{depeweg2018decomposition,
  author    = {Depeweg, Stefan and Hern{\'a}ndez-Lobato, Jos{\'e} Miguel and Doshi-Velez, Finale and Udluft, Steffen},
  title     = {Decomposition of Uncertainty in {Bayesian} Deep Learning for Efficient and Risk-Sensitive Learning},
  booktitle = {Proceedings of the 35th International Conference on Machine Learning},
  series    = {Proceedings of Machine Learning Research},
  volume    = {80},
  pages     = {1184--1193},
  year      = {2018},
  publisher = {PMLR},
  url       = {https://proceedings.mlr.press/v80/depeweg18a.html}
}

@inproceedings{du2024conu,
  author    = {Wang, Zhiyuan and Duan, Jinhao and Cheng, Lu and Zhang, Yue and Wang, Qingni and Shi, Xiaoshuang and Xu, Kaidi and Shen, Heng Tao and Zhu, Xiaofeng},
  title     = {{ConU}: Conformal Uncertainty in Large Language Models with Correctness Coverage Guarantees},
  booktitle = {Findings of the Association for Computational Linguistics: EMNLP 2024},
  pages     = {6886--6898},
  year      = {2024},
  publisher = {Association for Computational Linguistics},
  doi       = {10.18653/v1/2024.findings-emnlp.404},
  url       = {https://aclanthology.org/2024.findings-emnlp.404/}
}

@article{farquhar2024detecting,
  author  = {Farquhar, Sebastian and Kossen, Jannik and Kuhn, Lorenz and Gal, Yarin},
  title   = {Detecting Hallucinations in Large Language Models Using Semantic Entropy},
  journal = {Nature},
  volume  = {630},
  pages   = {625--630},
  year    = {2024},
  doi     = {10.1038/s41586-024-07421-0},
  url = {https://www.nature.com/articles/s41586-024-07421-0}
}

@misc{gemma4,
  author       = {{Google DeepMind}},
  title        = {{Gemma 4} Model Card: {gemma-4-E4B-it}},
  year         = {2026},
  howpublished = {Hugging Face model card},
  url          = {https://huggingface.co/google/gemma-4-E4B-it}
}

@article{hullermeier2021aleatoric,
  author  = {H{\"u}llermeier, Eyke and Waegeman, Willem},
  title   = {Aleatoric and Epistemic Uncertainty in Machine Learning: An Introduction to Concepts and Methods},
  journal = {Machine Learning},
  volume  = {110},
  number  = {3},
  pages   = {457--506},
  year    = {2021},
  doi     = {10.1007/s10994-021-05946-3},
  url = {https://arxiv.org/abs/1910.09457}
}

@inproceedings{javanmardi2025optimal,
  author  = {Javanmardi, Alireza and Zargarbashi, Soroush H. and Thies, Santo M. A. R. and Waegeman, Willem and Bojchevski, Aleksandar and H{\"u}llermeier, Eyke},
  title   = {Optimal Conformal Prediction under Epistemic Uncertainty},
  booktitle = {Proceedings of the 42nd Conference on Uncertainty in Artificial Intelligence},
  series    = {Proceedings of Machine Learning Research},
  volume    = {337},
  pages     = {2461--2479},
  year      = {2026},
  publisher = {PMLR},
  url       = {https://proceedings.mlr.press/v337/javanmardi26a.html}
}

@inproceedings{joshi2017triviaqa,
  author    = {Joshi, Mandar and Choi, Eunsol and Weld, Daniel S. and Zettlemoyer, Luke},
  title     = {{TriviaQA}: A Large Scale Distantly Supervised Challenge Dataset for Reading Comprehension},
  booktitle = {Proceedings of the 55th Annual Meeting of the Association for Computational Linguistics},
  pages     = {1601--1611},
  year      = {2017},
  publisher = {Association for Computational Linguistics},
  doi       = {10.18653/v1/P17-1147},
  url       = {https://aclanthology.org/P17-1147/}
}

@inproceedings{karim2025beyond,
  author    = {Karim, Ahmed and Wang, Qiao and Yuan, Zheng},
  title     = {Beyond the Score: Uncertainty-Calibrated {LLMs} for Automated Essay Assessment},
  booktitle = {Proceedings of the 2025 Conference on Empirical Methods in Natural Language Processing},
  year      = {2025},
  url       = {https://aclanthology.org/2025.emnlp-main.992/},
  pages = {19631--19636},
  publisher = {Association for Computational Linguistics},
  doi = {10.18653/v1/2025.emnlp-main.992}
}

@inproceedings{kirchhof2025reassessment,
  author    = {Kirchhof, Michael and Kasneci, Gjergji and Kasneci, Enkelejda},
  title     = {Position: Uncertainty Quantification Needs Reassessment for Large Language Model Agents},
  booktitle = {Proceedings of the 42nd International Conference on Machine Learning},
  series    = {Proceedings of Machine Learning Research},
  volume    = {267},
  pages     = {81665--81677},
  year      = {2025},
  publisher = {PMLR},
  url       = {https://proceedings.mlr.press/v267/kirchhof25b.html}
}

@inproceedings{belmadani2026judges,
  author    = {Belmadani, Ikram and El Khettari, Oumaima and Constant dit Beaufils, Pac{\^o}me and Dufour, Richard and Favre, Benoit},
  title     = {Who Judges the Judge? {E}valuating {LLM}-as-a-Judge for {French} Medical Open-Ended {QA}},
  booktitle = {Proceedings of the 1st Workshop on Linguistic Analysis for Health},
  pages     = {142--157},
  publisher = {Association for Computational Linguistics},
  year      = {2026},
  doi       = {10.18653/v1/2026.healing-1.12},
  url       = {https://aclanthology.org/2026.healing-1.12/}
}

@inproceedings{kuhn2023semantic,
  author    = {Kuhn, Lorenz and Gal, Yarin and Farquhar, Sebastian},
  title     = {Semantic Uncertainty: Linguistic Invariances for Uncertainty Estimation in Natural Language Generation},
  booktitle = {International Conference on Learning Representations},
  year      = {2023},
  url       = {https://openreview.net/forum?id=VD-AYtP0dve}
}

@article{kumar2023conformal,
  author  = {Kumar, Bhawesh and Lu, Charlie and Gupta, Gauri and Palepu, Anil and Bellamy, David and Raskar, Ramesh and Beam, Andrew},
  title   = {Conformal Prediction with Large Language Models for Multi-Choice Question Answering},
  journal = {arXiv preprint arXiv:2305.18404},
  year    = {2023},
  url     = {https://arxiv.org/abs/2305.18404}
}

@article{kwiatkowski2019nq,
  author  = {Kwiatkowski, Tom and Palomaki, Jennimaria and Redfield, Olivia and Collins, Michael and Parikh, Ankur and Alberti, Chris and others},
  title   = {{Natural Questions}: A Benchmark for Question Answering Research},
  journal = {Transactions of the Association for Computational Linguistics},
  volume  = {7},
  pages   = {452--466},
  year    = {2019},
  doi     = {10.1162/tacl_a_00276},
  url = {https://aclanthology.org/Q19-1026/}
}

@article{li2026setvalued,
  author  = {Li, Ye and Hu, Anqi and Ye, Yuanchang and Tong, Shiyan and Wang, Zhiyuan and Fu, Bo},
  title   = {Set-Valued Prediction for Large Language Models with Feasibility-Aware Coverage Guarantees},
  journal = {arXiv preprint arXiv:2603.22966},
  year    = {2026},
  url     = {https://arxiv.org/abs/2603.22966}
}

@inproceedings{mallen2023popqa,
  author    = {Mallen, Alex and Asai, Akari and Zhong, Victor and Das, Rajarshi and Khashabi, Daniel and Hajishirzi, Hannaneh},
  title     = {When Not to Trust Language Models: Investigating Effectiveness of Parametric and Non-Parametric Memories},
  booktitle = {Proceedings of the 61st Annual Meeting of the Association for Computational Linguistics},
  pages     = {9802--9822},
  year      = {2023},
  publisher = {Association for Computational Linguistics},
  doi       = {10.18653/v1/2023.acl-long.546},
  url       = {https://aclanthology.org/2023.acl-long.546/}
}

@inproceedings{manakul2023selfcheck,
  author    = {Manakul, Potsawee and Liusie, Adian and Gales, Mark J. F.},
  title     = {{SelfCheckGPT}: Zero-Resource Black-Box Hallucination Detection for Generative Large Language Models},
  booktitle = {Proceedings of the 2023 Conference on Empirical Methods in Natural Language Processing},
  pages     = {9004--9017},
  year      = {2023},
  publisher = {Association for Computational Linguistics},
  doi       = {10.18653/v1/2023.emnlp-main.557},
  url       = {https://aclanthology.org/2023.emnlp-main.557/}
}

@inproceedings{min2020ambigqa,
  author    = {Min, Sewon and Michael, Julian and Hajishirzi, Hannaneh and Zettlemoyer, Luke},
  title     = {{AmbigQA}: Answering Ambiguous Open-Domain Questions},
  booktitle = {Proceedings of the 2020 Conference on Empirical Methods in Natural Language Processing},
  pages     = {5783--5797},
  year      = {2020},
  publisher = {Association for Computational Linguistics},
  doi       = {10.18653/v1/2020.emnlp-main.466},
  url       = {https://aclanthology.org/2020.emnlp-main.466/}
}

@misc{ministral3,
  author       = {{Mistral AI}},
  title        = {{Ministral-3-8B-Instruct-2512} Model Card},
  year         = {2025},
  howpublished = {Hugging Face model card},
  url          = {https://huggingface.co/mistralai/Ministral-3-8B-Instruct-2512}
}

@inproceedings{nikitin2024kernel,
  author    = {Nikitin, Alexander and Kossen, Jannik and Gal, Yarin and Marttinen, Pekka},
  title     = {Kernel Language Entropy: Fine-Grained Uncertainty Quantification for {LLMs} from Semantic Similarities},
  booktitle = {Advances in Neural Information Processing Systems},
  volume    = {37},
  pages     = {8901--8929},
  year      = {2024},
  doi = {10.52202/079017-0283},
  url = {https://proceedings.neurips.cc/paper_files/paper/2024/hash/10c456d2160517581a234dfde15a7505-Abstract-Conference.html}
}

@inproceedings{noorani2025beyond,
  author  = {Noorani, Sima and Kiyani, Shayan and Pappas, George J. and Hassani, Hamed},
  title   = {Conformal Prediction Beyond the Seen: A Missing Mass Perspective for Uncertainty Quantification in Generative Models},
  booktitle = {Advances in Neural Information Processing Systems},
  volume    = {38},
  publisher = {Curran Associates, Inc.},
  year    = {2025},
  doi     = {10.52202/085713-1084},
  url     = {https://proceedings.neurips.cc/paper_files/paper/2025/hash/2e5911cb61db978c225cf62b6c029192-Abstract-Conference.html}
}

@article{openai2023gpt4,
  author  = {{OpenAI}},
  title   = {{GPT-4} Technical Report},
  journal = {arXiv preprint arXiv:2303.08774},
  year    = {2023},
  url     = {https://arxiv.org/abs/2303.08774}
}

@inproceedings{quach2024conformal,
  author    = {Quach, Victor and Fisch, Adam and Schuster, Tal and Yala, Adam and Sohn, Jae Ho and Jaakkola, Tommi S. and Barzilay, Regina},
  title     = {Conformal Language Modeling},
  booktitle = {International Conference on Learning Representations},
  year      = {2024},
  url       = {https://proceedings.iclr.cc/paper_files/paper/2024/hash/31421b112e5f7faf4fc577b74e45dab2-Abstract-Conference.html}
}

@misc{qwen3,
  author       = {{Qwen Team}},
  title        = {{Qwen3-8B} Model Card},
  year         = {2025},
  howpublished = {Hugging Face model card},
  url          = {https://huggingface.co/Qwen/Qwen3-8B}
}

@misc{qwen3_32b,
      title={Qwen3 Technical Report}, 
      author={Yang, An and Li, Anfeng and Yang, Baosong and Zhang, Beichen and Hui, Binyuan and Zheng, Bo and others},
      year={2025},
      eprint={2505.09388},
      archivePrefix={arXiv},
      primaryClass={cs.CL},
      url={https://arxiv.org/abs/2505.09388}, 
}

@inproceedings{romano2020aps,
  author    = {Romano, Yaniv and Sesia, Matteo and Cand{\`e}s, Emmanuel},
  title     = {Classification with Valid and Adaptive Coverage},
  booktitle = {Advances in Neural Information Processing Systems},
  volume    = {33},
  pages     = {3581--3591},
  year      = {2020},
  url = {https://proceedings.neurips.cc/paper_files/paper/2020/hash/244edd7e85dc81602b7615cd705545f5-Abstract.html}
}

@inproceedings{sale2025aleatoric,
  author    = {Sale, Yusuf and Javanmardi, Alireza and H{\"u}llermeier, Eyke},
  title     = {Aleatoric and Epistemic Uncertainty in Conformal Prediction},
  booktitle = {Proceedings of the Fourteenth Symposium on Conformal and Probabilistic Prediction with Applications},
  series    = {Proceedings of Machine Learning Research},
  volume    = {266},
  pages     = {784--786},
  year      = {2025},
  publisher = {PMLR},
  url       = {https://proceedings.mlr.press/v266/sale25a.html}
}

@article{touvron2023llama,
  author  = {Touvron, Hugo and Lavril, Thibaut and Izacard, Gautier and Martinet, Xavier and Lachaux, Marie-Anne and others},
  title   = {{LLaMA}: Open and Efficient Foundation Language Models},
  journal = {arXiv preprint arXiv:2302.13971},
  year    = {2023},
  url     = {https://arxiv.org/abs/2302.13971}
}

@misc{vovk2003mondrian,
  author      = {Vovk, Vladimir and Lindsay, David and Nouretdinov, Ilia and Gammerman, Alex},
  title       = {Mondrian Confidence Machine},
  year        = {2003},
  howpublished = {On-line Compression Modelling Project, Working Paper No. 4},
  url = {https://alrw.net/old/04.pdf}
}

@book{vovk2005,
  author    = {Vovk, Vladimir and Gammerman, Alex and Shafer, Glenn},
  title     = {Algorithmic Learning in a Random World},
  publisher = {Springer},
  year      = {2005},
  doi       = {10.1007/b106715}
}

@article{wei2024simpleqa,
  author  = {Wei, Jason and Nguyen, Karina and Chung, Hyung Won and Jiao, Yunxin Joy and Papay, Spencer and Glaese, Amelia and Schulman, John and Fedus, William},
  title   = {Measuring Short-Form Factuality in Large Language Models},
  journal = {arXiv preprint arXiv:2411.04368},
  year    = {2024},
  url     = {https://arxiv.org/abs/2411.04368}
}

@inproceedings{wimmer2023quantifying,
  author    = {Wimmer, Lisa and Sale, Yusuf and Hofman, Paul and Bischl, Bernd and H{\"u}llermeier, Eyke},
  title     = {Quantifying Aleatoric and Epistemic Uncertainty in Machine Learning: Are Conditional Entropy and Mutual Information Appropriate Measures?},
  booktitle = {Proceedings of the Thirty-Ninth Conference on Uncertainty in Artificial Intelligence},
  series    = {Proceedings of Machine Learning Research},
  volume    = {216},
  pages     = {2282--2292},
  year      = {2023},
  publisher = {PMLR},
  url       = {https://proceedings.mlr.press/v216/wimmer23a.html}
}

@article{xin2026paraphrase,
  author  = {Xin, Jiayi and Qiang, Evan and Zhu, Zihan and Li, Xiang and Su, Weijie J. and Long, Qi},
  title   = {Conformal Prediction with Paraphrase-Aware Scoring for {LLM} Uncertainty Quantification},
  journal = {arXiv preprint arXiv:2610.04239},
  year    = {2026},
  url     = {https://arxiv.org/abs/2610.04239}
}

@article{angelopoulos2024,
  author  = {Angelopoulos, Anastasios N. and Bates, Stephen},
  title   = {Conformal Prediction: A Gentle Introduction},
  journal = {Foundations and Trends in Machine Learning},
  volume  = {16},
  number  = {4},
  pages   = {494--591},
  year    = {2023},
  doi     = {10.1561/2200000101},
  url = {https://doi.org/10.1561/2200000101}
}

@inproceedings{bickfordsmith2024rethinking,
  author    = {Bickford Smith, Freddie and Kossen, Jannik and Trollope, Eleanor and Van Der Wilk, Mark and Foster, Adam and Rainforth, Tom},
  title     = {Rethinking Aleatoric and Epistemic Uncertainty},
  booktitle = {Proceedings of the 42nd International Conference on Machine Learning},
  series    = {Proceedings of Machine Learning Research},
  volume    = {267},
  pages     = {4345--4359},
  publisher = {PMLR},
  year      = {2025},
  url       = {https://proceedings.mlr.press/v267/bickford-smith25a.html}
}

@inproceedings{hamidieh2026complementing,
  author    = {Hamidieh, Kimia and Thost, Veronika and Gerych, Walter and Yurochkin, Mikhail and Ghassemi, Marzyeh},
  title     = {Complementing Self-Consistency with Cross-Model Disagreement for Uncertainty Quantification},
  booktitle = {The Fourteenth International Conference on Learning Representations},
  year      = {2026},
  url       = {https://proceedings.iclr.cc/paper_files/paper/2026/hash/83e9c3ea5d0b5de331708058637ada10-Abstract-Conference.html}
}

@inproceedings{guo2017calibration,
  author    = {Guo, Chuan and Pleiss, Geoff and Sun, Yu and Weinberger, Kilian Q.},
  title     = {On Calibration of Modern Neural Networks},
  booktitle = {Proceedings of the 34th International Conference on Machine Learning},
  series    = {Proceedings of Machine Learning Research},
  volume    = {70},
  pages     = {1321--1330},
  publisher = {PMLR},
  year      = {2017},
  url       = {https://proceedings.mlr.press/v70/guo17a.html}
}

@inproceedings{hou2023decomposing,
  author    = {Hou, Bairu and Liu, Yujian and Qian, Kaizhi and Andreas, Jacob and Chang, Shiyu and Zhang, Yang},
  title     = {Decomposing Uncertainty for Large Language Models through Input Clarification Ensembling},
  booktitle = {Proceedings of the 41st International Conference on Machine Learning},
  series    = {Proceedings of Machine Learning Research},
  volume    = {235},
  pages     = {19023--19042},
  publisher = {PMLR},
  year      = {2024},
  url       = {https://proceedings.mlr.press/v235/hou24b.html}
}

@inproceedings{hou2024icl,
  author    = {Ling, Chen and Zhao, Xujiang and Zhang, Xuchao and Cheng, Wei and Liu, Yanchi and Sun, Yiyou and Oishi, Mika and Osaki, Takao and Matsuda, Katsushi and Ji, Jie and Bai, Guangji and Zhao, Liang and Chen, Haifeng},
  title     = {Uncertainty Quantification for In-Context Learning of Large Language Models},
  booktitle = {Proceedings of the 2024 Conference of the North American Chapter of the Association for Computational Linguistics: Human Language Technologies},
  pages     = {3357--3370},
  publisher = {Association for Computational Linguistics},
  year      = {2024},
  doi       = {10.18653/v1/2024.naacl-long.184},
  url       = {https://aclanthology.org/2024.naacl-long.184/}
}

@inproceedings{kendall2017uncertainties,
  author    = {Kendall, Alex and Gal, Yarin},
  title     = {What Uncertainties Do We Need in {Bayesian} Deep Learning for Computer Vision?},
  booktitle = {Advances in Neural Information Processing Systems},
  volume    = {30},
  year      = {2017},
  url       = {https://proceedings.neurips.cc/paper_files/paper/2017/hash/2650d6089a6d640c5e85b2b88265dc2b-Abstract.html}
}

@article{lambrecht2026jucal,
  author  = {Heiss, Jakob and Lambrecht, S{\"o}ren and Weissteiner, Jakob and Wutte, Hanna and {\v{Z}}uri{\v{c}}, {\v{Z}}an and Teichmann, Josef and Yu, Bin},
  title   = {{JUCAL}: Jointly Calibrating Aleatoric and Epistemic Uncertainty in Classification Tasks},
  journal = {arXiv preprint arXiv:2602.20153},
  year    = {2026},
  url     = {https://arxiv.org/abs/2602.20153}
}

@inproceedings{spectraluncertainty2025,
  author    = {Walha, Nassim and Gruber, Sebastian G. and Decker, Thomas and Yang, Yinchong and Javanmardi, Alireza and H{\"u}llermeier, Eyke and Buettner, Florian},
  title     = {Fine-Grained Uncertainty Decomposition in Large Language Models: A Spectral Approach},
  booktitle = {Proceedings of the AAAI Conference on Artificial Intelligence},
  volume    = {40},
  pages     = {26090--26098},
  year      = {2026},
  doi       = {10.1609/aaai.v40i31.39811},
  url       = {https://ojs.aaai.org/index.php/AAAI/article/view/39811},
  note      = {Issue 31}
}

@inproceedings{wang2024blob,
  author    = {Wang, Yibin and Shi, Haizhou and Han, Ligong and Metaxas, Dimitris and Wang, Hao},
  title     = {{BLoB}: Bayesian Low-Rank Adaptation by Backpropagation for Large Language Models},
  booktitle = {Advances in Neural Information Processing Systems},
  volume    = {37},
  pages     = {67758--67794},
  year      = {2024},
  doi       = {10.52202/079017-2164},
  url       = {https://proceedings.neurips.cc/paper_files/paper/2024/hash/7d53575463291ea6b5a23cf6e571f59b-Abstract-Conference.html}
}

@inproceedings{yang2024laplaceloRA,
  author    = {Yang, Adam X. and Robeyns, Maxime and Wang, Xi and Aitchison, Laurence},
  title     = {Bayesian Low-Rank Adaptation for Large Language Models},
  booktitle = {International Conference on Learning Representations},
  year      = {2024},
  url       = {https://proceedings.iclr.cc/paper_files/paper/2024/hash/07c256a163a7559186ec1c71e95b9ec9-Abstract-Conference.html}
}
\appendix
\raggedbottom
\let\caelumsection\section
\renewcommand{\section}{\FloatBarrier\Needspace{5\baselineskip}\caelumsection}
\let\caelumsubsection\subsection
\renewcommand{\subsection}{\FloatBarrier\Needspace{4\baselineskip}\caelumsubsection}
\crefalias{section}{appendix}
\crefalias{subsection}{appendix}
\newsavebox{\caelumappendixbox}
\newenvironment{appendixblock}{%
  \par\FloatBarrier\setbox\caelumappendixbox=\vbox\bgroup
  \let\section\caelumsection
  \let\subsection\caelumsubsection
}{%
  \par\egroup
  \Needspace{\dimexpr\ht\caelumappendixbox+\dp\caelumappendixbox\relax}%
  \unvbox\caelumappendixbox
}

\section{Extended related work}\label{app:related}

\paragraph{Uncertainty estimation in LLMs.}
Token-level probabilities fail to capture semantic confidence. Paraphrases of the same answer can have very different likelihoods~\citep{kuhn2023semantic}. Semantic entropy~\citep{kuhn2023semantic} clusters generations via bidirectional entailment and computes entropy over clusters, and \citet{farquhar2024detecting} scale this to 30 model-task combinations, using it to detect confabulations. \citet{nikitin2024kernel} generalize this construction with kernel-based von Neumann entropy on a positive-semidefinite semantic-similarity kernel, and \citet{manakul2023selfcheck} detect hallucinations via cross-response consistency without token probabilities. These methods yield uncertainty scores for ranking, but give no coverage guarantees and no signal on the source of uncertainty.

\paragraph{Uncertainty decomposition for LLMs.}
The classical decomposition $H(Y|X) = \mathbb{E}_\theta[H(Y|X,\theta)] + \mathcal{I}(Y;\theta|X)$~\citep{depeweg2018decomposition,hullermeier2021aleatoric} marginalizes over model parameters, which is infeasible for large LLMs and has been criticized in the parametric setting~\citep{wimmer2023quantifying,bickfordsmith2024rethinking}. \citet{hou2023decomposing} sidestep both issues with Input Clarification Ensembling (ICE), replacing parameter variability with input-context variability so the procedure stays in the output space. \citet{spectraluncertainty2025} extend this with von Neumann entropy in a kernel-induced semantic space, outperforming ICE on ambiguity detection and correctness prediction. Other routes include information-theoretic epistemic estimators via iterative prompting~\citep{abbasiyadkori2024believe}, inter-model disagreement~\citep{hamidieh2026complementing}, mutual-information decomposition for in-context learning~\citep{hou2024icl}, and provable decomposition via higher-order calibration~\citep{ahdritz2025provable}. Weight-space Bayesian methods like Laplace-LoRA and BLoB~\citep{yang2024laplaceloRA,wang2024blob} require fine-tuning and do not apply to frozen LLMs. \citet{kirchhof2025reassessment} argue that uncertainty quantification for interactive LLM agents must account for underspecification, interaction, and how uncertainty is communicated. CAELUM returns prediction sets together with uncertainty diagnostics. Our analysis studies the generated-support limitation when these uncertainty signals are used inside conformal set construction.

\paragraph{Conformal prediction for language models.}
Conformal prediction~\citep{vovk2005,angelopoulos2024} gives distribution-free coverage under exchangeability. \citet{quach2024conformal} calibrate a sampling stopping rule and a rejection rule for sets of sampled open-ended responses, a sequence-level candidate-set mechanism rather than token-level conformal prediction. \citet{kumar2023conformal} construct prediction sets for multiple-choice QA, and \citet{karim2025beyond} calibrate LLM-based essay scoring. Conformal Information Pursuit~\citep{chan2025cip} uses conformal set sizes to drive interactive question answering, but treats uncertainty as a single signal. \citet{du2024conu} propose ConU, applying conformal prediction to free-form generation via self-consistency without decomposing uncertainty. \citet{sale2025aleatoric} analyze how aleatoric and epistemic uncertainty enter conformal prediction. CAELUM pairs conformal set construction for LLMs with a decomposed uncertainty weight. Concurrent work by \citet{li2026setvalued} constructs set-valued predictions for LLMs with feasibility-aware coverage at the population level. \citet{noorani2025beyond} allow prediction sets to include a catch-all element for unseen outputs, using Good--Turing estimates of the missing probability mass. In our case, in Theorem~\ref{thm:stratified} we condition on a learned stratum of queries predicted to be matchable before generation. We account for false matchability in this stratum through $\rho_{\hat m}$. In our experiments with a fixed sampling budget, the Good--Turing score detected support failure at AUROC $0.550$, while our learned detector operating after generation achieved $0.780$ (see Table~\ref{tab:missing-mass}).

\paragraph{ConU and the self-consistency proxy.}\label{app:conu-support}
ConU~\citep{du2024conu} scores each sampled answer using its semantic frequency and its agreement with other answers. During calibration, it takes the score of a candidate that matches the reference and computes a conformal quantile of these scores. At test time, it includes each candidate whose score falls below or equals that threshold. Its correctness-coverage formulation requires a correct candidate for every calibration query and assumes one exists for every test query. The guarantee therefore relies on the answer already being present among the samples; it does not cover references missing from the sampled support.

Our SC proxy instead takes a quantile of $1-\max_k p_k$ over all calibration questions and uses it as a cumulative-mass cutoff over frequency-ranked clusters. The calibration score does not use the identity of the correct cluster, so this quantile alone gives no correctness-coverage guarantee. The proxy can return small sets when most sampled answers fall in one cluster, whether or not that cluster is correct. Its coverage and set size in \cref{tab:main} are empirical results on the full test split. Comparisons with ConU require the same generated answers, correctness criterion, evaluation population, and set-size units.

\paragraph{Joint calibration of aleatoric and epistemic uncertainty.}
Temperature scaling is a scalar post-hoc calibration method~\citep{guo2017calibration}. Because it has only one control parameter, it does not by itself provide independent controls for aleatoric and epistemic components. CLEAR jointly calibrates aleatoric and epistemic uncertainty in regression. Its conformalized variant selects a mixture coefficient on a validation split and a final scale on a separate calibration split~\citep{azizi2026clear}. JUCAL jointly optimizes two probability-calibration parameters by negative log-likelihood for classifier ensembles, whose outputs can subsequently be conformalized~\citep{lambrecht2026jucal}. \citet{azizi2025semf} introduce SEMF for predictive intervals through supervised expectation maximization, also in a scalar output space. CAELUM adapts the split-calibration construction of conformalized CLEAR to estimates from an LLM, using a \emph{prompt ensemble} of $n$ clarifications of the same query~\citep{hou2023decomposing}, each producing $m$ samples. Following ICE, mean within-clarification entropy is the epistemic proxy, and the pooled-minus-within residual is the aleatoric proxy. The relevant prior theory has distinct scopes. \citet{sale2025aleatoric} analyze the aleatoric--epistemic distinction in conformal prediction, \citet{cabezas2025epistemic} place epistemic uncertainty in conformal scores, and \citet{javanmardi2025optimal} optimize under second-order epistemic information. \citet{xin2026paraphrase} also vary the input via paraphrasing in conformal prediction, but apply it to multiple-choice QA without a decomposition step.

\begin{appendixblock}
\subsection{Comparison with CLEAR and JUCAL}\label{app:family}

\begin{table}[H]
\caption{Joint calibration of aleatoric and epistemic uncertainty across output spaces.}
\label{tab:family}
\centering
\small
\begin{tabular}{lccc}
\toprule
 & \textbf{CLEAR} & \textbf{JUCAL} & \textbf{CAELUM} \\
\midrule
Output space & $\mathbb{R}$ & $\{1, \ldots, K\}$ & Open-ended text \\
Output & Interval & Calibrated probabilities & Semantic clusters \\
Aleatoric proxy & Quantile regression & Temperature ($c_1$) & Between-clarification entropy \\
Epistemic proxy & Ensemble diversity & Diversity adjustment ($c_2$) & Within-clarification entropy \\
Calibration loss & Quantile loss & NLL & Set size (APSS) \\
Requires ensemble & Yes & Yes & No (single LLM) \\
\bottomrule
\end{tabular}
\end{table}
\end{appendixblock}

\begin{appendixblock}
\section{Theoretical analysis}\label{app:theory}

\subsection{Proof of \texorpdfstring{\cref{prop:coverage}}{Proposition 1} (support-conditional finite-sample coverage)}\label{app:coverage-proof}

\begin{assumption}[Exchangeability of items, generations, and matchability]\label{ass:exch}
The calibration items and the test point are i.i.d.\ draws of the augmented tuple $(x, y^*, \mathcal{G}(x))$, where $\mathcal{G}(x)$ denotes the LLM's generated clarifications, sampled answers, semantic clusters, and seeded randomized-APS uniforms produced by the same fixed pipeline applied independently to each item. Matchability $M(x,y^*,\mathcal{G}(x))$, written $M(x,y^*)$ in \cref{sec:setup}, is computed from this tuple. Conditioning on the post-generation event $\{M=1\}$ therefore preserves exchangeability of the conditional sub-sample. For \cref{prop:coverage}, the stratum is true matchability $\{M=1\}$. For \cref{thm:stratified}, the stratum is the learned event $\{\hat m(x)=1\}$, where $\hat m$ is trained on $\Vzero$ and uses only $x$.
\end{assumption}
\end{appendixblock}

Under \cref{ass:exch}, the support-conditional guarantee of \cref{prop:coverage} holds with the usual split-conformal finite-sample slack on the quantile level.

\begin{remark}[Fixed preprocessing]\label{rem:preprocessing}
The support-conditional guarantee requires that the nonconformity score construction is fixed with respect to $\Vtwo$ except for the exchangeable calibration scores themselves. This holds because (a) $\lambdastar$ is selected on $\Vone$ alone; (b) the sentence encoder $\phi$ and clustering threshold $\delta$ are fixed independently of $\Vtwo$ and the test point (they may be chosen on $\Vone$); (c) each instance's uncertainty estimates are computed from its own generated answers, with no cross-instance dependencies; and (d) randomized APS uses independent seeded uniforms, which are part of the exchangeable score randomness. If any of these components were fine-tuned on $\Vtwo$, the exchangeability argument would break.
\end{remark}

\begin{proof}
Let $\Vtwo^M=\{i\in\Vtwo: M(x_i,y_i^*,\mathcal{G}(x_i))=1\}$ be the calibration subset whose reference answers fall in the generated support, and let $n_M=|\Vtwo^M|$. For this support-restricted variant define
\[
    \tauhat_M =
    Q_{\lceil(1-\alpha)(n_M+1)\rceil/n_M}
    \bigl(\{\snorm_{\lambdastar,i}:i\in\Vtwo^M\}\bigr).
\]
Once $\lambdastar$ is fixed from $\Vone$, the normalized nonconformity score $\snorm_{\lambdastar}(x, y^*) = s(x, y^*) / w_{\lambdastar}(x)$ is fixed relative to $\Vtwo^M$. For the smoothed-randomized APS we replace $s$ by $s_{\mathrm{rand}}$, which is part of $\mathcal{G}(x)$ via the seeded uniforms, and treat the uniforms as part of the exchangeable score randomness. By \cref{ass:exch}, conditioning on $\{M=1\}$ preserves exchangeability, so the scores $\{\snorm_i:i\in\Vtwo^M\}\cup\{\snorm_{N+1}:M_{N+1}=1\}$ are exchangeable. The conformal quantile $\tauhat_M$ ensures $\mathbb{P}(\snorm_{N+1}\leq \tauhat_M\mid M=1)\geq 1-\alpha$. The bound conditions on $n_M$ and is uniform in its realization by exchangeability of the ranks. On $\{M=1\}$, the deterministic APS event $\snorm \leq \tauhat_M$ is equivalent to $s \leq \tauhat_M \cdot w_{\lambdastar}(x)$ and implies that the matching cluster is included in $\Cset_{\lambdastar}(x)$. For the smoothed-randomized score, $s_{\mathrm{rand}}\geq s$, so the randomized event $\{\snorm^{\mathrm{rand}}\leq\tauhat_M\}$ is a sufficient condition for $\{s\leq\tauhat_M\,w_{\lambdastar}(x)\}$ and hence for inclusion (we sacrifice some efficiency for deterministic prediction sets). When $\eta_{\lambdastar}(x;\tauhat_M) = 1$, \cref{eq:prediction_set} contains every generated cluster, including the matching cluster on $\{M=1\}$, so the implication still holds at saturation. This yields the support-conditional lower bound.
\end{proof}

\begin{remark}[Coverage over joint randomness]\label{rem:joint-randomness}
The APS nonconformity score $s(x, y^*)$ depends on the LLM's stochastic generation (clarifications and answers sampled with temperature $> 0$). The support-conditional guarantee in \cref{prop:coverage} holds over \emph{both} the randomness of the calibration/test split \emph{and} the LLM's sampling randomness, conditional on the generated support containing the answer. The nonconformity scores are therefore not deterministic functions of $(x, y^*)$ alone.
\end{remark}

\begin{remark}[Role of the weight function]\label{rem:weight_role}
The weight function $\wlam(x)$ serves two purposes. First, by normalizing the raw scores as in \cref{eq:normalized_score} before computing the conformal quantile, it makes $\tauhat_\lambda$ depend on $\lambda$, enabling the optimization in Step~1 to distinguish between different uncertainty weightings. Without normalization, the calibration scores are $\lambda$-independent so $\tauhat$ is constant. The per-item threshold $\eta_\lambda(x;\tauhat)=\min(\tauhat\cdot\wlam(x),1)$ is then monotonically non-decreasing in $\lambda$ (because $\wlam$ is, for $\Uepi\geq 0$), so APSS is monotonically non-decreasing and the argmin returns $\lambdastar=0$. Second, through the prediction set criterion in \cref{eq:prediction_set}, the weight scales the effective threshold, expanding sets more aggressively for high-uncertainty inputs and redistributing set sizes across inputs. This mirrors the role of two-parameter joint calibration in conformalized CLEAR~\citep{azizi2026clear}, where a calibration constant varies with the aleatoric/epistemic mixing coefficient to balance coverage and set efficiency.
\end{remark}

\subsection{On asymptotic score stability (informal remark)}

\begin{remark}[Asymptotic score stability]\label{rem:conditional}
Exact \emph{conditional} coverage cannot generally be guaranteed in a distribution-free, finite-sample setting without making the prediction sets uninformative~\citep{barber2020}. We do not claim it. As a heuristic asymptotic, if (i) the number of independently generated answers $m \to \infty$, (ii) the answer partition is fixed and finite, with the clarifications held fixed, and (iii) calibration and test data are i.i.d., then by the law of large numbers the empirical cluster proportions $p_k(x)$ converge to their population counterparts, and their Shannon entropies converge by continuity. Convergence of VNE in an infinite-dimensional kernel space requires additional assumptions. Convergence of the conformal quantile $\tauhat$ also requires the calibration sample size to grow and regularity at the population quantile. We do not establish those convergence results here. This does not imply distribution-free conditional coverage for every fixed $x$. Our greedy clustering is a computationally efficient approximation that is order-dependent. Condition (ii) therefore need not hold. We report empirical coverage in \cref{tab:main} and validate per-cell behavior rather than relying on this asymptotic remark.
\end{remark}

\subsection{Coverage given matchability}\label{app:matched-coverage}

\cref{prop:coverage} is conditional on generated support. An item $x$ is \emph{matchable} if some generated cluster $G_k(x)$ contains the reference answer in the sense of \cref{app:implementation}. Unmatched items receive the maximal nonconformity score $s = 1$, are not included in any prediction set, and are therefore never covered by construction. Letting $\pi$ denote the population matchability rate, $\mathbb{P}(y^*_{N+1} \in \Cset_{\lambdastar}(x_{N+1})) \leq \pi$ regardless of $\alpha$. \cref{fig:cov_match} compares coverage and empirical support on the same test split. \Cref{tab:main} reports the match rate on that same test split. The support ceiling $\pi$ alone precludes nominal $1 - \alpha = 0.9$ coverage on hard cells (e.g.\ Gemma-SimpleQA, $\hat\pi_{\rm test} = 0.21$). Strengthening the generation or matching pipeline (retrieval, stronger generators, NLI-based equivalence, LLM judges) can lift $\pi$, but it lifts the ceiling for all conformal methods rather than uniquely improving CAELUM. \cref{thm:stratified} (\cref{sec:stratified}) makes this dependence explicit by replacing population matchability with a learnable stratum and a false matchability penalty.

\subsection{Proof of \texorpdfstring{\cref{thm:stratified}}{Theorem 1} (stratified coverage)}\label{app:stratified-proof}

\begin{proof}[Proof of \cref{thm:stratified}]
Let $\hat m: \mathcal{X} \to \{0, 1\}$ be a binary classifier trained on $\Vzero$, which is disjoint from both $\Vtwo$ and the test point. Define the predicted-matchable stratum
$\mathcal{S} = \{i\in\Vtwo : \hat m(x_i) = 1\}$. Because $\hat m$ depends on $x$ only through a fixed function (no calibration or test labels) and was trained on a disjoint split, $\hat m(\cdot)$ is constant in distribution under any permutation of $\Vtwo \cup \{(x_{N+1}, y^*_{N+1})\}$.

Score coverage on the stratum follows from Mondrian split conformal prediction~\citep{vovk2003mondrian,bostrom2020mondrian} applied to the taxonomy $\{\hat m = 0, \hat m = 1\}$. Specifically, conditioning on the event $\hat m(x_{N+1}) = 1$, the calibration sub-sample $\mathcal{S}$ together with the test point is exchangeable. Their joint distribution is invariant to permutations among items with $\hat m = 1$. Applying split conformal to this conditional distribution with the conformal quantile $\tauhat_{\mathrm{strat}}$ defined in \cref{eq:tau_strat} yields
\begin{equation*}
    \mathbb{P}\bigl(\snorm_{\lambda_0}(x_{N+1}, y^*_{N+1}) \leq \tauhat_{\mathrm{strat}}(\lambda_0) \,\big|\, \hat m(x_{N+1}) = 1\bigr) \geq 1 - \alpha,
\end{equation*}
where $\snorm=1/w_{\lambda_0}(x)$ for unmatched hard-score items. Let $M$ denote true matchability and $\rho_{\hat m}=\mathbb{P}(M=0\mid \hat m=1)$. For deterministic APS, coverage is the event $\{M=1,\snorm\leq\tauhat_{\mathrm{strat}}(\lambda_0)\}$. With randomized calibration scores and deterministic prediction sets, this score event is a subset of coverage because $s_{\rm rand}\geq s$. In both cases
\begin{align*}
\mathbb{P}(\text{covered}\mid \hat m=1)
&\geq \mathbb{P}(\snorm\leq\tauhat_{\mathrm{strat}}\mid \hat m=1)
   - \mathbb{P}(M=0,\snorm\leq\tauhat_{\mathrm{strat}}\mid \hat m=1) \\
&\geq 1-\alpha-\mathbb{P}(M=0\mid \hat m=1)
 = 1-\alpha-\rho_{\hat m},
\end{align*}
which is \cref{eq:cond_cov}. Since coverage implies $M=1$, the upper bound
$\mathbb{P}(\text{covered}\mid \hat m=1)\leq 1-\rho_{\hat m}$ follows immediately.

For the marginal lower bound \cref{eq:marginal_lb}, decompose the coverage event by the $\hat m$ taxonomy:
\begin{align*}
    \mathbb{P}\bigl(y^*_{N+1} \in \Cset_{\lambda_0}^{\mathrm{strat}}(x_{N+1})\bigr)
    &= \mathbb{P}\bigl(y^*_{N+1} \in \Cset_{\lambda_0}^{\mathrm{strat}}(x_{N+1}) \,\big|\, \hat m(x_{N+1}) = 1\bigr) \cdot \mathbb{P}\bigl(\hat m(x_{N+1}) = 1\bigr) \\
    &\quad + \mathbb{P}\bigl(y^*_{N+1} \in \Cset_{\lambda_0}^{\mathrm{strat}}(x_{N+1}),\, \hat m(x_{N+1}) = 0\bigr) \\
    &\geq (1 - \alpha-\rho_{\hat m})\,\mathbb{P}\bigl(\hat m(x_{N+1}) = 1\bigr).
\end{align*}
This is equivalent to \cref{eq:marginal_lb} because
$\rho_{\hat m}\mathbb{P}(\hat m=1)=\mathbb{P}(M=0,\hat m=1)$. The corollary below makes the upper-bound implication explicit.
\end{proof}

\begin{corollary}[Ceiling on hard cells]\label{cor:tight}
On the predicted-matchable stratum, coverage cannot exceed $1-\rho_{\hat m}$. Thus low classifier precision lowers attainable coverage within the selected stratum, regardless of the threshold. A low population matchability rate $\pi$ instead caps marginal coverage and may limit the fraction of queries a precise classifier can select.
\end{corollary}
\begin{proof}[Sketch]
Direct from the fact that every covered item must be matchable.
\end{proof}

\begin{appendixblock}
\section{Algorithm}\label{app:algorithm}

\Cref{alg:caelum} summarizes the full CAELUM pipeline used for the main experiments, including the optional matchability-stratified phase.

\begin{algorithm}[H]
\caption{CAELUM: Calibrated Aleatoric-Epistemic Language Uncertainty Modeling}
\label{alg:caelum}
\begin{algorithmic}[1]
\Require LLM $\Mmodel$; calibration data $\Dcal$; $n$ clarifications; $m$ answers per clarification; nonempty finite grid $\Lambda\subset[0,\infty)$; error rate $\alpha\in(0,1)$.
\Statex
\State \textbf{Phase 1: Two-Stage Sampling and Decomposition}
\For{each $(x_i, y_i^*) \in \Dcal$}
    \State Generate clarifications $W_1, \ldots, W_n$ from $\Mmodel$ given $x_i$
    \For{each clarification $W_j$}
        \State Sample $m$ answers $y_{j1}, \ldots, y_{jm} \sim \Mmodel(\cdot \mid W_j)$
    \EndFor
    \State Cluster all $nm$ answers semantically $\to$ sorted proportions $\{p_k\}$, representatives $\{\hat{y}_k\}$
    \State Compute $\Utot(x_i)$ as pooled VNE and $\Uepi(x_i)$ as mean per-clarification VNE (\cref{app:vne}); set $\Uale(x_i) = \max(\Utot(x_i) - \Uepi(x_i),0)$
    \State Apply the fixed reference matcher to record $M_i$ and, if matched, $k_i^*$
    \State Set $s_i\gets1$ if $M_i=0$, otherwise $s_i\gets\sum_{k<k_i^*}p_k(x_i)$, adding $U_i p_{k_i^*}(x_i)$ for randomized scoring with independent $U_i\sim\mathrm{Unif}(0,1)$ fixed across $\lambda$ \Comment{\cref{eq:score}}
\EndFor
\Statex
\State \textbf{Phase 2: Split Calibration}
\State Split $\Dcal$ into nonempty $\Vone$ and $\Vtwo$, plus a disjoint $\Vzero$ if Phase 3 is used
\For{each $\lambda \in \Lambda$} \Comment{$\lambda$ selection on $\Vone$}
    \State $\snorm_{\lambda,i} \gets s_i \,/\, \wlam(x_i)$ for all $i \in \Vone$ \Comment{Score normalization in \cref{eq:normalized_score}}
    \State $\tauhat_\lambda \gets Q_{\lceil(1-\alpha)(|\Vone|+1)\rceil/|\Vone|}(\{\snorm_{\lambda,i}\}_{i \in \Vone})$
    \State $\APSS(\lambda) \gets \frac{1}{|\Vone|}\sum_{i \in \Vone} |\Cset_\lambda(x_i)|$ using $\tauhat_\lambda$ and \cref{eq:prediction_set}
\EndFor
\State $\lambdastar \gets \arg\min_{\lambda\in\Lambda}\APSS(\lambda)$, choosing the smallest minimizing weight
\State $\snorm_{\lambdastar,i} \gets s_i \,/\, w_{\lambdastar}(x_i)$ for all $i \in \Vtwo$ \Comment{Score normalization}
\State $\tauhat \gets Q_{\lceil(1-\alpha)(|\Vtwo|+1)\rceil/|\Vtwo|}(\{\snorm_{\lambdastar,i}\}_{i \in \Vtwo})$ \Comment{Full-population empirical threshold}
\Statex \hspace{\algorithmicindent}For Proposition~\ref{prop:coverage}, replace $\Vtwo$ by $\Vtwo^M$ in this threshold
\Statex
\State \textbf{Phase 3 (optional, for the matchability-stratified variant of \cref{thm:stratified}):}
\State Train binary classifier $\hat m: \mathcal{X}\to\{0,1\}$ on $\Vzero$ (disjoint from $\Vone, \Vtwo$, test) using question features available before generation
\State Restrict $\Vtwo$ to $\mathcal{S}=\{i\in\Vtwo: \hat m(x_i)=1\}$ and recompute the threshold $\tauhat_{\mathrm{strat}}(\lambdastar)$ via \cref{eq:tau_strat} on $\mathcal{S}$
\Statex
\State For a new $x$, repeat Phase 1 sampling, clustering, and uncertainty estimation without a reference, apply \cref{eq:prediction_set} with $\tauhat$, or with $\tauhat_{\mathrm{strat}}$ in Phase 3, and report coverage on $\{\hat m=1\}$ separately
\Statex
\Ensure Prediction set $\Cset_{\lambdastar}(x)$ via \cref{eq:prediction_set}; diagnostic $\lambdastar$; per-instance $R(x)$ via \cref{eq:diagnostic}; with Phase~3, predicted-matchable set $\Cset_{\lambdastar}^{\mathrm{strat}}(x)$.
\end{algorithmic}
\end{algorithm}
\end{appendixblock}

\section{Dataset descriptions}\label{app:datasets}

\emph{Factual benchmarks.} These benchmarks test factual recall. A dataset label does not determine which component of prompt variability will be largest for a particular generator.

\begin{itemize}
    \item \textbf{TriviaQA}~\citep{joshi2017triviaqa}: Factual questions and answers authored by trivia enthusiasts. We use the 300-item curated subset released by \citet{spectraluncertainty2025} for direct comparability.
    \item \textbf{OpenNQ}~\citep{kwiatkowski2019nq}: Factual questions drawn from real Google search queries (Natural Questions development set). We use the 300-item curated subset released by \citet{spectraluncertainty2025}.
    \item \textbf{SimpleQA}~\citep{wei2024simpleqa}: Short-form factual questions designed as a challenging factuality benchmark, subsampled to 500 items. The benchmark was adversarially collected against GPT-4 responses to stress-test short-form factuality.
    \item \textbf{PopQA}~\citep{mallen2023popqa}: Entity-centric factual questions annotated with subject popularity, subsampled to 500 items.
\end{itemize}

\emph{Ambiguous benchmarks.} These benchmarks include questions or instructions with multiple valid interpretations, together with unambiguous items. Their annotations allow a separate check of the relationship between prompt variability and ambiguity.

\begin{itemize}
    \item \textbf{AmbigQA}~\citep{min2020ambigqa}: Open-domain questions from NQ, including questions that admit different interpretations. We use the 200-item curated subset of \citet{spectraluncertainty2025}.
    \item \textbf{AmbigInst}~\citep{hou2023decomposing}: Synthetic instruction-following tasks with ambiguous and unambiguous instructions. We use all 304 samples in the release of \citet{spectraluncertainty2025}.
\end{itemize}

\section{Implementation details}\label{app:implementation}

\paragraph{Notation summary.} For reference, the symbols used in the body and proofs are:
\begin{center}
\small
\begin{tabularx}{\textwidth}{@{}lX@{}}
\toprule
$X,Y,C$ & Query, pooled answer-cluster label, and clarification index in the Shannon identity, with $C$ uniform over clarifications and $\mathbb{P}(Y=k\mid C=i,X=x)=p_k^{(i)}(x)$ \\
$K=K(x),\,p_k^{(i)}(x)$ & Number of clusters and fraction of clarification $i$'s answers assigned to pooled cluster $k$ \\
$\log,\,0\log0$ & Natural logarithm, with the entropy convention $0\log0=0$ \\
$\Utot, \Uepi, \Uale$ & Pooled, epistemic, and aleatoric VNE signals, \cref{eq:epistemic} and \cref{app:vne} \\
$\wlam(x)=1+\Uale+\lambda\,\Uepi$ & Per-instance weight in the conformal score, \cref{eq:weight} \\
$\lambda^\star$ & Per-cell APSS-minimizing mixing coefficient, \cref{eq:lambda_selection} \\
$\snorm_\lambda = s/\wlam$ & Weight-normalized APS score, \cref{eq:normalized_score} \\
$\tauhat$ & Empirical full-$\Vtwo$ threshold in \cref{eq:tau}; $\tauhat_M$ is the support-restricted threshold in \cref{prop:coverage}; $\tauhat_{\mathrm{strat}}$ on $\mathcal{S}\subseteq\Vtwo$ \\
$M(x,y^*)$ & True matchability (reference appears in some sampled cluster) \\
$\hat m(x)\in\{0,1\}$ & Pre-generation logistic classifier predicting $M$ \\
$\pi=\mathbb{P}(M=1)$ & Population matchability rate \\
$\rho_{\hat m}=\mathbb{P}(M=0\mid \hat m=1)$ & False-matchability rate inside the predicted-matchable stratum \\
$\mathcal{S}=\{i\in\Vtwo:\hat m(x_i)=1\}$ & Calibration sub-sample for the stratified threshold \\
$\Vzero, \Vone, \Vtwo$ & Calibration splits: $\hat m$ training, $\lambda^\star$ selection, $\tauhat$ calibration \\
$R(x)=\log(\lambda^\star\Uepi+\epsilon)-\log(\Uale+\epsilon)$ & Diagnostic ratio for each instance, \cref{eq:diagnostic} \\
\bottomrule
\end{tabularx}
\end{center}

\paragraph{Lambda grid.} The grid $\Lambda$ consists of 10 linearly spaced values in $[0, 0.09]$ concatenated with 4001 log-spaced values in $[10^{-1}, 10^2]$, yielding 4011 candidate values. This provides fine resolution near $\lambda = 0$ (where the transition from aleatoric-only to mixed behavior occurs) and broader coverage at larger scales.

\paragraph{Matchability classifier features.} The inputs to $\hat m$ include the all-mpnet-base-v2 sentence embedding of the question and other sentence features: character length and token length, number of capitalized and numeric tokens, a year-pattern feature, and one-hot encoding of the first word of the question using a fixed vocabulary of $18$ words (\emph{what, who, when, where, why, how, which, is, are, was, were, did, do, does, can, name, list, describe}) and an \emph{other} category.

\paragraph{Data splits.} The main table uses a fixed 40\%/30\%/30\% split ($\Vone$/$\Vtwo$/Test) with seed 42 and deterministic APS, and \cref{tab:multiseed-apss} repeats that split for seeds 42--45 with seeded randomized APS. The coefficient SDs in \cref{tab:main,tab:lambda_coefficients} instead use deterministic APS on those four splits. For 200 samples per dataset, the seed 42 main split yields 80 $\Vone$ / 60 $\Vtwo$ / 60 Test. The pre-generation matchability experiment uses a separate 30\% $\Vzero$ / 30\% $\Vtwo$ / 40\% Test split, and the integrated $\lambdastar$ matchability ablation uses four 25\% splits.

\paragraph{Cluster alignment.} After clustering, the cluster proportions and representative answers are sorted in descending order of proportion. The correct cluster index must point into this sorted array, otherwise APS scoring is corrupted.

\paragraph{Correctness matching.} The reference answer is matched under the following tiers, applied in order to cluster representatives and then to raw cluster members for the lexical/overlap tiers: (1) exact substring (in either direction, after lower-casing and punctuation stripping); (2) normalized-substring match after Unicode and diacritic folding; (3) depluralized whole-word match for short answers; (4) single-token containment in short candidates ($\leq 6$ tokens) to handle entity answers; (5) ROUGE-L $> 0.15$; (6) Jaccard $> 0.25$. If no lexical/overlap tier matches, tier (7) uses embedding cosine $> 0.4$ (all-mpnet-base-v2) on cluster representatives. If no cluster matches, the sample is unmatched and receives the maximal nonconformity score ($s = 1.0$). These tiers are heuristic and can accept false positives as well as miss valid paraphrases. The embedding tier is permissive. A higher measured match rate is not by itself evidence of a better matcher. \cref{app:embedding} stress-tests the resulting match rates across encoders.

\section{von Neumann entropy decomposition}\label{app:vne}

We use an RBF kernel as in \citet{spectraluncertainty2025}, fixing $\gamma=1$ for all datasets; their AmbigInst experiments instead use $\gamma=100$. L2-normalized sentence embeddings $\phi(y)$ define $k(y,y')=\exp(-\gamma\|\phi(y)-\phi(y')\|^2)$ with $\gamma=1$. For $N$ answers, let $K_{ij}=k(y_i,y_j)/N$, so $\operatorname{Tr}K=1$. The von Neumann entropy is $H_{\rm VN}(K)=-\sum_r\mu_r\log\mu_r$, with non-negative eigenvalues $\mu_r$ and natural logarithms. We use unscaled entropy for both the pooled and per-clarification kernels.

The normalized Gram matrix has the same nonzero spectrum as the empirical density operator in the common reproducing-kernel Hilbert space. For the shared kernel feature map $\psi$, where $\langle\psi(y),\psi(y')\rangle=k(y,y')$, let $\rho_i=m^{-1}\sum_j\psi(y_{ij})\otimes\psi(y_{ij})$, where $u\otimes u$ is the operator $v\mapsto u\langle u,v\rangle$. With equal answer counts, $\rho_{\rm pool}=n^{-1}\sum_i\rho_i$, so entropy concavity gives $H_{\rm VN}(\rho_{\rm pool})\geq n^{-1}\sum_iH_{\rm VN}(\rho_i)$. Computing each Gram block separately does not alter this identity. The encoder, embeddings, kernel, and bandwidth are shared. The implementation clips negative numerical eigenvalues and floors the residual at zero. Unequal answer counts require size-weighted rather than uniform averaging for this identity. Finite-sample entropy estimates are biased, and this decomposition measures response variability under the chosen clarification procedure rather than ground-truth ambiguity.

\section{Embedding-sensitivity ablation}\label{app:embedding}

We recomputed clustering, VNE, and reference matching for three cells (TriviaQA / AmbigQA / PopQA on Qwen3-8B) with four sentence encoders, holding generations, splits, and numerical thresholds fixed. The default protocol uses all-mpnet-base-v2 with similarity threshold $\delta = 0.7$.

On the three audited cells, the default all-mpnet-base-v2 encoder ($\delta=0.7$) and an alternative MiniLM-L12 encoder differ in match rate by less than three percentage points and in AUROC for $\Uepi$ by less than $0.02$. MiniLM-L12 produces larger sets on all three cells. Larger encoders (bge-large-en-v1.5, e5-large-v2) result in excessive merging at $\delta=0.7$ and, for e5-large-v2, degenerate clustering. We follow the encoder used in \citet{spectraluncertainty2025} for direct comparability.

\begin{table}[H]
\caption{Embedding-sensitivity ablation. We vary only the sentence encoder, which is used for clustering, VNE, and reference matching. mpnet (our default) and MiniLM-L12 give similar match rates and $\Uepi$ AUROC, but MiniLM-L12 produces larger sets on all three cells. bge-large becomes too permissive at $\delta=0.7$ and degrades $\Uepi$ AUROC on two of three cells. e5-large yields 100\% match (degenerate clustering) at the same threshold. Results are unranked because the encoders also change the support labels. The coefficients use deterministic APS and the split with seed 42.}
\label{tab:embedding_ablation}
\centering
\small
\begingroup
\small
\setlength{\tabcolsep}{3.5pt}
\renewcommand{\arraystretch}{1.12}
\begin{tabular*}{\linewidth}{@{\extracolsep{\fill}}lrrrrr@{}}
\toprule
Encoder & Match\% & $\lambda^\star$ & Coverage & APSS & $\Uepi$ AUROC \\
\midrule
\multicolumn{6}{@{}l}{\textbf{TriviaQA}}\\[3pt]
all-mpnet-base-v2 & 78.0 & 47.4 & 0.811 & 1.82 & 0.709 \\
all-MiniLM-L12-v2 & 75.3 & 9.58 & 0.767 & 2.02 & 0.724 \\
bge-large-en-v1.5 & 93.7 & 17.1 & 0.933 & 1.28 & 0.519 \\
e5-large-v2 & 100.0 & 0.00 & 1.000 & 1.00 & n/a \\
\midrule[0.3pt]
\addlinespace[2pt]
\multicolumn{6}{@{}l}{\textbf{AmbigQA}}\\[3pt]
all-mpnet-base-v2 & 67.0 & 7.54 & 0.667 & 1.30 & 0.632 \\
all-MiniLM-L12-v2 & 66.5 & 78.0 & 0.617 & 1.55 & 0.636 \\
bge-large-en-v1.5 & 88.0 & 24.8 & 0.883 & 1.00 & 0.736 \\
e5-large-v2 & 100.0 & 0.00 & 1.000 & 1.00 & n/a \\
\midrule[0.3pt]
\addlinespace[2pt]
\multicolumn{6}{@{}l}{\textbf{PopQA}}\\[3pt]
all-mpnet-base-v2 & 53.6 & 3.15 & 0.533 & 1.74 & 0.780 \\
all-MiniLM-L12-v2 & 52.2 & 43.9 & 0.527 & 2.13 & 0.778 \\
bge-large-en-v1.5 & 76.2 & 56.4 & 0.727 & 1.65 & 0.772 \\
e5-large-v2 & 100.0 & 0.00 & 1.000 & 1.00 & n/a \\
\bottomrule
\end{tabular*}
\endgroup

\end{table}

\section{Clarifier-quality sensitivity}\label{app:clarifier_sensitivity}

The decomposition reflects the LLM's clarification diversity, so clarifier quality is part of the measured pipeline. All main experiments use the answer LLM itself as the clarifier, which keeps the support-ceiling analysis attributable to a single sampled generator. External clarifiers are not used in the reported experiments. An external clarifier may change matchability, and its benefit would require a separate evaluation.

\section{Evaluation metrics}\label{app:metrics}

\begin{itemize}
    \item \textbf{Marginal coverage.} Fraction of test queries where $y^* \in \Cset(x)$. For sampled semantic sets this rate is at most the test support rate. The support-conditional and learned stratum guarantees are stated separately (we use $\alpha=0.1$).
    \item \textbf{APSS (Average Prediction Set Size).} Mean number of semantic clusters in prediction sets across test queries indexed by $T$: $\APSS = \frac{1}{|T|}\sum_{i \in T} |\Cset(x_i)|$. Lower is better at equal coverage, as smaller sets are more informative.
    \item \textbf{Support-failure / hallucination AUROC.} Area under the ROC curve using $\Uepi$ (or $\Utot$, $\Uale$) as a predictor of support failure, where a sample is labeled positive if the reference answer does not match any generated cluster under the fixed matching rule. This often indicates hallucination, but the measured event is absence from generated support.
    \item \textbf{Ambiguity AUROC.} Area under the ROC curve using $\Uale$ as a predictor of the binary ambiguity label, available for AmbigQA and AmbigInst in the main benchmark.
    \item \textbf{Diagnostic $\lambdastar$.} The calibrated epistemic weight selected on $\Vone$ via \cref{eq:lambda_selection}. The value describes the fitted score weight. A regime interpretation requires a nonflat objective and split stability.
\end{itemize}

\section{Full comparison of semantic prediction sets}\label{app:complete-methods}
Table~\ref{tab:complete-methods} expands the main comparison with paired coverage and APSS for all six methods. It uses the same generated answers and deterministic seed-42 split. Set size alone cannot establish efficiency when the methods achieve different coverage. Table~\ref{tab:ncapss} provides a separate comparison at matched coverage.

\begin{table}[H]
\caption{Coverage and average set size for all six set rules at nominal $\alpha=0.1$. Every pair of columns reports both quantities for the same method and test split. SC proxy is the simplified self-consistency baseline, not original ConU. As in Table~\ref{tab:main}, \textbf{bold} and \underline{underline} mark the smallest and second-smallest displayed sizes among methods tied for the highest coverage in each row.}
\label{tab:complete-methods}
\centering
\begingroup
\small
\setlength{\tabcolsep}{1.8pt}
\renewcommand{\arraystretch}{1.12}
\begin{tabular*}{\linewidth}{@{\extracolsep{\fill}}lrr@{\hspace{8pt}}rr@{\hspace{8pt}}rr@{\hspace{8pt}}rr@{\hspace{8pt}}rr@{\hspace{8pt}}rr@{}}
\toprule
& \multicolumn{2}{c}{CAELUM} & \multicolumn{2}{c}{Vanilla} & \multicolumn{2}{c}{Equal} & \multicolumn{2}{c}{Epi. only} & \multicolumn{2}{c}{SC proxy} & \multicolumn{2}{c}{SE-CP} \\
\cmidrule(lr){2-3}\cmidrule(lr){4-5}\cmidrule(lr){6-7}\cmidrule(lr){8-9}\cmidrule(lr){10-11}\cmidrule(lr){12-13}
Dataset & Cov. & APSS & Cov. & APSS & Cov. & APSS & Cov. & APSS & Cov. & APSS & Cov. & APSS \\
\midrule
\multicolumn{13}{@{}l}{\textbf{Qwen3-8B}}\\[3pt]
TriviaQA & 0.811 & \underline{1.82} & 0.811 & 1.84 & 0.811 & 1.84 & 0.811 & \textbf{1.81} & 0.778 & 1.02 & 0.811 & 1.84 \\
AmbigQA & 0.667 & \textbf{1.30} & 0.667 & \textbf{1.30} & 0.667 & \textbf{1.30} & 0.667 & \textbf{1.30} & 0.650 & 1.00 & 0.667 & \textbf{1.30} \\
OpenNQ & 0.700 & \underline{1.20} & 0.700 & \underline{1.20} & 0.700 & \underline{1.20} & 0.700 & \underline{1.20} & 0.700 & \textbf{1.00} & 0.700 & \underline{1.20} \\
AmbigInst & 0.935 & \textbf{1.00} & 0.935 & \textbf{1.00} & 0.935 & \textbf{1.00} & 0.935 & \textbf{1.00} & 0.935 & \underline{1.05} & 0.935 & \textbf{1.00} \\
SimpleQA & 0.367 & \underline{1.77} & 0.367 & \underline{1.77} & 0.367 & \underline{1.77} & 0.367 & \textbf{1.75} & 0.327 & 1.02 & 0.367 & \underline{1.77} \\
PopQA & 0.533 & \textbf{1.74} & 0.533 & 1.80 & 0.533 & 1.79 & 0.533 & \underline{1.75} & 0.520 & 1.02 & 0.533 & 1.79 \\
\midrule[0.3pt]
\addlinespace[2pt]
\multicolumn{13}{@{}l}{\textbf{Ministral-3-8B-Instruct-2512}}\\[3pt]
TriviaQA & 0.811 & \textbf{2.27} & 0.811 & 2.34 & 0.811 & 2.37 & 0.811 & \underline{2.28} & 0.767 & 1.07 & 0.811 & 2.37 \\
AmbigQA & 0.733 & \textbf{3.90} & 0.717 & 4.08 & 0.717 & 4.05 & 0.733 & \underline{3.92} & 0.717 & 1.90 & 0.717 & 4.05 \\
OpenNQ & 0.733 & \textbf{1.61} & 0.733 & \underline{1.63} & 0.733 & \underline{1.63} & 0.722 & 1.60 & 0.711 & 1.00 & 0.733 & \underline{1.63} \\
AmbigInst & 0.859 & \textbf{2.32} & 0.859 & \textbf{2.32} & 0.859 & 2.75 & 0.859 & 2.99 & 0.859 & \underline{2.63} & 0.859 & 2.75 \\
SimpleQA & 0.340 & 2.31 & 0.340 & \underline{2.30} & 0.340 & 2.31 & 0.340 & \textbf{2.29} & 0.287 & 1.00 & 0.340 & 2.31 \\
PopQA & 0.487 & \textbf{3.57} & 0.487 & \underline{3.62} & 0.487 & \textbf{3.57} & 0.487 & \textbf{3.57} & 0.447 & 1.12 & 0.487 & \textbf{3.57} \\
\midrule[0.3pt]
\addlinespace[2pt]
\multicolumn{13}{@{}l}{\textbf{Gemma-4-E4B-it}}\\[3pt]
TriviaQA & 0.678 & \underline{1.74} & 0.678 & 1.77 & 0.678 & 1.77 & 0.678 & \textbf{1.73} & 0.656 & 1.01 & 0.678 & 1.77 \\
AmbigQA & 0.633 & \textbf{2.17} & 0.633 & \underline{2.18} & 0.633 & \textbf{2.17} & 0.633 & \textbf{2.17} & 0.583 & 1.05 & 0.633 & \textbf{2.17} \\
OpenNQ & 0.556 & \textbf{1.51} & 0.556 & \textbf{1.51} & 0.556 & \textbf{1.51} & 0.556 & \textbf{1.51} & 0.544 & 1.00 & 0.556 & \textbf{1.51} \\
AmbigInst & 0.935 & \underline{1.96} & 0.924 & 1.97 & 0.935 & 1.99 & 0.935 & \textbf{1.89} & 0.870 & 1.42 & 0.935 & 1.99 \\
SimpleQA & 0.207 & \textbf{1.66} & 0.207 & \textbf{1.66} & 0.207 & \underline{1.67} & 0.207 & \underline{1.67} & 0.180 & 1.00 & 0.207 & \underline{1.67} \\
PopQA & 0.373 & \underline{1.94} & 0.373 & \textbf{1.93} & 0.373 & 1.95 & 0.373 & \textbf{1.93} & 0.347 & 1.02 & 0.373 & 1.95 \\
\bottomrule
\end{tabular*}
\endgroup

\end{table}

\section{Weight selection on a separate validation subset}\label{app:lambda-selection}

The procedure for selecting $\lambda^\star$ in \cref{eq:lambda_selection} computes the conformal threshold $\hat\tau_\lambda$ on $\Vone$ and measures APSS \emph{on the same $\Vone$} for each candidate $\lambda$. With a 4011-point grid on a small $\Vone$ ($\approx 80$--$200$ items per cell), this is susceptible to selection bias.

A held-out variant splits $\Vone$ into $\Vone^{\mathrm{train}}$ (60\%) and $\Vone^{\mathrm{eval}}$ (40\%) and computes $\tauhat_\lambda$ on $\Vone^{\mathrm{train}}$ but evaluates the APSS for the argmin on $\Vone^{\mathrm{eval}}$. This separates threshold estimation from weight evaluation, whereas the original procedure uses all of $\Vone$ for both.

The main 18-cell comparison uses selection on all of $\Vone$. Across seeds, CAELUM and Vanilla have nearly identical cell-mean APSS, and their differences lie within the reported split variation (\cref{tab:multiseed-apss}).

\paragraph{Standardized weight and set-score selection.}\label{app:standardized-weight}
The judged rerun in \cref{app:historical-scale} and the semantic sets in \cref{app:missing-mass} use a different variant. Let $\mu_{\rm ale}$ and $\mu_{\rm epi}$ be the means of $\Uale$ and $\Uepi$ over all of $\Vone$, replacing a mean below $10^{-8}$ by $1$. This variant uses
\[
    w^{\rm std}_\lambda(x)=1+\frac{\Uale(x)}{\mu_{\rm ale}}
        +\lambda\frac{\Uepi(x)}{\mu_{\rm epi}}.
\]
It uses randomized APS and the 60/40 split of $\Vone$ described above. For each weight, the threshold is fitted on $\Vone^{\mathrm{train}}$, and the selection objective on $\Vone^{\mathrm{eval}}$ is mean set size plus $2/\alpha$ times the fraction of references absent from the prediction set. The first minimizing grid weight is selected. The judged rerun uses all items in each subset. The abstention experiment instead restricts both this threshold calculation and the selection objective to matched items, and fits the final semantic threshold on matched items in the first half of $\Vtwo$. The uncertainty means still use all of $\Vone$. These weights and their selection objective differ from \cref{tab:main}, so their fitted coefficients are not directly comparable.

\begin{table}[H]
\caption{Per-cell APSS (mean $\pm$ SD across seeds $42$--$45$) and $\lambdastar$ at nominal $\alpha = 0.1$. \textbf{Bold} and \underline{underline} mark the smallest and second-smallest displayed mean APSS, including ties. APSS values and their ranks are descriptive because coverage is not matched across methods. CAELUM and Vanilla agree within their per-cell split SD on all $18$ cells. The largest CAELUM APSS SD is $0.37$ and the largest Vanilla APSS SD is $0.38$. SE-CP and TrueAPS have larger SD on a few high-variance cells. $\lambdastar$ varies widely across seeds on cells with a nearly flat objective (e.g.\ Qwen3-TriviaQA $14.6 \pm 23.8$), even though the resulting APSS is much more stable. Flat objectives do not identify a regime from $\lambdastar$.}
\label{tab:multiseed-apss}
\centering
\footnotesize
\setlength{\tabcolsep}{3pt}
\begingroup
\small
\setlength{\tabcolsep}{3.5pt}
\renewcommand{\arraystretch}{1.12}
\begin{tabular*}{\linewidth}{@{\extracolsep{\fill}}lrrrrrr@{}}
\toprule
& & \multicolumn{4}{c}{Average set size (mean $\pm$ split SD)} & \\
\cmidrule(lr){3-6}
Dataset & Match\% & CAELUM & Vanilla & SE-CP & TrueAPS & $\lambda^\star$ \\
\midrule
\multicolumn{7}{@{}l}{\textbf{Qwen3-8B}}\\[3pt]
TriviaQA & 78.0 & \underline{1.82}$\pm$0.13 & \underline{1.82}$\pm$0.12 & \textbf{1.81}$\pm$0.12 & 1.84$\pm$0.12 & 14.6$\pm$23.8 \\
AmbigQA & 67.0 & \textbf{1.30}$\pm$0.08 & \textbf{1.30}$\pm$0.08 & \textbf{1.30}$\pm$0.08 & \textbf{1.30}$\pm$0.08 & 13.4$\pm$8.7 \\
OpenNQ & 63.7 & \textbf{1.19}$\pm$0.03 & \textbf{1.19}$\pm$0.03 & \textbf{1.19}$\pm$0.03 & \textbf{1.19}$\pm$0.03 & 0.0$\pm$0.0 \\
AmbigInst & 100.0 & 1.66$\pm$0.05 & 1.67$\pm$0.05 & \textbf{1.00}$\pm$0.00 & \underline{1.59}$\pm$0.07 & 0.0$\pm$0.0 \\
SimpleQA & 35.6 & \textbf{1.73}$\pm$0.09 & \textbf{1.73}$\pm$0.09 & \textbf{1.73}$\pm$0.09 & \textbf{1.73}$\pm$0.09 & 8.4$\pm$14.5 \\
PopQA & 53.6 & \underline{1.82}$\pm$0.04 & \underline{1.82}$\pm$0.04 & \textbf{1.81}$\pm$0.03 & 1.83$\pm$0.04 & 4.2$\pm$4.5 \\
\midrule[0.3pt]
\addlinespace[2pt]
\multicolumn{7}{@{}l}{\textbf{Ministral-3-8B-Instruct-2512}}\\[3pt]
TriviaQA & 84.7 & \underline{2.84}$\pm$0.28 & \underline{2.84}$\pm$0.29 & \textbf{2.45}$\pm$0.36 & 2.85$\pm$0.28 & 9.6$\pm$13.2 \\
AmbigQA & 67.0 & \underline{3.72}$\pm$0.37 & 3.74$\pm$0.38 & \textbf{3.70}$\pm$0.40 & 3.76$\pm$0.40 & 21.1$\pm$33.7 \\
OpenNQ & 67.0 & \textbf{1.74}$\pm$0.08 & \textbf{1.74}$\pm$0.08 & \textbf{1.74}$\pm$0.08 & \textbf{1.74}$\pm$0.08 & 17.0$\pm$10.0 \\
AmbigInst & 91.4 & 4.55$\pm$0.15 & 4.42$\pm$0.34 & \textbf{2.64}$\pm$0.78 & \underline{3.83}$\pm$0.74 & 1.2$\pm$1.6 \\
SimpleQA & 42.6 & \textbf{2.45}$\pm$0.15 & \textbf{2.45}$\pm$0.15 & \textbf{2.45}$\pm$0.15 & \underline{2.46}$\pm$0.15 & 0.2$\pm$0.1 \\
PopQA & 51.6 & \underline{3.36}$\pm$0.36 & \underline{3.36}$\pm$0.36 & \textbf{3.33}$\pm$0.36 & 3.39$\pm$0.35 & 0.4$\pm$0.5 \\
\midrule[0.3pt]
\addlinespace[2pt]
\multicolumn{7}{@{}l}{\textbf{Gemma-4-E4B-it}}\\[3pt]
TriviaQA & 68.7 & \textbf{1.75}$\pm$0.02 & \textbf{1.75}$\pm$0.03 & \textbf{1.75}$\pm$0.03 & \textbf{1.75}$\pm$0.02 & 4.7$\pm$6.2 \\
AmbigQA & 58.0 & 2.18$\pm$0.04 & \underline{2.17}$\pm$0.04 & \textbf{2.16}$\pm$0.04 & 2.18$\pm$0.04 & 6.4$\pm$7.4 \\
OpenNQ & 51.7 & \textbf{1.56}$\pm$0.06 & \textbf{1.56}$\pm$0.06 & \textbf{1.56}$\pm$0.06 & \textbf{1.56}$\pm$0.06 & 0.0$\pm$0.0 \\
AmbigInst & 96.1 & 2.48$\pm$0.20 & 2.50$\pm$0.19 & \textbf{1.90}$\pm$0.09 & \underline{2.25}$\pm$0.16 & 24.6$\pm$22.9 \\
SimpleQA & 21.2 & \textbf{1.74}$\pm$0.05 & \textbf{1.74}$\pm$0.05 & \underline{1.75}$\pm$0.05 & \underline{1.75}$\pm$0.05 & 0.1$\pm$0.1 \\
PopQA & 38.0 & \textbf{1.92}$\pm$0.06 & \textbf{1.92}$\pm$0.06 & \underline{1.93}$\pm$0.06 & \underline{1.93}$\pm$0.06 & 3.8$\pm$6.0 \\
\bottomrule
\end{tabular*}
\endgroup

\end{table}

\begin{appendixblock}
\section{Selected weight coefficients}\label{app:lambda_heatmap}

\Cref{tab:lambda_coefficients} reports the coefficients from \cref{tab:main} using deterministic APS with seed 42. Their magnitudes alone do not establish stability or identify an uncertainty regime.

\begin{table}[H]
\caption{Selected $\lambda^\star$ at seed 42 under deterministic APS, rounded to two decimals. Parentheses give the population SD of $\lambda^\star$ over seeds 42--45 under the same deterministic APS protocol.}
\label{tab:lambda_coefficients}
\centering
\begingroup
\small
\setlength{\tabcolsep}{12pt}
\renewcommand{\arraystretch}{1.12}
\begin{tabular*}{\linewidth}{@{\extracolsep{\fill}}lrrr@{}}
\toprule
\textbf{Dataset} & \multicolumn{1}{c}{\textbf{Qwen3-8B}} & \multicolumn{1}{c}{\textbf{Ministral-3-8B}} & \multicolumn{1}{c}{\textbf{Gemma-4-E4B-it}} \\
& & \multicolumn{1}{c}{\textbf{Instruct-2512}} & \\
\midrule
TriviaQA & 47.42 (14.5) & 26.96 (25.5) & 45.74 (13.6) \\
AmbigQA & 7.54 (33.9) & 18.41 (24.5) & 9.32 (5.2) \\
OpenNQ & 0.00 (0.0) & 6.19 (14.8) & 0.00 (15.0) \\
AmbigInst & 0.00 (0.0) & 0.00 (25.5) & 6.17 (18.9) \\
SimpleQA & 0.00 (12.2) & 0.32 (20.9) & 0.28 (0.1) \\
PopQA & 3.15 (38.7) & 1.43 (22.5) & 8.51 (7.9) \\
\bottomrule
\end{tabular*}
\endgroup

\end{table}
\end{appendixblock}

\section{Matched-coverage APSS (NCAPSS)}\label{app:ncapss}

Smaller sets in \cref{tab:main} can reflect lower coverage. To compare sizes at matched coverage, \emph{normalized-coverage APSS} (NCAPSS) sets $q^\star=\min(1-\alpha,\text{validation match rate})$ for each cell, sweeps each method over $\alpha\in[0.05,0.4]$, and interpolates APSS on its test coverage curve at $q^\star$; if every swept coverage exceeds $q^\star$, it reports the smallest swept APSS. A dagger marks methods whose sweep cannot reach the target. This use of test coverage makes NCAPSS a post-hoc diagnostic, not a deployable selection rule. Like the main table, the sweep uses deterministic APS.

\begin{table}[H]
\caption{APSS at matched coverage $q^\star$ (smallest swept APSS when every swept coverage exceeds $q^\star$). \textbf{Bold} and \underline{underline} mark the smallest and second-smallest displayed sizes among methods that reach the target. Daggers mark unattainable targets and are excluded from the ranking. Vanilla beats CAELUM on $2/11$ matched cells. The SC proxy cannot reach the target on $9/18$.}
\label{tab:ncapss}
\centering
\footnotesize
\setlength{\tabcolsep}{3.5pt}
\begingroup
\small
\setlength{\tabcolsep}{3.5pt}
\renewcommand{\arraystretch}{1.12}
\begin{tabular*}{\linewidth}{@{\extracolsep{\fill}}lrrrrrrr@{}}
\toprule
& & \multicolumn{6}{c}{Average set size at matched coverage $\downarrow$} \\
\cmidrule(lr){3-8}
Dataset & $q^\star$ & CAELUM & Vanilla & Equal & Epi. & SC proxy & SE-CP \\
\midrule
\multicolumn{8}{@{}l}{\textbf{Qwen3-8B}}\\[3pt]
TriviaQA & 0.77 & \textbf{1.00} & \textbf{1.00} & \textbf{1.00} & \textbf{1.00} & \textbf{1.00} & \textbf{1.00} \\
AmbigQA & 0.67 & 1.20 & \textbf{1.07} & \underline{1.10} & 1.20 & \textbf{1.07} & \underline{1.10} \\
OpenNQ & 0.61 & \underline{1.06} & 1.17 & 1.14 & \underline{1.06} & \textbf{1.00} & 1.14 \\
AmbigInst & 0.90 & \textbf{1.00} & \textbf{1.00} & \textbf{1.00} & \textbf{1.00} & \underline{1.01} & \textbf{1.00} \\
SimpleQA & 0.35 & \textbf{1.68} & \textbf{1.68} & \underline{1.73} & \textbf{1.68} & 1.00$^{\dagger}$ & \underline{1.73} \\
PopQA & 0.54 & \underline{1.64} & 1.66 & \textbf{1.56} & 1.67 & 1.00$^{\dagger}$ & \textbf{1.56} \\
\midrule[0.3pt]
\addlinespace[2pt]
\multicolumn{8}{@{}l}{\textbf{Ministral-3-8B-Instruct-2512}}\\[3pt]
TriviaQA & 0.86 & 1.92$^{\dagger}$ & 1.92$^{\dagger}$ & 2.37$^{\dagger}$ & 2.28$^{\dagger}$ & 1.00$^{\dagger}$ & 2.37$^{\dagger}$ \\
AmbigQA & 0.64 & \textbf{1.07} & \underline{1.08} & \underline{1.08} & \textbf{1.07} & 1.38 & \underline{1.08} \\
OpenNQ & 0.64 & 1.31 & 1.31 & 1.22 & \underline{1.13} & \textbf{1.00} & 1.22 \\
AmbigInst & 0.90 & 3.92$^{\dagger}$ & 2.32$^{\dagger}$ & 3.25$^{\dagger}$ & 3.86$^{\dagger}$ & 2.32$^{\dagger}$ & 3.25$^{\dagger}$ \\
SimpleQA & 0.46 & 2.27$^{\dagger}$ & 2.29$^{\dagger}$ & 2.27$^{\dagger}$ & 2.27$^{\dagger}$ & 1.00$^{\dagger}$ & 2.27$^{\dagger}$ \\
PopQA & 0.53 & 3.17$^{\dagger}$ & 3.31$^{\dagger}$ & 3.15$^{\dagger}$ & 3.23$^{\dagger}$ & 1.08$^{\dagger}$ & 3.15$^{\dagger}$ \\
\midrule[0.3pt]
\addlinespace[2pt]
\multicolumn{8}{@{}l}{\textbf{Gemma-4-E4B-it}}\\[3pt]
TriviaQA & 0.69 & 1.71$^{\dagger}$ & 1.61$^{\dagger}$ & 1.66$^{\dagger}$ & 1.71$^{\dagger}$ & 1.00$^{\dagger}$ & 1.66$^{\dagger}$ \\
AmbigQA & 0.56 & 1.57 & 1.92 & 1.90 & \underline{1.55} & \textbf{1.00} & 1.90 \\
OpenNQ & 0.50 & 1.26 & 1.39 & 1.34 & \underline{1.19} & \textbf{1.00} & 1.34 \\
AmbigInst & 0.90 & 1.65 & \underline{1.57} & \textbf{1.51} & 1.66 & 1.66 & \textbf{1.51} \\
SimpleQA & 0.21 & 1.65$^{\dagger}$ & 1.63$^{\dagger}$ & 1.65$^{\dagger}$ & 1.65$^{\dagger}$ & 1.00$^{\dagger}$ & 1.65$^{\dagger}$ \\
PopQA & 0.38 & 1.81$^{\dagger}$ & 1.86$^{\dagger}$ & 1.86$^{\dagger}$ & 1.80$^{\dagger}$ & 1.00$^{\dagger}$ & 1.86$^{\dagger}$ \\
\bottomrule
\end{tabular*}
\endgroup

\end{table}

\begingroup
\setlength{\intextsep}{6pt}
\begin{appendixblock}
\section{Support failure AUROC}\label{app:auroc_bars}

\begin{table}[H]
\caption{Support failure AUROC at seed 42. Each $\Delta$ subtracts the AUROC of $\Utot$ on the [0,1] scale. \textbf{Bold} and \underline{underline} mark the best and second-best AUROC.}
\label{tab:auroc}
\centering
\small
\begingroup
\small
\setlength{\tabcolsep}{3.5pt}
\renewcommand{\arraystretch}{1.12}
\begin{tabular*}{\linewidth}{@{\extracolsep{\fill}}lrrrrrrr@{}}
\toprule
& & \multicolumn{4}{c}{Support-failure AUROC $\uparrow$} & \multicolumn{2}{c}{Difference vs. $\Utot$} \\
\cmidrule(lr){3-6}\cmidrule(lr){7-8}
Dataset & Match\% & $\Utot$ & $\Uepi$ & $\Uale$ & $R(x)$ & $\Delta_{\rm epi}$ & $\Delta_R$ \\
\midrule
\multicolumn{8}{@{}l}{\textbf{Qwen3-8B}}\\[3pt]
TriviaQA & 78.0 & 0.658 & \underline{0.709} & 0.571 & \textbf{0.730} & +0.050 & +0.072 \\
AmbigQA & 67.0 & 0.574 & \underline{0.632} & 0.449 & \textbf{0.677} & +0.059 & +0.104 \\
OpenNQ & 63.7 & \underline{0.629} & \textbf{0.650} & 0.581 & 0.419 & +0.021 & $-$0.210 \\
SimpleQA & 35.6 & \textbf{0.588} & \underline{0.583} & 0.558 & 0.442 & $-$0.005 & $-$0.146 \\
PopQA & 53.6 & 0.761 & \textbf{0.780} & 0.641 & \underline{0.764} & +0.019 & +0.003 \\
\midrule[0.3pt]
\addlinespace[2pt]
\multicolumn{8}{@{}l}{\textbf{Ministral-3-8B-Instruct-2512}}\\[3pt]
TriviaQA & 84.7 & 0.765 & \textbf{0.823} & 0.632 & \underline{0.803} & +0.058 & +0.038 \\
AmbigQA & 67.0 & \textbf{0.666} & \underline{0.665} & 0.632 & 0.453 & $-$0.001 & $-$0.213 \\
OpenNQ & 67.0 & 0.751 & \textbf{0.802} & 0.584 & \underline{0.763} & +0.051 & +0.011 \\
AmbigInst & 91.4 & \textbf{0.735} & \underline{0.724} & 0.682 & 0.318 & $-$0.011 & $-$0.418 \\
SimpleQA & 42.6 & 0.528 & \underline{0.565} & 0.470 & \textbf{0.598} & +0.037 & +0.069 \\
PopQA & 51.6 & 0.707 & \textbf{0.729} & 0.630 & \underline{0.721} & +0.022 & +0.014 \\
\midrule[0.3pt]
\addlinespace[2pt]
\multicolumn{8}{@{}l}{\textbf{Gemma-4-E4B-it}}\\[3pt]
TriviaQA & 68.7 & 0.727 & \textbf{0.773} & 0.606 & \underline{0.733} & +0.046 & +0.006 \\
AmbigQA & 58.0 & \underline{0.600} & \textbf{0.605} & 0.541 & 0.529 & +0.005 & $-$0.072 \\
OpenNQ & 51.7 & \textbf{0.672} & \underline{0.665} & 0.606 & 0.395 & $-$0.007 & $-$0.278 \\
AmbigInst & 96.1 & \textbf{0.849} & 0.766 & \underline{0.787} & 0.574 & $-$0.083 & $-$0.275 \\
SimpleQA & 21.2 & 0.389 & 0.399 & \underline{0.409} & \textbf{0.457} & +0.010 & +0.068 \\
PopQA & 38.0 & \textbf{0.785} & 0.749 & \underline{0.761} & 0.607 & $-$0.036 & $-$0.177 \\
\bottomrule
\end{tabular*}
\endgroup

\end{table}
\end{appendixblock}

\begin{figure}[H]
\centering
\includegraphics[width=0.94\textwidth]{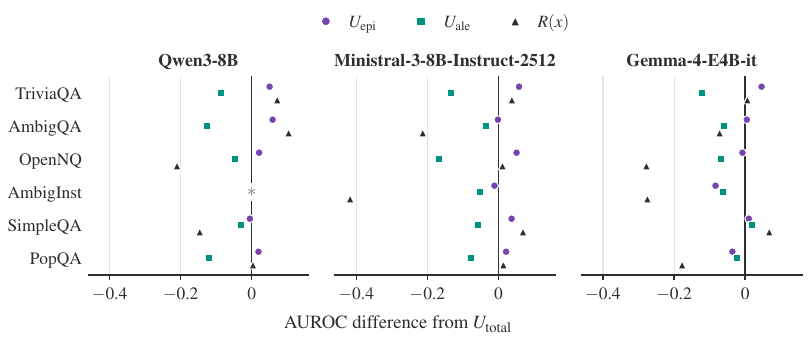}
\caption{Support failure AUROC differences relative to $\Utot$ at seed 42. Positive values indicate better ranking. $^\ast$Qwen3-8B/AmbigInst has no support failures and therefore no defined AUROC.}
\label{fig:auroc_signals}
\end{figure}
\endgroup
\paragraph{Paired AUROC comparisons.}
An inverse-variance meta-analysis using DeLong estimates across the 17 cells gives a fixed-effect $\Delta_\textsc{epi} = 0.019$ ($95\%$ CI $[0.007, 0.032]$, $p = 0.003$) and a random-effect estimate of $0.019$ ($p = 0.004$), with $I^2 = 0.04$ indicating negligible between-cell heterogeneity. The DeLong test accounts for per-cell AUROC variance, whereas a one-sided sign test uses only the $11/17$ wins ($p \approx 0.17$). Both summaries treat cells as independent, although the three generators answer the same questions.

\paragraph{Variation across splits.}
The multi-seed evaluation (seeds $42$--$45$) gives $\Uepi > \Utot$ on $15/17$ cells with mean $\bar\Delta_\textsc{epi} = +0.012$. $\Uepi$ seed std is at most $0.094$ (Gemma-AmbigInst) and is $\leq 0.052$ on the other cells.

\paragraph{The ratio score.}
$R(x)$ obtains gains over $\Utot$ on TriviaQA (Qwen3 $+0.072$, Ministral $+0.038$, Gemma $+0.006$) and on Qwen3-AmbigQA ($+0.104$), but underperforms $\Utot$ on $8/17$ cells, including all four with $\lambdastar=0$ (mean $\Delta_R = -0.083$ across the 17 cells). On a separate selective-answering routing AUC evaluation that uses randomized APS for weight selection, $R(x)$ loses to $\Uepi$ on $16/18$ cells (mean $\Delta_{R - \Uepi} = -0.046$, two-sided sign-test $p = 0.001$, Wilcoxon $p = 5\!\times\!10^{-5}$). Away from the numerical floor, $\log\lambdastar$ contributes a cell-wide constant to $R$, while the varying aleatoric denominator can change the ordering. At $\lambdastar=0$, the epistemic term disappears.

\section{Selective answering (full table)}\label{app:r_protocol}

\begin{table}[H]
\caption{Selective-answering experiment with deterministic APS for weight selection. AUC of accuracy as a function of escalation rate, when items are ranked by each signal (higher AUC = better). \emph{Random} is a reference policy. \emph{Oracle} (escalating incorrect items first) is the upper bound. $\Delta_{R(x)}$ is the absolute AUC gain of $R(x)$ over Random (\emph{not} a percentage). \textbf{Bold} and \underline{underline} mark the best and second-best signals in each row among $\{\Utot, \Uepi, R(x)\}$. We aggregate across the three models per dataset (mean over models).}
\label{tab:r_protocol}
\centering
\small
\begingroup
\small
\setlength{\tabcolsep}{3.5pt}
\renewcommand{\arraystretch}{1.12}
\begin{tabular*}{\linewidth}{@{\extracolsep{\fill}}lrrrrrr@{}}
\toprule
Dataset (mean over 3 LLMs) & Random & $\Utot$ & $\Uepi$ & $R(x)$ & Oracle & $\Delta_{R-\rm random}$ \\
\midrule
TriviaQA & 0.865 & \underline{0.917} & \textbf{0.923} & 0.915 & 0.963 & +0.049 \\
AmbigQA & 0.801 & \textbf{0.854} & \underline{0.852} & 0.819 & 0.928 & +0.018 \\
OpenNQ & 0.834 & \underline{0.868} & \textbf{0.871} & 0.827 & 0.936 & $-$0.007 \\
AmbigInst & 0.924 & \textbf{0.963} & \underline{0.956} & 0.910 & 0.986 & $-$0.014 \\
SimpleQA & 0.626 & \underline{0.645} & \textbf{0.647} & 0.631 & 0.728 & +0.005 \\
PopQA & 0.738 & \textbf{0.787} & \underline{0.785} & 0.766 & 0.838 & +0.028 \\
\bottomrule
\end{tabular*}
\endgroup

\end{table}

\section{Matched-stratum coverage with \texorpdfstring{$\lambdastar$}{lambda-star}-weighted scores}\label{app:matchable-with-lambdastar}

The empirical stratified-coverage results in \cref{sec:experiments} use the Vanilla-CP score ($\lambda = 0$) on the matchable stratum, to isolate the effect of stratifying by matchability. Replacing the score with the $\lambdastar$-weighted form $\snorm_{\lambdastar,i} = s_i / w_{\lambdastar}(x_i)$ also satisfies the bound because \cref{thm:stratified} holds for any $\lambda_0$ fixed or selected using $\Vone$, and $\lambdastar$ selected on $\Vone$ is by construction independent of $\Vtwo$ and the test point.

\paragraph{Integrated pipeline.} We (i) split each cell 25/25/25/25 into $\Vzero$ (matchability classifier on pre-generation features), $\Vone$ ($\lambdastar$ argmin), $\Vtwo$ (matched-stratum $\hat\tau$), and Test; (ii) run the standard two-parameter CAELUM calibration on $\Vone, \Vtwo$ to obtain $\lambdastar$; (iii) compute $\tauhat_{\mathrm{strat}}(\lambdastar)$ via \cref{eq:tau_strat} on the predicted-matchable subset of $\Vtwo$; (iv) construct $\Cset_{\lambdastar}^{\mathrm{strat}}(x)$ on items with $\hat m(x) = 1$ in test. Because the only change is the per-item weight and $\lambdastar$ is fixed before $\Vtwo$, the support-aware bound of \cref{thm:stratified} is unchanged.

\paragraph{Empirical instantiation.} Across the same $18$ cells (seed $42$, $4$-way split), the integrated variant passes the slack-adjusted empirical diagnostic on $10/18$ cells and gives matched-stratum coverage $0.72 \pm 0.21$, including three oracle-fallback rows. On the $15$ learned-classifier rows, matched-stratum coverage is $0.67 \pm 0.19$ and classifier $\Vtwo$ AUC is $0.64 \pm 0.07$. These $\pm$ values are standard deviations across cells. Selected $\lambdastar$ values include Qwen3-TriviaQA $29.8$, Ministral-AmbigQA $11.9$, and Gemma-AmbigQA $37.5$, with $0$ selected on $12$ cells. On the same $18$ cells, the body's $3$-way variant (pre-gen classifier, $\lambda = 0$ on the predicted-matchable stratum, $4$ random splits) gives learned-only matched-stratum coverage $0.71 \pm 0.14$ and full support-audit coverage $0.82 \pm 0.17$ when flagged oracle-fallback rows are included. The integrated variant checks the stratified construction with a $\lambdastar$-weighted score. The main empirical evaluation uses $\lambda=0$ to isolate matchability stratification. Per-cell numbers are reported in the supplementary release.

\begin{appendixblock}
\section{Per-cell stratified coverage figure}\label{app:stratified-figure}

\Cref{fig:stratified} reports per-cell stratified coverage under the pre-generation matchability classifier, with learned and oracle-fallback evaluations shown separately.

\begin{figure}[H]
\centering
\includegraphics[width=\textwidth]{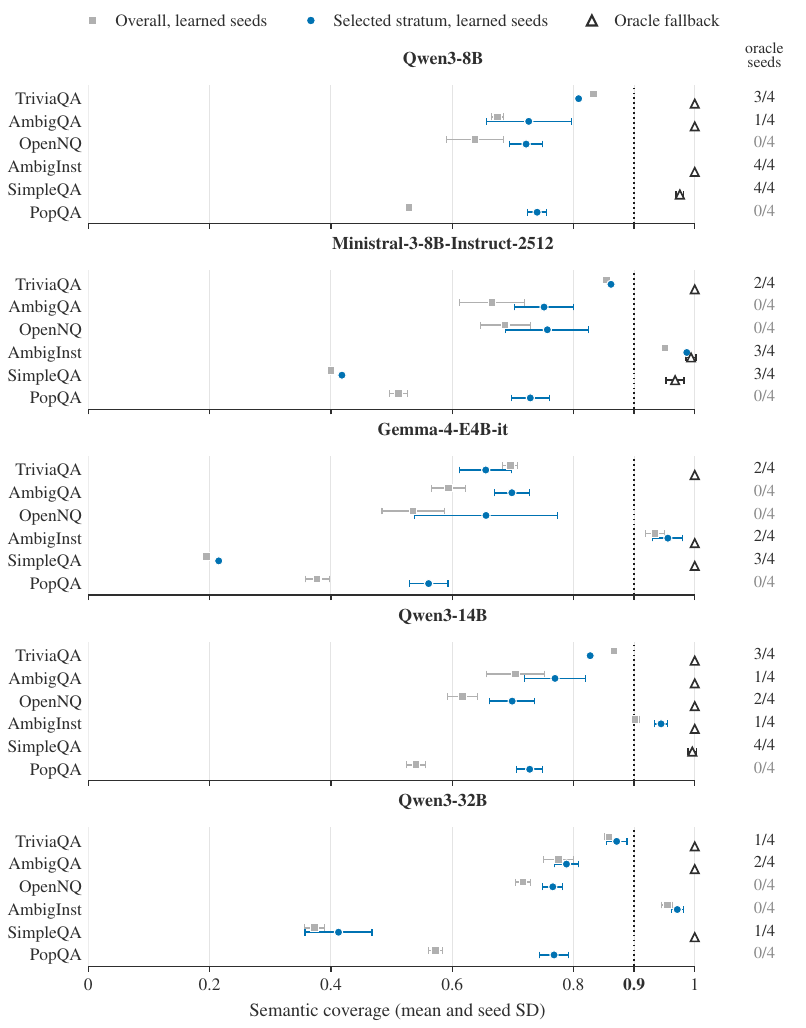}
\caption{Matchability stratification across four random splits. Blue circles and gray squares compare selected-stratum and overall coverage using only seeds with a learned classifier. Hollow black triangles show oracle-fallback stratum coverage separately. Counts at right give oracle seeds out of four. Error bars are split SD within each subset, so a one-seed subset has zero SD. The dotted line marks nominal $0.9$ coverage, not the $\rho_{\hat m}$-adjusted theorem bound. Missing learned points mean all four seeds used oracle labels.}
\label{fig:stratified}
\end{figure}
\end{appendixblock}

\begin{appendixblock}
\section{Cross-model matchability transfer (full table with bootstrap CIs)}\label{app:transfer}

\Cref{tab:transfer-full} reports the point AUCs, bootstrap CIs, $\Delta$, and label agreement. Pre-generation question features are the all-mpnet-base-v2 sentence embedding, character/token length, capitalized- and numeric-token counts, a year-pattern indicator, and a one-hot question-prefix bucket. We train a logistic regression on the matchability labels from one 4--9B LLM in our family, then evaluate AUC against another LLM's matchability labels on the same $2{,}104$ question records across the six datasets, with all six ordered transfer directions reported below. We align records by dataset and question ID and verify their question text and reference answers. Repeated question texts retain their own matchability labels, including when their references differ.

\begin{table}[H]
\caption{Cross-LLM transfer on $n=2{,}104$ aligned questions per model pair, reused for fitting and scoring. AUC$_{\mathrm{src}}$ and AUC$_{\mathrm{tgt}}$ score the same classifier against source and target LLM labels. Pairwise $95\%$ bootstrap CIs use $B=1000$ resamples of (label, score) pairs with the fitted model held fixed, so they omit uncertainty from fitting the model. $\Delta=\mathrm{AUC}_{\mathrm{tgt}}-\mathrm{AUC}_{\mathrm{src}}$. Agreement is the fraction of identical source and target labels. The mean row weights the six directions equally and reports point estimates only. These results measure label transfer on the same questions, not unseen-question generalization.}
\label{tab:transfer-full}
\centering
\small
\begingroup
\setlength{\tabcolsep}{5.5pt}
\begin{tabular*}{\linewidth}{@{\extracolsep{\fill}}lrlllr@{}}
\toprule
Source LLM $\to$ Target LLM & $n$ & AUC$_{\mathrm{src}}\uparrow$ ($95\%$ CI) & AUC$_{\mathrm{tgt}}\uparrow$ ($95\%$ CI) & $\Delta$ & Agreement \\
\midrule
Qwen3 $\to$ Ministral & 2104 & $0.793$ $[0.772, 0.813]$ & $0.771$ $[0.750, 0.791]$ & $-0.022$ & 0.831 \\
Qwen3 $\to$ Gemma & 2104 & $0.793$ $[0.772, 0.813]$ & $0.797$ $[0.778, 0.815]$ & $+0.004$ & 0.817 \\
Ministral $\to$ Qwen3 & 2104 & $0.788$ $[0.769, 0.807]$ & $0.771$ $[0.750, 0.790]$ & $-0.017$ & 0.831 \\
Ministral $\to$ Gemma & 2104 & $0.788$ $[0.769, 0.807]$ & $0.782$ $[0.763, 0.801]$ & $-0.006$ & 0.796 \\
Gemma $\to$ Qwen3 & 2104 & $0.821$ $[0.803, 0.838]$ & $0.782$ $[0.762, 0.801]$ & $-0.039$ & 0.817 \\
Gemma $\to$ Ministral & 2104 & $0.821$ $[0.803, 0.838]$ & $0.766$ $[0.745, 0.785]$ & $-0.055$ & 0.796 \\
\midrule
\textbf{Mean over 6 pairs} & \textemdash & $0.801$ & $0.778$ & $-0.023$ & $0.815$ \\
\bottomrule
\end{tabular*}
\endgroup

\end{table}
\end{appendixblock}

\begin{figure}[H]
\centering
\includegraphics[width=\textwidth]{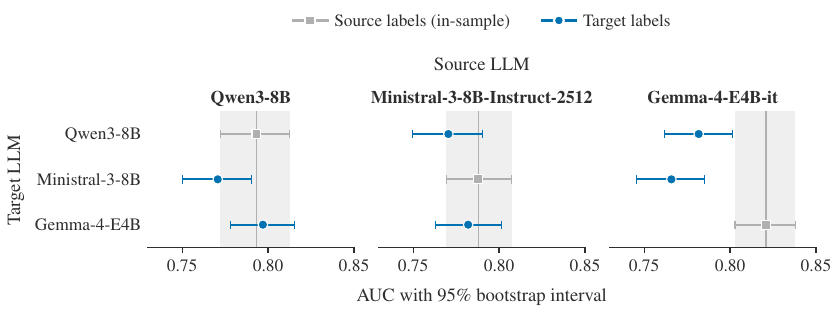}
\caption{Transfer of matchability labels across generators on the same 2,104 question records. Each panel names the source LLM whose matchability labels train the classifier. Rows name the target LLM whose labels are used for evaluation. Gray squares show AUC against source labels and blue circles show AUC against target labels, with 95\% bootstrap intervals. The source result sits in the row of the source LLM, and its interval is shaded across the panel for comparison. Fitting and evaluation reuse the same questions, so this is not an estimate of generalization to unseen questions.}
\label{fig:transfer}
\end{figure}

Across model pairs, roughly four out of five questions receive the same matchability label. For comparison, let $p_{\rm src}$ and $p_{\rm tgt}$ denote the source and target matchability rates. Independent-label agreement $p_{\rm src}p_{\rm tgt}+(1-p_{\rm src})(1-p_{\rm tgt})$ is $0.50$--$0.53$ at these marginal match rates. The agreement exceeds that independent-label baseline. Since the same questions and datasets are reused, shared benchmark structure and question difficulty may both contribute. \cref{fig:transfer} visualizes the per-pair AUCs.

\section{Reproducibility and further analyses}\label{app:further}
\subsection{Experiment protocols and reproducibility}\label{app:protocol}
The main experiments and additional analyses differ in their scores, matching rules, and data splits. A sweep over split seeds reuses generated answers. In the sweep of set sizes, CAELUM's calibrator seed follows the split seed, while Vanilla, Equal-weight, Epistemic-only, and TrueAPS retain the default calibrator seed of $42$. SE-CP uses deterministic APS. These sweeps do not measure variation across independent generation runs.

\begin{table}[H]
\caption{Protocols used for the main analyses. Both columns of the comparison at equal sample counts use newly generated answers. That comparison is separate from the other scale and judge analyses. Stratum rows that substitute oracle labels are reported separately as oracle reference results.}
\label{tab:protocol}
\centering
\footnotesize
\renewcommand{\arraystretch}{1.15}
\setlength{\tabcolsep}{4pt}
\begin{tabularx}{\linewidth}{@{}>{\raggedright\arraybackslash}p{0.21\linewidth}>{\raggedright\arraybackslash}X>{\raggedright\arraybackslash}X>{\raggedright\arraybackslash}X@{}}
\toprule
Evidence & Questions / labels & Scores & Split / seeds \\
\midrule
Main comparison and NCAPSS & 18 cells, heuristic & Deterministic APS & 40/30/30; 42, with $\lambda^\star$ SD over 42--45 \\
\addlinespace[3pt]
Sets across splits & Same generations & Randomized APS for CAELUM, Vanilla, Equal, Epi., TrueAPS; deterministic SE-CP & 40/30/30; 42--45 \\
\addlinespace[3pt]
Support failure ranking & 18 cells, heuristic & $\Utot$, $\Uepi$, $\Uale$, $R(x)$. For $R(x)$, deterministic APS in Table~\ref{tab:auroc}, randomized APS across splits & 40/30/30; 42 for Table~\ref{tab:auroc}, 42--45 across splits \\
\addlinespace[3pt]
Selective answering & 18 cells, heuristic & $R(x)$ with deterministic APS in Table~\ref{tab:r_protocol}; randomized APS for comparison with $\Uepi$ & 40/30/30; 42 \\
\addlinespace[3pt]
Learned stratum & 30 cells, heuristic & Randomized APS, $\lambda=0$ & $\Vzero/\Vtwo$/test 30/30/40; 42--45 \\
\addlinespace[3pt]
Integrated stratum & 18 cells, heuristic & Randomized APS, fitted $\lambda$ & Four 25\% splits; 42 \\
\addlinespace[3pt]
Transfer across generators & Aligned questions, heuristic & Logistic probability & Fit and score the same 2,104 questions \\
\addlinespace[3pt]
Scale comparison & 30 cells, judged & Standardized variant (\cref{app:standardized-weight}), randomized APS & 40/30/30; 42--45 \\
\addlinespace[3pt]
Classifier audit & 30 cells, judged & Question-only logistic probability & $\Vzero/\Vtwo$/test 30/30/40; 42--45 \\
\addlinespace[3pt]
Abstention & 29 cells, judged & Standardized variant; matched semantic calibration (\cref{app:missing-mass}) & Four 25\% splits; $\Vtwo$ halved; 42--45 \\
\addlinespace[3pt]
Scale at equal sample counts & 12 cells, heuristic & Support counts over all questions & Generation seed 42; no calibration split \\
\addlinespace[3pt]
\bottomrule
\end{tabularx}
\end{table}

\paragraph{Finite-sample quantiles.}
The theoretical construction uses the conformal rank $k=\lceil(1-\alpha)(n+1)\rceil$, with threshold $+\infty$ when $k>n$ (including an empty stratum). For $k\leq n$, the implementation used for these results calls NumPy's \texttt{quantile} at $k/n$ with \texttt{method='higher'}, which selects rank $k+1$ whenever $k<n$. For example, $n=60$, $\alpha=0.1$ gives rank 56 rather than 55. This is conservative at ordinary sample sizes. The stratified procedure returns $+\infty$ when fewer than ten calibration items are selected. The reported results use these numerical conventions.

\paragraph{Matching and tie handling.} Two further implementation details affect a few reported values. A PopQA reference alias consisting only of a flag emoji normalizes to an empty string and therefore matches any answer, which concerns six of the $10{,}520$ match labels. Set construction compares cumulative mass with $\tauhat\,\wlam(x)$ rather than the normalized score, so a cluster can be excluded at an exact floating-point tie. Changing both leaves every coverage value in \cref{tab:main} and the stratified summaries unchanged at the reported precision; it changes two APSS values in \cref{tab:main} by at most $0.02$, one deterministic coverage value at split seed $44$ ($0.880$ to $0.870$), a few PopQA-derived values in the third decimal, and several fitted $\lambdastar$ values, consistent with the nearly flat selection objective (\cref{app:stability}). The released code keeps both details, so rerunning it in the documented software environment reproduces the reported values.

\paragraph{Compute.}
The main experiments use 2,104 questions per generator, with five clarifications and 50 answers per question. We used DCSR clusters at the University of Lausanne for generation and uncertainty computation. Code, results, and details of GPU hardware and compute usage are available at \url{https://github.com/Unco3892/caelum}.

\paragraph{Assets and licenses.}
The repository documents the datasets, models, and software used in our experiments, with links to their sources and licenses. CAELUM's original code and software documentation use the MIT License. Third-party assets retain their respective licenses and attribution requirements. Model weights are obtained from their original publishers.

\subsection{Classifier operating points and PopQA}\label{app:operating}
The learned-only summaries in \cref{sec:experiments} exclude rows whose $\Vtwo$ AUC is below $0.55$. This filter uses $\Vtwo$ labels, so Theorem~\ref{thm:stratified} does not apply conditional on passing it.

The main matchability experiment uses a logistic classifier with $C=0.1$, class-balanced training, and threshold 0.5 on question features. Its oracle substitution after single-class training or calibration AUC below 0.55 uses labels unavailable at deployment. We exclude these rows from learned-only summaries. The learned-only summary is selected using calibration labels. Theorem~\ref{thm:stratified} therefore does not apply conditional on passing that filter. A deployment evaluation must fix the classifier and any retention rule before $\Vtwo$. The empirical coverage check compares selected-stratum test coverage with $1-\alpha-1/\sqrt{N}$, where $N$ is the number of calibration questions in the selected stratum. This tolerance is heuristic, not a confidence bound or a test of Theorem~\ref{thm:stratified}.

A separate audit uses the 30 cells with judged labels, seeds 42--45, and a 30/30/40 train/calibration/test split with no oracle substitution. With matchable as the positive class, macro AUROC is $0.620\pm0.107$ over 116 test folds containing both classes. Brier score is $0.216\pm0.056$ and ten-bin ECE is $0.180\pm0.090$ over all 120 folds. The $\pm$ values are fold standard deviations, not confidence intervals.

The pre-generation AUROC of $0.649$ in \cref{tab:detection-baselines} uses a different allocation, with 50\% of each cell for training and 25\% for testing, rather than 30\% and 40\% here. It is not a second estimate from the same fitted classifiers and test items.

\begin{table}[H]
\caption{Classifier audit with judged labels, macro-averaged over cell--seed folds. Precision and $\rho$ are defined only when the selected subset is nonempty (120 folds at thresholds 0.3 and 0.5, and 104 at 0.7). \textbf{Bold} and \underline{underline} mark the best and second-best displayed precision, recall, and false matchability rate. Higher thresholds reduce the selected population as well as its false matchability penalty.}
\label{tab:operating}
\centering
\small
\begin{tabular}{@{}rrrrr@{}}
\toprule
Threshold & Fraction selected & Precision $\uparrow$ & Recall $\uparrow$ & $\rho_{\hat m}\downarrow$ \\
\midrule
0.3 & 0.966 & 0.671 & \textbf{0.970} & 0.329 \\
0.5 & 0.578 & \underline{0.735} & \underline{0.636} & \underline{0.265} \\
0.7 & 0.192 & \textbf{0.839} & 0.207 & \textbf{0.161} \\
\bottomrule
\end{tabular}
\end{table}

For PopQA at threshold 0.5, pooling the held-out predictions over 20 overlapping model--seed folds gives selection fraction 0.424, precision 0.715, recall 0.593, and $\rho=0.285$. These numbers describe a different judged protocol from the stratum results based on heuristic labels. They do not close the PopQA coverage gap. An explanation based on the popularity tail requires an analysis conditioned on popularity. Observations about model size and sampling budget alone cannot identify that cause. Augmented abstention is a separate task, not a repair of the classifier using question features.

\paragraph{Sampling budget, escalation, and held-out model families.}
In three additional analyses, we ask whether more samples, a larger fallback generator, or a different training set would change our results. In the README of our repository, we describe commands for doing this. First, we take $m\in\{1,2,4,6,8,10\}$ previously generated answers per clarification, so between 5 and 50 answers per question. We then re-cluster the answers and apply our heuristic matcher. On the 26 cells where the match rate with 5 answers is below 0.9, the match rate increases by between 3.0 and 13.8 percentage points as we increase $m$ from 1 to 10. But the mean gain per additional answer per clarification falls from 2.2 percentage points between $m=1$ and $m=2$ to 0.3 percentage points between $m=8$ and $m=10$. The highest match rate on PopQA with 50 answers is 0.572 (Qwen3-32B). The four remaining AmbigInst cells all begin with a match rate of 0.9 or more, and none improve by more than 0.7 percentage points with additional answers.

Second, we evaluate an escalation policy using judged labels. When the base model's learned support-failure score reaches a threshold, the policy uses Qwen3-14B or Qwen3-32B as a fallback. We choose the escalation budget on $\Vone$ and then calibrate abstention using the outcomes of the combined base and fallback policy. On SimpleQA, the selected escalation budget was zero for every base model. So no query was escalated. On PopQA, some queries were escalated. But none of the items that would otherwise abstain became answered and covered with the Qwen3-14B fallback. With Qwen3-32B, at most 1.3\% did so on average over split seeds, under both the switching and merged-set policies.

Third, when we use the learned support-failure score from \cref{app:missing-mass}, with all cells of the target's model family excluded from training (all sizes of Qwen3 for a Qwen3 target), its mean detection AUROC across cells is 0.751. If we exclude only the target cell, it is 0.780. Both means are over 29 cells. They exclude Qwen3-8B--AmbigInst, which has no support failures under judged labels and therefore no defined AUROC.

A tenfold increase in budget raises PopQA matchability by 3.4 to 9.6 percentage points, but not enough to reach our target. Escalation converts almost no abstentions into covered answers. Taken together, these results are consistent with gaps in the generators' knowledge of these questions, but they do not establish why the failures occur.

\paragraph{Exploratory popularity probe.}
We examine popularity as a possible predictive feature by comparing logistic classifiers using the non-embedding question metadata with and without $\log(1+s_{\rm pop})$, where $s_{\rm pop}$ is PopQA's supplied subject-popularity value. The probe uses five PopQA cells with judged labels, seeds 42--45, 30/30/40 splits, threshold 0.5, and the same $C=0.1$ and class balancing. Scaling is fitted only on the training portion. No oracle labels replace predictions.

Across 20 overlapping test folds, mean AUROC rises from 0.688 to 0.795, precision from 0.710 to 0.757, and recall from 0.552 to 0.679. Selection rises from 0.397 to 0.458 and $\rho$ falls from 0.290 to 0.243. This exploratory comparison uses metadata features rather than the question embeddings used by the main classifier. Popularity is supplied dataset metadata and may not be available from a new question alone. The selected population's precision remains below 0.9. The association motivates further study of popularity as a feature, but deployment and a causal explanation remain unresolved.

\subsection{Reference-matching sensitivity}\label{app:matcher-audit}
The support ceiling applies to any fixed matching rule, but its measured value will depend on that rule. In an audit of 4,118 cases labeled ``unmatched'' by the heuristic matcher across the 30 cells and six datasets, Gemini 2.5 Flash-Lite found 168 matches, and measured matchability in each cell increased by 0--5 percentage points. Gemini 2.5 Flash independently looked at the same 4,118 cases. In 97.4\% of them, the two judges agreed on whether they were matched or not. Flash agreed with 73 of the 168 new matches found by Flash-Lite. And in no cell did their matchability estimates differ by more than 2.2 percentage points. This is a false-negative sensitivity check of the heuristic matcher. It does not measure the false-positive rate among matches accepted by the heuristic. This agreement applies only to cases adjudicated by the judges, not to all generated answers. And it would not be sufficient to assure the semantic correctness of any particular match they accepted.

Each judge is presented with the question, acceptable reference answers, and a numbered representative of each cluster. They are asked to select a cluster that asserts the same person, date, number, fact or entity as one of the reference answers. The cluster answer could use paraphrase, abbreviations or additional context, but not assert a different fact. They were asked to respond with \texttt{ANSWER: N}, where $N$ is the number of the cluster (starting at 1) or 0 if there was no match. Temperature was set to 0. A valid positive index corresponding to one of the clusters would replace the heuristic ``unmatched'' label. Otherwise, their response (if unparseable, out of range or zero) would leave the case unmatched.

Encoder and matching choices must be fixed before using $\Vtwo$ or test labels. Tuning them on a separate training or validation split preserves the split-conformal setup. Tuning them on $\Vtwo$ or test invalidates that setup. Maximizing match rate is not a sufficient tuning objective, since permissive similarity thresholds can increase apparent support while destroying correctness discrimination. The bge-large and e5-large results in Table~\ref{tab:embedding_ablation} illustrate this problem. The French medical QA judge study of \citet{belmadani2026judges} concerns a different domain and does not validate this matcher.

\subsection{Weight movement and stability}\label{app:stability}
In Section~\ref{sec:conformal}, we check whether any set in the selection grid changes. If none of the sets change, then the coefficient cannot be identified using APSS. With $V=\Vone$ and $K_i=K(x_i)$, a small fraction of changing sets would allow only a small change in the mean set size. Indeed, $\operatorname{range}(\mathrm{APSS})\leq |V|^{-1}\sum_i (K_i-1)\mathbf{1}\{i\text{ moves}\}$.

A separate sweep on held-out test items covers four split seeds and 30 cells, with $\lambda$ on a sparse grid from $0$ to $10^6$. At $\lambda=0$, 97.7\% of sets include all of the sampled clusters (96.3\% among items with more than one cluster). Over the sweep, 3.7\% of the test items experience a change in set size. Over the sweep's points up to 100, 3.6\% of them change. Using all questions in each cell and heuristic labels, 25 of 30 cells have support-failure rate above $\alpha$. On average over the four seeds, none of these cells changes more than 6.3\% of its sets. The other five cells, all of which are of type AmbigInst, move between 0.3\% and 37.2\% of their sets, the most on Ministral--AmbigInst (37.2\%) and Qwen3-32B--AmbigInst (12.8\%). Changes in set size are measured on the test split. Support failure rates use all questions in each cell.

We also look at how often the reference of a matched question across all splits lies in the highest-ranked cluster (under deterministic APS, this item would get score zero). In the five cells below $\alpha$, this holds for 93\% of matched items in Qwen3-14B and 94\% in Qwen3-8B. These sets move at most 1.6\%. For the other three cells, it holds for 79--91\% of matched items. Their sets move 11.4--37.2\%. Qwen3-14B has a failure rate of 0.095, below $\alpha$. Taken together, across these five cells, we could not reliably set a cutoff for how concentrated the matched scores are at zero. These shares refer to deterministic APS, whereas the sweep used APS with randomization, where the highest-ranked cluster does not get score zero.

\paragraph{Validation audit on the full grid.}
We also recompute the diagnostic on $\Vone$ using the 18 main cells, deterministic APS, the seed 42 split, and all 4,011 candidate weights. Table~\ref{tab:weight-stability} shows three completely flat cells. In the remaining cells, the objective distinguishes some weight regions but still has multiple minimizing grid points. All 18 main coefficients equal the first minimizer in the recomputed grid. Tied fractions range from 5 of 4,011 points to the entire grid. These fractions describe this nonuniform grid, not a confidence region or probability. For example, the minimizing region on Ministral--TriviaQA is narrow, approximately $[26.96,28.94]$, even though it contains more than one grid point.

Failure rates above $\alpha$ do not rule out movement on $\Vone$. Ministral--TriviaQA, Ministral--AmbigQA, and Qwen3-8B--TriviaQA have validation failure rates of 12.5\%, 41.25\%, and 23.3\%, but 29.2\%, 16.2\%, and 11.7\% of their sets move. Ministral--PopQA also has 44.5\% failure and 7.5\% movement. These validation results differ from the seed averages in the held-out sweep.

\begin{table}[H]
\caption{Validation calculation on the 18 main cells. $n_1$ is the selection-split size, $f_{\rm move}$ the fraction of items whose set size changes across the grid, $\Delta$APSS the maximum minus minimum validation APSS, and tied grid the fraction of candidate weights attaining that minimum. No cell has a unique minimizing grid point.}
\label{tab:weight-stability}
\centering
\footnotesize
\renewcommand{\arraystretch}{1.1}
\begin{tabular*}{\linewidth}{@{\extracolsep{\fill}}llrrrr@{}}
\toprule
Model & Dataset & $n_1$ & $f_{\rm move}$ (\%) & $\Delta$APSS & Tied grid (\%) \\
\midrule
Qwen3-8B & TriviaQA & 120 & 11.7 & 0.092 & 10.8 \\
 & AmbigQA & 80 & 1.2 & 0.013 & 37.3 \\
 & OpenNQ & 120 & 0.8 & 0.017 & 6.9 \\
 & AmbigInst & 121 & 0.0 & 0.000 & 100.0 \\
 & SimpleQA & 200 & 0.0 & 0.000 & 100.0 \\
 & PopQA & 200 & 3.5 & 0.020 & 44.6 \\
\midrule
Ministral-3-8B & TriviaQA & 120 & 29.2 & 0.258 & 0.2 \\
 & AmbigQA & 80 & 16.2 & 0.125 & 24.5 \\
 & OpenNQ & 120 & 4.2 & 0.033 & 40.2 \\
 & AmbigInst & 121 & 56.2 & 1.091 & 1.0 \\
 & SimpleQA & 200 & 1.0 & 0.010 & 26.0 \\
 & PopQA & 200 & 7.5 & 0.030 & 12.7 \\
\midrule
Gemma-4-E4B-it & TriviaQA & 120 & 5.0 & 0.042 & 8.1 \\
 & AmbigQA & 80 & 6.2 & 0.050 & 34.3 \\
 & OpenNQ & 120 & 0.0 & 0.000 & 100.0 \\
 & AmbigInst & 121 & 39.7 & 0.198 & 0.1 \\
 & SimpleQA & 200 & 1.0 & 0.005 & 41.1 \\
 & PopQA & 200 & 2.0 & 0.015 & 35.6 \\
\bottomrule
\end{tabular*}

\end{table}

\subsection{Qwen3-14B and Qwen3-32B}\label{app:historical-scale}
These results use different generated samples from the fixed sampling budget comparison in \cref{app:model-scale}. The same six benchmarks were evaluated with Qwen3-14B and Qwen3-32B. Table~\ref{tab:historical-scale} uses labels obtained by having a judge revisit heuristic negatives, 40/30/30 splits, and seeds 42--45. Coverage, APSS, and $\lambda^\star$ use the standardized variant of \cref{app:standardized-weight} with randomized APS and selection over all items in $\Vone^{\mathrm{eval}}$. AUROC is computed from the uncertainty signals on the same held-out subsets. The support rate uses all items, so it need not equal the split-mean coverage exactly. Across all 30 cells of this judged rerun, including the 18 main cells, mean CAELUM coverage tracks $\hat\pi$ with a maximum absolute gap of $0.030$ (Qwen3-14B--AmbigQA, $0.700$ versus $0.730$).

The five-model transfer matrix uses heuristic labels for the same 2,104 records, aligned by dataset and question ID. The classifiers are fitted and scored without a held-out question split. Mean source-label AUC is 0.793 and target-label AUC 0.772 over 20 ordered pairs. Every transfer interval has a lower endpoint above 0.72. Table~\ref{tab:historical-transfer} gives the complete matrix. This extension supports cross-generator label transfer for these questions. It does not establish transfer to new domains, unseen questions, or API models.

\begingroup
\setlength{\intextsep}{4pt}
\setlength{\abovecaptionskip}{4pt}
\begin{table}[H]
\caption{Results for larger generators. Values are means over four overlapping split seeds, except all-question $\hat\pi$. Coverage, APSS, and $\lambda^\star$ use the standardized variant of \cref{app:standardized-weight}. The $\lambda^\star$ column gives the mean $\pm$ SD over these seeds and is not directly comparable with \cref{tab:main}. \textbf{Bold} and \underline{underline} mark the larger and smaller displayed AUROCs, and displayed ties are both in bold. Win counts use unrounded values. $\Uepi$ exceeds $\Utot$ on 8 of these 12 cells, with mean difference $+0.031$. It does not improve ranking on every larger-model cell.}
\label{tab:historical-scale}
\centering
\footnotesize
\renewcommand{\arraystretch}{1.1}
\begingroup
\small
\setlength{\tabcolsep}{3.5pt}
\renewcommand{\arraystretch}{1.12}
\begin{tabular*}{\linewidth}{@{\extracolsep{\fill}}lrrrrrrr@{}}
\toprule
& \multicolumn{2}{c}{Support / coverage} & \multicolumn{3}{c}{Support-failure AUROC $\uparrow$} & & \\
\cmidrule(lr){2-3}\cmidrule(lr){4-6}
Dataset & $\hat\pi$ & Cov. & $\Uepi$ & $\Utot$ & $\Delta_{\rm epi}$ & APSS & $\lambda^\star$ \\
\midrule
\multicolumn{8}{@{}l}{\textbf{Qwen3-14B}}\\[3pt]
TriviaQA & 0.840 & 0.822 & \textbf{0.744} & \textbf{0.744} & -0.001 & 1.56 & 0.0$\pm$0.0 \\
AmbigQA & 0.730 & 0.700 & \textbf{0.756} & \underline{0.681} & +0.075 & 1.45 & 0.0$\pm$0.0 \\
OpenNQ & 0.690 & 0.686 & \textbf{0.718} & \underline{0.648} & +0.070 & 1.16 & 0.0$\pm$0.0 \\
AmbigInst & 0.918 & 0.910 & \textbf{0.801} & \underline{0.702} & +0.098 & 1.92 & 0.0$\pm$0.0 \\
SimpleQA & 0.370 & 0.365 & \textbf{0.547} & \underline{0.543} & +0.005 & 1.48 & 0.0$\pm$0.0 \\
PopQA & 0.550 & 0.542 & \textbf{0.746} & \underline{0.744} & +0.002 & 1.94 & 2.1$\pm$2.1 \\
\midrule[0.3pt]
\addlinespace[2pt]
\multicolumn{8}{@{}l}{\textbf{Qwen3-32B}}\\[3pt]
TriviaQA & 0.873 & 0.858 & \textbf{0.766} & \underline{0.741} & +0.025 & 1.61 & 0.9$\pm$1.2 \\
AmbigQA & 0.805 & 0.817 & \textbf{0.771} & \underline{0.711} & +0.060 & 1.53 & 0.9$\pm$1.6 \\
OpenNQ & 0.740 & 0.744 & \textbf{0.761} & \underline{0.701} & +0.061 & 1.22 & 0.0$\pm$0.0 \\
AmbigInst & 0.961 & 0.962 & \underline{0.696} & \textbf{0.718} & -0.021 & 2.63 & 21.3$\pm$19.0 \\
SimpleQA & 0.406 & 0.390 & \underline{0.566} & \textbf{0.568} & -0.002 & 1.80 & 0.0$\pm$0.0 \\
PopQA & 0.576 & 0.563 & \underline{0.768} & \textbf{0.770} & -0.002 & 2.43 & 1.3$\pm$2.2 \\
\bottomrule
\end{tabular*}
\endgroup

\end{table}

\begingroup
\small
\setlength{\tabcolsep}{3.5pt}
\setlength{\LTcapwidth}{\linewidth}
\setlength{\LTpre}{7pt}
\setlength{\LTpost}{7pt}
\renewcommand{\arraystretch}{1.2}
\begin{longtable}{@{}p{74pt}lrrr@{}}
\caption{Matchability label transfer across five generators on aligned questions. Source AUC is in-sample, and target AUC uses the same fitted scores. Bootstrap intervals use 1,000 resamples of record labels and fixed fitted scores. These estimates use saved generations from five models, aligned by dataset and question ID. The three-model table uses the corresponding 18-cell subset. The scale comparison in \cref{app:model-scale} uses different generations.}
\label{tab:historical-transfer}\\
\toprule
\textbf{Source LLM} & \textbf{Target LLM} & \textbf{Source AUC}$\uparrow$ & \textbf{Transfer AUC}$\uparrow$ (95\% CI) & \textbf{Agreement} \\
\midrule
\endfirsthead
\multicolumn{5}{@{}l}{\textit{Table~\thetable\ (continued)}}\\[3pt]
\toprule
\textbf{Source LLM} & \textbf{Target LLM} & \textbf{Source AUC}$\uparrow$ & \textbf{Transfer AUC}$\uparrow$ (95\% CI) & \textbf{Agreement} \\
\midrule
\endhead
\midrule
\multicolumn{5}{r@{}}{\footnotesize Continued on next page}\\
\endfoot
\bottomrule
\endlastfoot
\multirow{4}{74pt}{\raggedright\textbf{Qwen3-8B}} & Ministral-3-8B-Instruct-2512 & 0.793 & 0.771 [0.750, 0.791] & 0.831 \\*
 & Gemma-4-E4B-it & 0.793 & 0.797 [0.778, 0.815] & 0.817 \\*
 & Qwen3-14B & 0.793 & 0.752 [0.730, 0.772] & 0.854 \\*
 & Qwen3-32B & 0.793 & 0.763 [0.742, 0.782] & 0.846 \\
\midrule[0.3pt]
\addlinespace[2pt]
\multirow{4}{74pt}{\raggedright\textbf{Ministral-3-8B-Instruct-2512}} & Qwen3-8B & 0.788 & 0.771 [0.750, 0.790] & 0.831 \\*
 & Gemma-4-E4B-it & 0.788 & 0.782 [0.763, 0.801] & 0.796 \\*
 & Qwen3-14B & 0.788 & 0.747 [0.724, 0.767] & 0.822 \\*
 & Qwen3-32B & 0.788 & 0.764 [0.742, 0.784] & 0.831 \\
\midrule[0.3pt]
\addlinespace[2pt]
\multirow{4}{74pt}{\raggedright\textbf{Gemma-4-E4B-it}} & Qwen3-8B & 0.821 & 0.782 [0.762, 0.801] & 0.817 \\*
 & Ministral-3-8B-Instruct-2512 & 0.821 & 0.766 [0.745, 0.785] & 0.796 \\*
 & Qwen3-14B & 0.821 & 0.755 [0.733, 0.775] & 0.797 \\*
 & Qwen3-32B & 0.821 & 0.769 [0.749, 0.790] & 0.786 \\
\midrule[0.3pt]
\addlinespace[2pt]
\multirow{4}{74pt}{\raggedright\textbf{Qwen3-14B}} & Qwen3-8B & 0.778 & 0.781 [0.762, 0.801] & 0.854 \\*
 & Ministral-3-8B-Instruct-2512 & 0.778 & 0.773 [0.752, 0.791] & 0.822 \\*
 & Gemma-4-E4B-it & 0.778 & 0.800 [0.781, 0.818] & 0.797 \\*
 & Qwen3-32B & 0.778 & 0.782 [0.762, 0.802] & 0.870 \\
\midrule[0.3pt]
\addlinespace[2pt]
\multirow{4}{74pt}{\raggedright\textbf{Qwen3-32B}} & Qwen3-8B & 0.783 & 0.769 [0.748, 0.788] & 0.846 \\*
 & Ministral-3-8B-Instruct-2512 & 0.783 & 0.766 [0.744, 0.785] & 0.831 \\*
 & Gemma-4-E4B-it & 0.783 & 0.787 [0.767, 0.806] & 0.786 \\*
 & Qwen3-14B & 0.783 & 0.756 [0.735, 0.777] & 0.870 \\

\end{longtable}
\endgroup
\endgroup

\subsection{Detection of support failure}\label{app:detection-baselines}
We compare the additional support-failure scores in Table~\ref{tab:detection-baselines} on the same 29 cells with judged labels, excluding Qwen3-8B/AmbigInst because it has no support failures. The evaluation uses seeds 42--45 and the four-way split described below. CAELUM's learned detector is trained on other cells and evaluated on the omitted cell. The per-cell logistic and gradient-boosting scores use features available after generation. The pre-generation classifier uses question features only, so it has access to less information.

\begin{table}[H]
\caption{Additional detector comparison on 29 cells with judged labels. References identify score definitions. All values come from our evaluation. For the frequency baseline, $p_1$ is the largest cluster mass. Values are mean $\pm$ population SD across cell means, after averaging over seeds 42--45. Higher AUROC and AUPRC and lower FPR at 90\% TPR are better. AUPRC's chance baseline is the support-failure prevalence, which varies by cell. \textbf{Bold} and \underline{underline} indicate the best and second-best displayed mean, including ties. These are different splits and labels from the main 18-cell AUROC table.}
\label{tab:detection-baselines}
\centering
\begingroup
\small
\setlength{\tabcolsep}{3.5pt}
\renewcommand{\arraystretch}{1.12}
\begin{tabular*}{\linewidth}{@{\extracolsep{\fill}}lrrr@{}}
\toprule
Support-failure score & AUROC $\uparrow$ & AUPRC $\uparrow$ & FPR at 90\% TPR $\downarrow$ \\
\midrule
CAELUM detector (ours) & \textbf{0.780}$\pm$0.08 & \textbf{0.648}$\pm$0.16 & \textbf{0.542}$\pm$0.17 \\
Logistic regression (per cell) & \underline{0.716}$\pm$0.09 & \underline{0.573}$\pm$0.17 & 0.661$\pm$0.16 \\
Gradient boosting (per cell) & 0.709$\pm$0.10 & 0.566$\pm$0.17 & 0.659$\pm$0.16 \\
$\Uepi$~\citep{spectraluncertainty2025} & 0.696$\pm$0.10 & 0.529$\pm$0.17 & \underline{0.636}$\pm$0.15 \\
$\Utot$~\citep{spectraluncertainty2025} & 0.674$\pm$0.10 & 0.510$\pm$0.16 & 0.678$\pm$0.16 \\
$\Uale$~\citep{spectraluncertainty2025} & 0.604$\pm$0.09 & 0.457$\pm$0.16 & 0.749$\pm$0.14 \\
$1-p_1$ (cluster frequency) & 0.585$\pm$0.09 & 0.434$\pm$0.17 & 0.958$\pm$0.12 \\
Pre-generation classifier (ours) & 0.649$\pm$0.09 & 0.506$\pm$0.17 & 0.720$\pm$0.12 \\
Good--Turing missing mass~\citep{noorani2025beyond} & 0.550$\pm$0.05 & 0.400$\pm$0.19 & 1.000$\pm$0.00 \\
\bottomrule
\end{tabular*}
\endgroup

\end{table}

\subsection{Missing mass and abstention at a fixed sampling budget}\label{app:missing-mass}
We compare rules that add a catch-all label for a reference absent from the sampled support. Its augmented coverage counts either a matchable reference included in the semantic set, or an unmatchable reference accompanied by that label. This is an enlarged output space. It does not increase the rate at which correct semantic answers are generated.

The comparison considers 29 non-degenerate cells, omitting Qwen3-8B--AmbigInst. Four split seeds are used for each cell, with judged support labels. To form semantic sets, the standardized variant of \cref{app:standardized-weight} is used, including its restriction to matched items. From each cell, four splits are taken at 25\% each. For weight selection, $\Vone$ is used. To fit the semantic threshold, the first half of $\Vtwo$ is used. To calibrate triggers, the second half is used. The final 25\% is used as test data. CAELUM's detector is trained on cells other than this one. It therefore does not see the first quarter of this cell. To compare against the calibration module in CPQ, the full $\Vtwo$ and a fixed budget of 50 answers are used.

\paragraph{Learned support-failure detector.}
This comparison is separate from the main results across the 18 cells. The detector is a logistic regression (L2-regularized, with $C=1$, balanced class weights) of features standardized over the pool of records from the other cells. Its positive label is lack of a reference match in the sampled clusters under the judged support labels. The 15 features are $\Uepi, \Uale, \Utot, p_1$ and $p_2$ (the two largest cluster masses, with $p_2 = 0$ for a single cluster), their difference, $K$, $1/K$, the Shannon entropy of the cluster masses, the maximum and standard deviation of the entropies of the clarifications, a pre-generation matchability probability of the question, its length in characters and in tokens separated by whitespace, and the proportion of generated answers that refuse the question. For each of the other 29 cells in the full 30-cell collection, all its records are included in the pool. But the cell in question is excluded. One of the features is a pre-generation matchability probability. This comes from a logistic regression using question-only features to predict class (match vs. no match), with $C = 0.1$ and balanced class weights on the pool. On this pool, therefore, the probabilities are in-sample. Both regressions use seed 42 and a 5,000-iteration limit. No hyperparameter tuning was done in either regression. And the scores of items in the target cell were kept fixed across the four splits of the data. Even if a cell is excluded, it is possible that the same questions from another generator will appear in another cell in the pool. The family-held-out analysis in \cref{app:operating} tests a different exclusion rule.

\paragraph{Abstention threshold.}
Let $a(x)$ be the support-failure score. The catch-all label is added when $a(x)\geq t$. On the second half of $\Vtwo$, let $n_b$ be the sample size and $\widehat R_b(t)$ the fraction not covered by either the semantic set or a correctly added catch-all label. Of the scores $0$, $1+10^{-9}$, and those actually observed, the rule picks the largest value of $t$ for which
\[
    \frac{n_b}{n_b+1}\widehat R_b(t)+\frac{1}{n_b+1}\leq\alpha.
\]
If none of these values satisfies this inequality, the catch-all label is added to all sets. In this case, it could still be that a semantic set misses a matched reference, which the catch-all label does not cover. The same threshold rule supports the Good--Turing trigger.

We evaluate the Good--Turing missing mass score used by \citet{noorani2025beyond} and their CPQ calibration module. The Good--Turing trigger uses our risk-controlled threshold rule. Neither row reproduces their adaptive querying policy.

Table~\ref{tab:missing-mass} reports means across cell means. Abstention rate is the frequency of including the added label. Precision is the fraction of such cases where the reference is absent from support. Parentheses count cells with defined precision. The other means use all 29 cells. Ordinary semantic coverage is $0.652\pm0.19$ and semantic APSS is $2.021\pm0.77$ (cell SD). The CPQ calibration rule has no standalone ranking score, hence the dash. Its higher augmented coverage also comes with more abstention.

At the fixed sampling budget, the calibration module attaches the catch-all label to every set on 24 of 29 cells. CAELUM's learned detector reduces that frequency in this experiment, while its 0.459 abstention rate remains substantial. This extension operates after generation and addresses a different decision from the classifier using question features in Theorem~\ref{thm:stratified}. It does not close that classifier's PopQA gap.

\section{Additional support audits and escalation pilots}\label{app:model-scale}

\paragraph{Generator scale at a fixed sampling budget.}
To test whether scaling the generator removes the sampled-support limitation, we regenerate answers with Qwen3-8B~\citep{qwen3} and Qwen3-32B~\citep{qwen3_32b} on the same $2{,}104$ questions and reference answers as the main experiment. Both models use FP16 without $4$-bit quantization, generation seed $42$, $n=5$ self-generated clarifications at temperature $0.7$, and $m=10$ answers per clarification at temperature $0.5$ (up to $512$ new tokens for the clarification list and $128$ per answer, with thinking mode disabled). We retain all-mpnet-base-v2 with clustering threshold $0.7$, VNE $\gamma=1$, and the original heuristic reference matcher (\cref{app:implementation}). Both columns of \cref{tab:fresh_scale_support} come from this generation campaign. Neither reuses the main table's generations. The ordered questions and references are identical across the two columns.

\begin{table}[H]
\caption{Support rates at a fixed generation budget. $\hat\pi$ is the fraction of references matched to any generated cluster under the heuristic matcher, not calibrated semantic coverage. Counts cover all queries, not a held-out test split. Percentages and differences are descriptive estimates from one generation seed. The final row pools all $2{,}104$ questions with equal weight. \textbf{Bold} and \underline{underline} mark the higher and lower support rates within each row.}
\label{tab:fresh_scale_support}
\centering
\small
\begingroup
\small
\setlength{\tabcolsep}{3.5pt}
\renewcommand{\arraystretch}{1.15}
\begin{tabular*}{\linewidth}{@{\extracolsep{\fill}}lcccc@{}}
\toprule
\textbf{Dataset} & \textbf{$N$} & \multicolumn{2}{c}{\textbf{Matched references / $N$ (rate)}} & \textbf{$\Delta\hat\pi$ (pp)} \\
\cmidrule(lr){3-4}
& & Qwen3-8B & Qwen3-32B & \\
\midrule
TriviaQA & 300 & $232/300$ (\underline{77.3\%}) & $253/300$ (\textbf{84.3\%}) & $+7.0$ \\
AmbigQA & 200 & $128/200$ (\underline{64.0\%}) & $155/200$ (\textbf{77.5\%}) & $+13.5$ \\
OpenNQ & 300 & $184/300$ (\underline{61.3\%}) & $216/300$ (\textbf{72.0\%}) & $+10.7$ \\
AmbigInst & 304 & $301/304$ (\textbf{99.0\%}) & $295/304$ (\underline{97.0\%}) & $-2.0$ \\
SimpleQA & 500 & $175/500$ (\underline{35.0\%}) & $183/500$ (\textbf{36.6\%}) & $+1.6$ \\
PopQA & 500 & $265/500$ (\underline{53.0\%}) & $287/500$ (\textbf{57.4\%}) & $+4.4$ \\
\midrule
\textbf{All questions} & 2,104 & $1285/2104$ (\underline{61.1\%}) & $1389/2104$ (\textbf{66.0\%}) & $+4.9$ \\
\bottomrule
\end{tabular*}
\endgroup

\end{table}

The $32$B generator has higher empirical matchability on five of six datasets, with the largest gains on AmbigQA, OpenNQ, and TriviaQA. AmbigInst is already near saturation and decreases slightly. The pooled rate rises from $61.1\%$ to $66.0\%$, but SimpleQA and PopQA remain at $36.6\%$ and $57.4\%$. Scaling therefore does not remove the missing-support problem in this fixed sampling budget audit. These measurements do not establish calibrated coverage, cross-scale classifier transfer, or an advantage in epistemic ranking. Those require separate evaluations. They also do not identify the cause of the remaining errors, such as entity popularity.

\paragraph{Scope of the escalation pilots.}
The following escalation and abstention pilots use previously generated answers and different protocols from the scale comparison above. They are \emph{not} used for the main claims or pooled with those support rates.

\paragraph{Phi-4 fallback escalation.}
We evaluate a model-escalation policy on $96$ pair-seed evaluations using four base LLMs (the three main models plus Qwen3-14B), six datasets, and seeds $42$--$45$, with Phi-4 as the fallback answer generator. The policy answers with the base model unless a risk score computed from the base model triggers escalation. No non-oracle policy uses test labels. Phi-4 is stronger on average. Base top-1 accuracy is $0.589$ versus fallback $0.669$, and base support is $0.629$ versus fallback support $0.713$. For set-valued escalation, base coverage is $0.623$ versus fallback coverage $0.706$, base APSS is $1.88$ versus fallback APSS $1.83$, and the union support is $0.766$. Phi-4 has higher mean support in this pilot, but the useful-escalation pool is only $0.143$ for top-1 accuracy and $0.139$ for set coverage, while the harmful-escalation pool is non-negligible ($0.062$ and $0.056$ respectively).

\begin{table}[H]
\caption{Phi-4 escalation pilot using previously generated answers. Each entry averages the 96 combinations of base generator, dataset, and split equally. Best non-oracle rows are selected separately at each fixed escalation budget. Oracle rows use test labels and are upper bounds, not deployable policies. Full summaries are in the supplementary release.}
\label{tab:model-scale-pilot}
\centering
\small
\setlength{\tabcolsep}{3.5pt}
\renewcommand{\arraystretch}{1.15}
\begin{tabular*}{\linewidth}{@{\extracolsep{\fill}}lcccc@{}}
\toprule
\textbf{Policy} & \textbf{Escalation budget} & \textbf{Accuracy / coverage} & \textbf{$\Delta$ vs base} & \textbf{Oracle $\Delta$} \\
\midrule
\multicolumn{5}{@{}l}{\textbf{Answer top-1}}\\[2pt]
LearnedNet-RF & $20\%$ & $0.615$ acc. & $+0.027$ & $+0.134$ \\
CAELUM margin & $50\%$ & $0.642$ acc. & $+0.054$ & $+0.143$ \\
\midrule[0.3pt]
\multicolumn{5}{@{}l}{\textbf{Set coverage}}\\[2pt]
LearnedNet-RF & $20\%$ & $0.646$ cov. & $+0.023$ & $+0.131$ \\
Learned logistic & $50\%$ & $0.674$ cov. & $+0.051$ & $+0.139$ \\
\bottomrule
\end{tabular*}
\end{table}

The ranking scores only partially identify helpful escalation. For answer escalation, the best helpful-escalation AUROC is $0.625$ for $\Utot$ and $0.613$ for $\Uepi$. For set-valued escalation, $\Uepi$ is best at $0.603$. A useful escalation rule must distinguish cases where the stronger model adds missing support from those where it replaces a correct base answer with a wrong one.

\paragraph{Qwen3.5-35B abstain pilot.}
We also ran a small Qwen3.5-35B-A3B pilot on OpenNQ, SimpleQA, and PopQA with an augmented label space that adds ``abstain (reference absent from sampled support)''. This tests abstention when the reference is absent from sampled support. The augmented set reaches $0.978$ mean coverage over the three cells, compared with $0.689$ semantic coverage on generated clusters, but it does so with a high abstain rate ($0.822$) and low abstain precision for unmatched items ($0.358$). Frequent inclusion of the abstention label raises recall for support failures, so high augmented coverage alone does not establish a deployable answering policy.

\begin{table}[H]
\caption{Qwen3.5-35B-A3B pilot with an added abstention label on three factual datasets. Semantic coverage and APSS count only generated clusters. Augmented metrics also count the ``abstain (not in support)'' label. Abstention rate is its frequency of inclusion. The final row averages the three datasets equally, each with 30 test questions. Lower APSS and abstention rates are preferable at comparable coverage.}
\label{tab:qwen35-abstain-pilot}
\centering
\small
\setlength{\tabcolsep}{3.5pt}
\renewcommand{\arraystretch}{1.15}
\begin{tabular*}{\linewidth}{@{\extracolsep{\fill}}lcccccc@{}}
\toprule
\textbf{Dataset} & \textbf{Match rate} & \multicolumn{2}{c}{\textbf{Coverage}} & \multicolumn{2}{c}{\textbf{APSS}} & \textbf{Abstention rate} \\
\cmidrule(lr){3-4}\cmidrule(lr){5-6}
& & Semantic & Augmented & Semantic & Augmented & \\
\midrule
OpenNQ & $0.767$ & $0.767$ & $1.000$ & $1.47$ & $2.30$ & $0.833$ \\
SimpleQA & $0.633$ & $0.633$ & $0.967$ & $2.33$ & $3.30$ & $0.967$ \\
PopQA & $0.667$ & $0.667$ & $0.967$ & $4.17$ & $4.83$ & $0.667$ \\
\midrule
\textbf{Dataset mean} & $0.689$ & $0.689$ & $0.978$ & $2.66$ & $3.48$ & $0.822$ \\
\bottomrule
\end{tabular*}
\end{table}

\begin{appendixblock}
\section{Examples of generated answers}\label{app:examples}
Table~\ref{tab:examples} contains three Qwen3-8B/TriviaQA records checked against the source uncertainty and raw generation records. Questions, reference answers, and representative responses are reproduced verbatim. Cluster proportions and uncertainty values come from those records. We recompute $R$ with the main cell's $\lambda^\star$ and $\epsilon=10^{-6}$, and determine set size using its main threshold. These examples illustrate the recorded outputs, not identified uncertainty regimes.

\setlength{\intextsep}{4pt}
\setlength{\abovecaptionskip}{4pt}
\begin{table}[H]
\caption{Examples from the Qwen3-8B/TriviaQA cell in the main experiments. Labels describe whether the reference matches sampled support. They do not infer an intervention's benefit from $R$ or treat prompt variability as annotated ambiguity. A generated response can be incorrect even when it is reproduced faithfully here.}
\label{tab:examples}
\centering
\small
\renewcommand{\arraystretch}{1.1}
\begingroup
\small
\renewcommand{\arraystretch}{1.0}
\begin{tabularx}{\linewidth}{@{}lX@{}}
\toprule
\multicolumn{2}{@{}l}{\textbf{1. Reference in a single sampled cluster}} \\
Source & Qwen3-8B / TriviaQA, question ID 251; zero-based cache row 251. \\
Question & In which city was Pablo Picasso born, in 1881? \\
Reference & Malaga \\
Clusters & (1.00) ``Pablo Picasso was born in Malaga, Spain, in 1881.'' \\
Signals & $\Uale=0.212$, $\Uepi=0.007$, $R(x)=+0.45$, $|\Cset(x)|=1$. \\
\midrule
\multicolumn{2}{@{}l}{\textbf{2. Reference absent from sampled clusters}} \\
Source & Qwen3-8B / TriviaQA, question ID 39; zero-based cache row 39. \\
Question & The Museum of Bags and Purses is in which European city? \\
Reference & Amsterdam \\
Clusters & (0.72) ``The Museum of Bags and Purses is located in the European city of Antwerp, Belgium.'' \\
 & (0.28) ``The Museum of Bags and Purses is located in the European city of Madrid, Spain.'' \\
Signals & $\Uale=0.270$, $\Uepi=0.392$, $R(x)=+4.23$, $|\Cset(x)|=2$. \\
\midrule
\multicolumn{2}{@{}l}{\textbf{3. Reference present among three sampled clusters}} \\
Source & Qwen3-8B / TriviaQA, question ID 35; zero-based cache row 35. \\
Question & Who directed the Noughties films One Day in September, The Last King of Scotland and Touching the Void? \\
Reference & KEVIN MACDONALD \\
Clusters & (0.96) ``The director responsible for the Noughties films *One Day in September*, *The Last King of Scotland*, and *Touching the Void* is Kevin Macdonald.'' \\
 & (0.02) ``The three Noughties films were directed by Andy Anderson.'' \\
 & (0.02) ``The three Noughties films were directed by Mike Leigh.'' \\
Signals & $\Uale=0.429$, $\Uepi=0.412$, $R(x)=+3.82$, $|\Cset(x)|=3$. \\
\bottomrule
\end{tabularx}
\endgroup

\end{table}
\end{appendixblock}

\paragraph{Scope of the checks.}
Raw generation strings verify the origin of the displayed representatives. The uncertainty records contain proportions but not a member-to-cluster assignment for every answer. Multiplying each proportion by 50 recovers integer cluster counts, but is not an independent rerun of the embedding clustering. The examples therefore verify the source records and arithmetic rather than certify every semantic-clustering decision.

\end{document}